\documentclass[hidelinks,11pt,a4paper+]{article}
\usepackage{lineno} 
\usepackage{amsmath}  
\usepackage{graphicx}
\usepackage{subcaption}
\usepackage{latexsym}
\usepackage{amssymb}
\usepackage{amsthm}
\usepackage{amsfonts}
\usepackage{booktabs}
\usepackage{bbm}
\usepackage{lineno}
\usepackage{algorithm}
\usepackage{algorithmic} 
\usepackage[ruled,vlined,linesnumbered,algo2e,resetcount]{algorithm2e}
\usepackage[usenames, dvipsnames]{color}
\usepackage[round]{natbib}
\usepackage{authblk}
\usepackage{hyperref}
\usepackage{multirow}
\usepackage{lipsum}
\usepackage{mwe}
\usepackage{threeparttable}
\usepackage{lscape}
\newcounter{equationset}	
\usepackage{mathtools}
\usepackage{appendix}
\usepackage{setspace} 
\usepackage{todonotes}
\usepackage{etoolbox}
\presetkeys{todonotes}{color=yellow!20}{}
\usepackage{titlesec}
\usepackage{comment}
\usepackage[top=2.7cm]{geometry}
\usepackage{longtable}
\usepackage{array}
\usepackage{caption}
\usepackage{lscape}
\usepackage{ulem}

\newcommand*\linenomathpatch[1]{%
  \cspreto{#1}{\linenomath}%
  \cspreto{#1*}{\linenomath}%
  \csappto{end#1}{\endlinenomath}%
  \csappto{end#1*}{\endlinenomath}%
}

\apptocmd{\thebibliography}{\setlength{\itemsep}{3.5pt plus 0pt}}{}{}

\linenomathpatch{equation}
\linenomathpatch{gather}
\linenomathpatch{multline}
\linenomathpatch{align}
\linenomathpatch{alignat}
\linenomathpatch{flalign}

\makeatletter
\newcommand{\srcsize}{\@setfontsize{\srcsize}{8.5pt}{8.5pt}}
\newcommand{\Tablesize}{\@setfontsize{\srcsize}{8pt}{8pt}}
\makeatother
\DeclarePairedDelimiter\ceil{\lceil}{\rceil}

\graphicspath{{fig/},{/data/eps/}}
\DeclareGraphicsExtensions{.eps,eps}
\titlespacing\section{2pt}{6pt}{2pt}
\titlespacing\subsection{2pt}{4pt}{0pt}
\titlespacing\subsubsection{2pt}{2pt}{0pt}
\newtheorem{prop}{Proposition}
\newcommand{\beq}[1]{\begin{equation}\label{#1}}
\newcommand{\eeq}{\end{equation}}

\newcommand{\wg}{\color{black}}  
\newcommand{\mm}{\color{black}}

\makeatletter
\newcommand*{\rom}[1]{\expandafter\@slowromancap\romannumeral #1@}

\makeatother

\title{A Branch-and-Price Algorithm for Fast and Equitable Last-Mile Relief Aid Distribution}
\author[1]{Mahdi Mostajabdaveh}
\author[2]{F. Sibel Salman \thanks{Corresponding Author. \\ 
 E-mail address: \href{ssalman@ku.edu.tr }{ssalman@ku.edu.tr } Phone: +90 505 401 9602}}
\author[3]{Walter J. Gutjahr}
\affil[1]{Department of Mathematics and Industrial Engineering, Polytechnique Montr\'{e}al, 2500 Chem., Montreal, H3T 1J4, Canada }
\affil[2]{College of Engineering, Ko{\c c} University, Sar{\i}yer, 34450, Istanbul, Turkey}
\affil[3]{Department of Statistics and Operations Research, University of Vienna, Universit\"{a}tsring 1, 1010, Vienna, Austria }

\begin{document}

\setlength{\abovedisplayskip}{4pt}
\setlength{\belowdisplayskip}{2pt}
\setlength\paperheight {279.4mm}%
\setlength\paperwidth  {210mm}%
\setlength{\belowcaptionskip}{-10pt}
\allowdisplaybreaks









\maketitle

\newpage

\begin{abstract}

The distribution of relief supplies to shelters is a critical aspect of post-disaster humanitarian logistics. In major disasters, prepositioned supplies often fall short of meeting all demands. We address the problem of planning vehicle routes from a distribution center to shelters while allocating limited relief supplies. To balance efficiency and equity, we formulate a bi-objective problem: minimizing a Gini-index-based measure of inequity in unsatisfied demand for fair distribution and minimizing total travel time for timely delivery. We propose a Mixed Integer Programming (MIP) model and use the $\epsilon$-constraint method to handle the bi-objective nature. By deriving mathematical properties of the optimal solution, we introduce valid inequalities and design an algorithm for optimal delivery allocations given feasible vehicle routes. A branch-and-price (B\&P) algorithm is developed to solve the problem efficiently. Computational tests on realistic datasets from a past earthquake in Van, Turkey, and predicted data for Istanbul's Kartal region show that the B\&P algorithm significantly outperforms commercial MIP solvers.{ Our bi-objective approach reduces aid distribution inequity by 34\% without compromising efficiency. Results indicate that when time constraints are very loose or tight, lexicographic optimization prioritizing demand coverage over fairness is effective. For moderately restrictive time constraints, a balanced approach is essential to avoid inequitable outcomes.}

\end{abstract}

\noindent
{\small
\emph{Keywords: } Humanitarian logistics, Equitable aid distribution, Multi-objective, Branch-and-Price }

\section{Introduction}

After a disaster, affected individuals urgently need core relief items such as drinking water, food, quilts, blankets, sleeping sacks, jackets, tents, lighting equipment, medical goods, and other essential non-food items \citep{UNHCRList}. These items are typically provided at temporary shelters where disaster-impacted individuals find refuge after the catastrophic event. 
In the devastating Maras earthquakes that struck Turkey in February 2023, the impact was profound, resulting in over 54,000 fatalities, more than 100,000 individuals injured, and an additional 2.5 million people in need of shelter across { Southeastern} Turkey and { Northwestern} Syria. During the initial response phase, emergency tents were distributed to those affected, eventually leading to the establishment of tent cities spanning 11 provinces in Southeastern and Eastern Turkey. Over time, the affected population transitioned to temporary housing solutions by establishing container cities with better infrastructure and services. By May 2023, over 400 tent cities and numerous container settlements with a collective capacity of approximately 100,000 containers were accommodating nearly one million displaced individuals.

In large-scale disasters like the Maras earthquakes,  logistical challenges are immense due to limited resources and compromised accessibility during the initial stages of the crisis. Given the impossibility of addressing the needs of all affected individuals simultaneously and the complexity of the operations, developing strategies and solutions for equitable and efficient distribution of aid is vital. Governmental and non-governmental organizations typically store inventory in one or more depots and assign a set of shelters to each depot as part of their disaster preparedness plan. These depots are strategically located based on geographic coverage, accessibility, risk assessment, and proximity to vulnerable populations to ensure efficient and swift disaster response. During the early phase of disaster response, only locally prepositioned supplies are available for distribution. 

In large-scale disasters like the Maras earthquakes, { logistical challenges arise from limited resources and restricted accessibility during the crisis's initial stages. Addressing all affected individuals simultaneously is impossible, making strategies for equitable and efficient aid distribution crucial. Governmental and non-governmental organizations typically store inventory in depots and assign shelters to each as part of disaster preparedness. } These depots are strategically positioned based on geographic coverage, accessibility, risk assessment, and proximity to vulnerable populations to enable swift and efficient responses.   { In the early disaster phase,} only locally prepositioned supplies are available for distribution.  The scarce supplies should be delivered swiftly to alleviate the suffering  \citep{Hu2019}.

The distribution of commodities to the shelters is based on the needs, information about the state of the road network, and the available number of vehicles { together with their capacities,} drivers, and volunteers { gathered} during an early assessment. This raises the question of how to efficiently { allocate and distribute} the relief goods to the shelters. { If demands at all shelters could be fulfilled or  the delivery amount to each shelter is determined at first, then the distribution decisions}  would constitute a classical vehicle routing problem (VRP). { However, optimizing the relief amounts to be delivered to the shelters and the routes of the vehicles simulatenously may provide a better solution. This is the problem we address in our study.} 

{ In the early hours of the disaster, supply is scarce and not all demand can be satisfied.} For instance, in the immediate aftermath of the Maras earthquakes in 2023, provinces such as Hatay and Adıyaman were left with a large amount of unsatisfied demand, whereas other provinces, such as Maras and Gaziantep, were more advantageous. In situations where part of the needs of beneficiaries must remain unfulfilled, ``effectiveness'' or ``efficiency'' cannot be the only goal; equity becomes an essential consideration.


{ Equitable relief distribution remains an open research area in operations research and humanitarian operations. This paper develops and solves a bi-objective optimization model as the basis for a decision support tool to help aid agencies design effective and equitable distribution plans. The model addresses distributing supplies from a single depot by dispatching available vehicles to shelters. It involves (i) determining the delivery amount to each shelter to partially or fully meet its demand and (ii) planning vehicle routes while accounting for road damages and accessibility. } Figure~1  illustrates the structure of a solution by a small example.

{\mm
\begin{figure}[H]
  \includegraphics[width=0.55\textwidth]{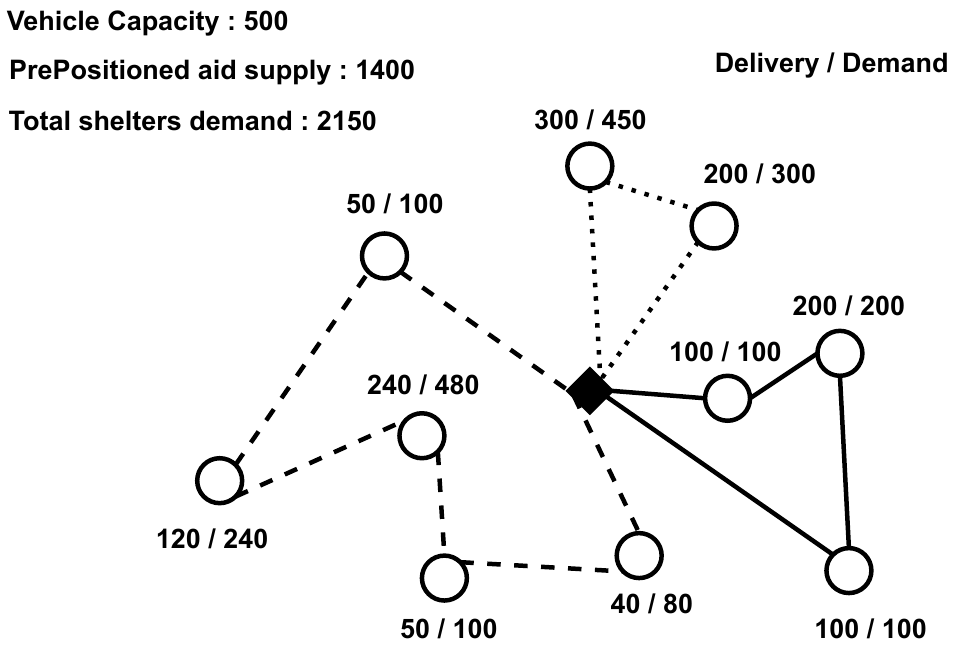}
  \centering
  \caption{ \small Example of three vehicle routes (shown with different line types) delivering to shelters from a single depot. Numbers above each shelter show delivered amounts and corresponding demands, with demand satisfaction varying: 100\% for the two rightmost shelters, but only 50\% for the leftmost.}
  \label{fig:Vanmap}
\end{figure}
\vspace{-0.5em}
}

Within a bi-objective framework for determining Pareto-optimal solutions, we consider two largely conflicting objectives. The first objective aims to maximize the efficiency of the operation by minimizing the overall transportation time of the vehicles. In addition, we also impose a time limit for the completion of each individual vehicle route to achieve a service guarantee. The second objective evaluates {  both the total (or average) unsatisfied demand and the degree of inequity in its distribution among the affected population. }

Evaluation measures for allocation of benefits to the individuals of a population have been proposed under the name of {\it Social Welfare Functions} (SWFs) (see \cite{karsu2015inequity}, \cite{adler2019measuring}, \cite{chen2021guide}). In this paper, we apply a special SWF, defined as a weighted sum of two components: (i) average unsatisfied demand per victim and (ii) Gini's mean absolute difference of the values of unsatisfied demand across all victims. Gini's mean absolute difference builds on the well-known {\it Gini index}\footnote{\mm 
The Gini index is a measure of inequality within a distribution.
A Gini index of 0 represents perfect equality,
while a Gini index of 1 corresponds to maximal inequality.}, arguably the most prominent inequity measure in the economic literature. For studies that use Gini index-related measures in operation research, please refer to Section~\ref{sec:lit_equity}.


The problem presented in this study extends the classical Capacitated Vehicle Routing Problem (CVRP) by determining the delivery amounts at each location, while considering service level constraints and optimizing for two objectives. 
Although our focus is on humanitarian logistics applications, it is worth mentioning that CVRP extensions addressing both efficiency and equity are also relevant in commercial logistics as exemplified by \cite{toth2014vehicle}. 
 \cite{vidal2020concise} highlighted the desirability of incorporating appropriate equity measures to capture service equity in vehicle routing problems.  \cite{matl2017workload} reviewed and analyzed equity measures in vehicle routing models, proposing the Gini index as a promising measure for capturing inequity in VRPs.

{ To our knowledge, no prior research has tackled a bi-objective problem that integrates explicit routing, delivery amount decisions, and equity considerations. More importantly, our proposed problem and algorithm offer practical value, helping humanitarian operations managers allocate scarce resources more effectively and equitably. By optimizing post-disaster resource distribution, we aim to ensure equitable access to essential supplies and support community recovery. Our study contributes to the VRP and humanitarian logistics literature as follows:  
1) We incorporate a social welfare function based on the Gini index to balance effectiveness and equity in relief distribution.  
2) We propose a bi-objective VRP-based MIP formulation, using the $\epsilon$-constraint method to address efficiency and equity trade-offs.  
3) We derive mathematical properties of optimal deliveries. Based on these properties, we introduce two novel valid inequalities to tighten the LP relaxation of the model, significantly accelerating our column generation approach. 
4) We develop an efficient solution approach to address computational challenges of a Gini-index-based objective, including a column generation method and a GRASP heuristic for the pricing problem.  
5) We design a Branch-and-Price (B\&P) algorithm, enhanced with a heuristic to generate  an initial set of columns leading to a feasible solution.  
6) Our computational study uses datasets from Istanbul's Kartal district and Turkey's Van province, offering insights into the trade-offs between objectives and the price of fairness in real-world scenarios.}

The rest of the article is organized as follows. Section~\ref{sec:literature} discusses related works. The formal problem definition, a mathematical model,
the social welfare objective function, as well as properties of the optimal solution are presented in Section~\ref{sec:problem_defination}. In Section~\ref{sec:solution_method}, we introduce our exact solution method.  Section~\ref{sec:computational_test} is devoted to our computational experiments. Finally, we conclude our work and discuss future research directions in Section~\ref{sec:conclution}.

\section{Literature review}\label{sec:literature}

\subsection{Relief aid distribution}

In recent years, {numerous} articles focused on the challenges associated with last-mile disaster relief, with regard to various aspects such as relief aid distribution (\cite{lin2011logistics}, \cite{penna2018vehicle}, \cite{bruni2018fast}, \cite{Sheu2014}), medical/technical crew routing and team deployment (\cite{talarico2015ambulance}, \cite{duque2016network}, \cite{Chen2012}), 
{\mm road restoration post-disaster networks (\cite{duque2016network}, \cite{akbari2021online}, \cite{ajam2022routing})} or victim evacuation from affected areas (\cite{zhu2019emergency}, \cite{molina2018multi}). 
{\mm A considerable portion of these papers develops models with multiple objectives to account for the diverse evaluation criteria used in humanitarian operations. Works such as \cite{lin2011logistics}, \cite{chang2014greedy}, \cite{wang2014multi}, \cite{huang2015modeling}, \cite{molina2018multi}, \cite{ferrer2018multi}, cf.~also the survey~\cite{gutjahr2016multicriteria}, exemplify this approach. For multi-objective vehicle routing in a more general context, we refer to~\cite{jozefowiez2008multi,zajac2021objectives}.} {\mm For comprehensive insights into mathematical models for the humanitarian supply chain and emergency logistics management, we recommend reviewing works such as \cite{kundu2022emergency}, \cite{habib2016mathematical}, \cite{ozdamar2015models} or \cite{anaya2014relief}. 
}

An early paper by \cite{balcik2008last} addressed relief supply allocation and vehicle routing, proposing a two-phase approach to minimize logistics and penalty costs for unmet or delayed demands. Their algorithm first determines delivery schedules, followed by inventory allocation. \cite{nolz2011risk} focused on post-disaster drinking water distribution, aiming to reduce operational risk, water access travel time, and total vehicle travel time. \cite{chang2014greedy} implemented a multi-objective genetic algorithm for relief distribution during Taiwan's Chi-Chi earthquake. \cite{huang2015modeling} studied aid distribution and allocation, emphasizing life-saving utility, fairness, and delay costs. \cite{moreno2018effective} optimized distribution center locations and fleet sizing under uncertainty with a two-stage stochastic model to reduce logistics costs and human suffering. Vehicle routing is a crucial aspect of last-mile delivery, and for articles focusing on the vehicle routing problem in humanitarian operations, readers can refer to the review by \cite{anuar2021vehicle}.

While many studies concentrate on the effective and efficient distribution of relief aid in post-disaster settings, none of the mentioned studies address the equity of distributed aids. This work stands out from previous efforts by considering both the efficiency and equity of distributions.

\subsection{Equity in humanitarian logistics}\label{sec:lit_equity}

{The humanitarian logistics literature increasingly addresses equity (fairness) in post-disaster logistics, particularly concerning emergency response times. \cite{holguin2013appropriate} introduced deprivation cost to quantify human suffering in these contexts. However, \cite{gutjahr2018equity} highlighted that solely minimizing total deprivation costs, although initially expected to promote equitable solutions, was still insufficient to ensure equity. To address this, they proposed a measure based on Gini's mean absolute difference in deprivation costs, an aggregation function (\cite{kostreva2004equitable}) that balances effectiveness and equity. Their case study on the 2015 Nepal earthquake demonstrated significant equity gains with minimal effectiveness losses. Similarly, \cite{alem2022revisiting} used the Gini coefficient to optimize the equity-effectiveness trade-off in location-allocation problems, introducing valid inequalities and a clustering-based Lorenz curve construction to streamline constraints and variables. Their model was tested on data from flood and landslide disasters in Rio de Janeiro state, Brazil.} 

In recent years, the Gini index and Gini's mean absolute difference have been utilized to incorporate fairness considerations in various problems such as relief aid distribution in disaster response (\cite{gutjahr2018equity}, \cite{park2020supply}), humanitarian food distribution (\cite{eisenhandler2019humanitarian}, \cite{eisenhandler2019Segment}), shelter location in disaster preparedness (\cite{mostajabdaveh2018inequity}, \cite{gutjahr2021inequity}), and equitable vaccine distribution to control an epidemic (\cite{enayati2020optimal}, \cite{gutjahr2023fair}). For a recent and comprehensive examination of equity measures in Operations Research studies, including Gini-related measures, readers are referred to \cite{chen2021guide}.

{\mm 
Several papers investigated humanitarian logistics problems 
using other approaches for ensuring equity, such as the maxmin function (Rawlsian approach).
\cite{campbell2008} were among the first to apply equity concepts to disaster relief supply distribution, introducing an equity-related objective defined as minimizing maximum arrival time to demand points. They employed local search heuristics to obtain significantly different solutions compared to conventional objectives. 
Some articles use models representing uncertainty, see, for example, \cite{noyan2015stochastic}, \cite{noyan2018stochastic}, \cite{sabouhi2018robust}, \cite{fianu2018markov}, \cite{cheng2021modeling}, and \cite{li2024distributional}. Other works examine the relief item distribution problem in a multi-period setting; see \cite{yu2018novel}, \cite{huang2019equitable}, \cite{ccankaya2019humanitarian}, and \cite{ekici2020inventory}.
We refer the reader to the review by \cite{donmez2022fairness} for fairness in humanitarian logistics operations.}
{\mm Our work differs from the studies mentioned above by specifically addressing the problem of aid distribution, and by using a Gini-index-related metric.}

{In order to locate our paper within the existing literature, we can use the classification in the recent survey by \cite{vidal2020concise}. Therein, regarding models with equity objectives, three types of papers are distinguished:
papers dealing with (i) workload balance, (ii) service equity, or (iii) collaborative planning.
Our current work addresses the second of these types, service equity. In this category, \cite{vidal2020concise} indicate only two publications, namely \cite{balcik2010review} and \cite{huang2012models}. 
}
{The survey by \cite{balcik2010review} is close to our work as it focuses on equity considerations in vehicle routing. The Gini index, the main metric concept of the current paper, is already mentioned. Moreover, the authors already refer to supplies as equity determinants, as opposed to arrival times or similar measures which prevail in the most other articles on equitable routing. However, no publication dealing with equitable supply by a Gini-based approach is referred to in the survey.

\cite{huang2012models} studied equitable aid distribution, but defined equity in terms of standard deviation and of maximum pairwise difference between individual service levels, where the service level is determined for each aid recipient from delivery speed and delivery amount. 
More recently, \cite{ghasbeh2022equitable} focused on maximizing equity and minimizing response time in post-disaster relief aid distribution, where equity is addressed by minimizing a weighted sum of unmet demand for non-emergency commodities. Both \cite{huang2012models} and  \cite{ghasbeh2022equitable} use equity concepts that are different from our approach as they do not rely on the Gini index.}

{\cite{eisenhandler2019humanitarian} introduced the humanitarian pickup and distribution problem for food donation logistics, addressing site selection, routing, and allocation under capacity and working-hour constraints. The objective function combines total delivered units with a Gini-based equity measure. A heuristic method was tested on data from food banks in Israel and Houston, Texas. \cite{eisenhandler2019Segment} extended this work with a segment-based formulation and a new heuristic to handle complex scenarios, including multi-vehicle cases and time windows. Our relief distribution problem shares similarities with these studies but differs in its use of a more general equity measure, its application context, structure, and solution method. Specifically, we address multi-vehicle cases with a bi-objective formulation for equity and travel time, assuming all demand points require service. Importantly, we compute exact solutions, tackling the challenge noted in \cite{eisenhandler2019Segment}, where the authors state that solving their model by an approach using column generation and branch-and-price would be a complex task.}

\cite{avishan2023humanitarian} address the problem of routing relief logistics teams after a major disaster. Their model decides on the service time of teams at each location and considers the equity of the service time received by affected sites. Equity is indirectly managed through a utility function, where utility decreases each time unit when the service time exceeds a site's critical need. The study utilizes adjustable robust optimization to address the uncertainty in the travel times of relief teams. To solve the problem efficiently, a heuristic approach is proposed. There are notable differences with our model.
We address the frequent case where relief supplies fall short of demand. Contrary to the approach based on service time, our study explicitly calculates the amount of relief supplies delivered to each location. Additionally, we directly consider equity.
Furthermore, instead of relying on heuristics, we employ an efficient exact algorithm to solve the problem.


{\mm In summary, the literature on relief aid distribution that combines vehicle routing decisions with decisions on deliveries while considering equity remains relatively limited. Here, we introduce a novel bi-objective optimization model for determining a distribution plan that dictates vehicle routes and delivery quantities to shelters. This model aims to minimize two objectives: (i) total traveling time and (ii) inequity of relief supply distribution. Notably, none of the previously mentioned studies have presented an exact solution method that efficiently addresses problems of similar structure for real-world application sizes. Our problem focuses on the post-disaster response stage, necessitating swift identification of solutions. We believe that our developed efficient branch-and-price method for equitable relief item routing constitutes a significant contribution in terms of both practical utility and methodological advancement.} {Table 1 surveys related publications in comparison to our work.}
\begin{table}[ht]
\Tablesize
\centering
\begin{tabular}{@{}l|c|c|c|c|c@{}}
\toprule
\multicolumn{1}{c|}{\textbf{Study}} & \textbf{\begin{tabular}[c]{@{}c@{}}Multi-\\ Objective\end{tabular}} & \textbf{\begin{tabular}[c]{@{}c@{}}Equity \\ Measure\end{tabular}} & \textbf{\begin{tabular}[c]{@{}c@{}}Routing \\ Aspects\end{tabular}} & \textbf{\begin{tabular}[c]{@{}c@{}}Decision \\ on \\ Delivery \\ Quantities\end{tabular}} & \textbf{Method} \\ \midrule
Campbell et al. (2008) & No & \begin{tabular}[c]{@{}c@{}}MinMax \\ Arrival Time\end{tabular} & \begin{tabular}[c]{@{}c@{}}VRP/TSP \\ with \\ vehicle capacity\end{tabular} & No & \begin{tabular}[c]{@{}c@{}}Local search\\ with insertion/\\ improvement\end{tabular} \\ \midrule
Huang et al. (2012) & No & \begin{tabular}[c]{@{}c@{}}Min \\ Service level \\ differences\end{tabular} & \begin{tabular}[c]{@{}c@{}}Vehicle-to-route\\ assignment\end{tabular} & Yes & CPLEX / GRASP \\ \midrule
Huang et al. (2015) & \begin{tabular}[c]{@{}c@{}}Yes, \\ Max lifesaving utility, \\ Min delay cost, \\ Min variance \\ of demand fill rates\end{tabular} & \begin{tabular}[c]{@{}c@{}}Demand \\ fill rate\\ variance\end{tabular} & Network flow & Yes & \begin{tabular}[c]{@{}c@{}}Convex quadratic \\ network flow, \\ VI algorithm\end{tabular} \\ \midrule
Moreno et al. (2018) & \begin{tabular}[c]{@{}c@{}}Yes, \\ Min deprivation \\ costs,\\ Min logistics costs\end{tabular} & \begin{tabular}[c]{@{}c@{}}Deprivation\\ costs\end{tabular} & \begin{tabular}[c]{@{}c@{}}Vehicle-to-node \\ assignment\end{tabular} & Yes & \begin{tabular}[c]{@{}c@{}}Two-stage\\ stochastic model, \\ heuristics\end{tabular} \\ \midrule
\begin{tabular}[c]{@{}l@{}}Eisenhandler\\ and Tzur (2019a)\end{tabular} & \begin{tabular}[c]{@{}c@{}}Yes, \\ Max food allocation, \\ Max equity of food \\ distribution\end{tabular} & \begin{tabular}[c]{@{}c@{}}Gini-related \\ measure\end{tabular} & \begin{tabular}[c]{@{}c@{}}Single vehicle\\ Pickup and \\ delivery, \\ Vehicle capacity\end{tabular} & Yes & \begin{tabular}[c]{@{}c@{}}Large \\ neighborhood \\ search\end{tabular} \\ \midrule
\begin{tabular}[c]{@{}l@{}}Eisenhandler \\ and Tzur (2019b)\end{tabular} & \begin{tabular}[c]{@{}c@{}}Yes, \\ Max food \\ allocation, \\ Max equity of \\ food distribution\end{tabular} & \begin{tabular}[c]{@{}c@{}}Gini-related \\ measure\end{tabular} & \begin{tabular}[c]{@{}c@{}}Pickup and \\ delivery, \\ Vehicle capacity, \\ Time windows\end{tabular} & Yes & Matheuristic \\ \midrule
Ghasbeh et al. (2022) & \begin{tabular}[c]{@{}c@{}}Yes, \\ Min unmet demand, \\ Min response time\end{tabular} & \begin{tabular}[c]{@{}c@{}}Weighted \\ unmet \\ demand\end{tabular} & \begin{tabular}[c]{@{}c@{}}Location-\\ routing\end{tabular} & Yes & \begin{tabular}[c]{@{}c@{}}ALNS and \\ local search\end{tabular} \\ \midrule
Avishan et al. (2023) & \begin{tabular}[c]{@{}c@{}}Yes, \\ Min unmet demand, \\ Min max service \\ time difference\end{tabular} & \begin{tabular}[c]{@{}c@{}}Utility of \\ service \\ time\end{tabular} & \begin{tabular}[c]{@{}c@{}}Pickup and \\ delivery, \\ Time windows\end{tabular} & No & \begin{tabular}[c]{@{}c@{}}Robust \\ optimization, \\ Heuristic\end{tabular} \\ \midrule
\multicolumn{1}{c|}{Ours} & \begin{tabular}[c]{@{}c@{}}Yes, \\ Min total travel time, \\ Min inequity \\ of relief supply \\ distribution\end{tabular} & \begin{tabular}[c]{@{}c@{}}Gini-related\\ measure\end{tabular} & \begin{tabular}[c]{@{}c@{}}VRP with \\ vehicle capacity\\ and \\ travel time limit\end{tabular} & Yes & \begin{tabular}[c]{@{}c@{}}Branch-\\ and-Price\end{tabular} \\ \bottomrule
\end{tabular}
\caption{\centering Summary of objectives, equity measures, routing decisions, and methodologies in relief aid distribution studies considering equity}
\end{table}

\section{Problem definition and mathematical models}\label{sec:problem_defination}
For the response phase after a disaster, we investigate the problem of planning vehicle routes between a depot and shelter locations while simultaneously assigning relief supplies to the shelters. {A depot is a distribution center where relief items are pre-positioned in preparation for a disaster.} {\mm Shelters are assumed to receive supplies from a single depot, as it is conventional in practice for a shelter to be assigned to a single distribution center.}

The problem is defined on {a} directed graph $G = (V, A)$. The node set $V = \{0, . . . , n,n+1\}$ contains the depot that is represented by both node 0 and node $n+1$, and $n$ shelters. We denote the set of all shelters as $V_c = V \setminus \{0,n+1\}$. {\wg The set $A$ is defined as $\{(i, j) \, | \, i,j \in V,\, i \neq j, \, i\neq n+1,\, j\neq 0\}$.} To each edge~$(i,j)$, a travel time $t_{i,j}$ is assigned, which is calculated as the time for traversing the shortest path between $i$ and $j$ in the post-disaster road network. {\mm We assume that the necessary information on the road network, such as information on time required for road restoration and traffic condition, has already been collected during an initial assessment phase and has been taken into account in the computation of the numbers~$t_{i,j}$.}\footnote{It is worth mentioning that in practice, the values of the travel times $t_{i,j}$ can be time-dependent in the aftermath of a disaster. For example, some damaged roads may be repaired, which reduces the values of $t_{ij}$ (see, e.g., \cite{moreno2019branch}, \cite{moreno2020heterogeneous}). In the present work, we consider a small time horizon, such that changes of travel times do not play a relevant role, but future work should address the situation of longer time horizons where the assumption of constant travel times has to be relaxed.
}
Travel times are not necessarily symmetric, so $t_{i,j}$ needs not be equal to~$t_{j,i}$.

{\mm Note that by the assumption above, the given road network graph has been extended to an (almost) complete graph~$G$.
While traversing edge~$(i,j)$ of~$G$, i.e., while traveling from node $i$ to node $j$, a route may physically pass through other nodes that lie on the shortest path between $i$ and $j$. This does not imply that the shelters located at these intermediate nodes are visited.}
{
For example, suppose the physical network consists of three nodes, a depot and two shelters indexed 1 and 2, that lie on one and the same straight road in the indicated order. Then a vehicle may start at the depot, visit shelter~1, visit shelter~2, and return then to the depot while passing through node 1 once again. In the graph~$G$, where the depot is indexed as node 0 or node 3, this second passage does not count as a visit to shelter~1, since in~$G$, the backward trip only corresponds to a traversal of edge (2,3), but neither of (2,1) nor of (1,3).
We would also like to note that the triangle inequality might not hold in the original road network, but it holds in the complete graph~G by definition of $t_{i,j}$, since if $t_{i,j}$ would be larger than $t_{i,k}+t_{k,j}$, then the path defining $t_{i,j}$ would not be a shortest path.
}


Each shelter $i$ requires an amount of relief supply, $d_i$, which is calculated {by an} assessment of the number of victims who need relief items at shelter location~$i$. We are concerned with a single commodity, {or equivalently, a package combining several commodities with fixed proportions that are composed} according to the common needs of victims. 
We assume that only the prepositioned relief supply at the depot, $C$, is available for distribution among victims. {Furthermore,} we consider the (frequent) situation where the prepositioned supply $C$ is less than the total demand of shelters: $C < \sum_{i \in V_c} d_i$, {which is the challenging case from the viewpoint of equity}.

A fleet {of $m$ homogeneous vehicles}, each with a capacity of $Q$, is based at the depot. Each vehicle performs only one route that starts from the depot and visits some of the shelters on the post-disaster network. The total duration of each route {has to} be less than $\Psi$. {Each shelter is allowed to} be visited only once for the purpose of delivering goods; however, {it} can act as {an} intermediate node
{on a route}. We do not allow split delivery, 
{i.e., the demand of a shelter is (partially or completely) covered by at most one vehicle.} 

We address two objectives for the problem, time efficiency and unsatisfied demand. To represent time efficiency, we minimize the total relief response time, which is the sum of all vehicles' traveling times. As for the second objective, we combine the average unsatisfied demand per victim with an equity measure: Gini's mean absolute difference of unsatisfied demand ratios between individuals. A detailed explanation of this measure can be found in Section~\ref{sec:equityMeasure}. Table \ref{Model:PAR} presents the notation {for} the problem's parameters. In the following sections, we introduce a compact formulation of the problem.

\begin{table}[h]
\centering
\small
\caption{Parameters}
\label{Model:PAR}
\begin{tabular}{@{}l|l@{}}
\toprule
Parameter & Definition \\ \midrule
  $Q$     & Vehicle capacity \\ 
  $d_i$   & Demand for relief supplies at shelter node $i$ \\
  $D$     & {Total demand, $D = \sum_i d_i$} \\
  $t_{i,j}$& Time it takes to travel between nodes $i$ and $j$ \\
  $C$ 	  & Total amount of available relief supply \\
  $\Psi$  & Maximum delivery time (time when the last supply {has to be} delivered) \\
  $m$     & {Number of vehicles} \\       
  \bottomrule
\end{tabular}
\end{table}


\subsection{Mathematical model}\label{sec:P2Math}
To formulate our problem, we first follow the vehicle flow formulation for the {VRP with maximum travel time duration}. This formulation is easy to understand, implement, and adapt to additional assumptions. We refer to this formulation as the VF model in this work. The decision variables are as defined in Table \ref{Table:VFvariables}. 

\begin{table}[!htbp]
\small
\centering
\caption{Decision Variables}
\label{Table:VFvariables}
\begin{tabular}{@{}llll@{}}
\toprule
Notation & Description & Variable type & Domain\\ \midrule
$x_{i,j}$&\begin{tabular}[c]{@{}c@{}} 1, if the vehicle directly travels from node $i$ to node $j$
\end{tabular}  & binary & $\forall i,j \in V$ \\
$v_{i}$  & The amount of relief supply to node $i$ & non-negative & $\forall i \in V_c$ \\ 
$u_{i} $ & The amount of {delivered} vehicle load when reaching node $i$ & non-negative & $\forall i \in V_c$ \\
$w_{i}$  & The time that the vehicle visits node $i$, $u_0 = 0$ & non-negative &$\forall i \in V$ \\ \bottomrule
\end{tabular}
\end{table}
\begin{align}
\allowdisplaybreaks
\text{objective 1: }\min \: & z_1 = \sum_{(i,j) \in A}t_{i,j}{x_{i,j}}  \\
\text{objective 2: } \min \: & z_2 = D \cdot {\cal I}(v) \label{Obj:VFequity}\\
\mbox{s.t. } 
& \sum\limits_{j \in V  \setminus \{0\}}{x_{i,j}} = 1,  \quad \forall i \in V_c, \label{Con:VFsinglevisit}\\
& \sum\limits_{j \in V \setminus \{n+1\} } x_{j,i}  = 1,  \qquad i \in V_c, \label{Con:VFbalance}\\
& \sum\limits_{j \in V \setminus \{0\}} x_{0,j} \le m, \label{Con:VFdepotout}\\
& \sum\limits_{i \in V \setminus \{n+1\}} x_{i,n+1} \le m, \label{Con:VFdepotin}\\
& v_i \leq d_i, \qquad \forall i \in V_c, \label{Con:VFdelivery} \\
& \sum\limits_{i \in V_c} v_i\leq C, \label{Con:VFSupply}\\
& w_j \geq w_i + t_{i,j} - (1-x_{i,j})M, \qquad \forall i,j \in V, \label{Con:VF_MTZ1} \\
& w_j \leq \Psi, \qquad  \forall j \in V_c  \label{Con:VFMaxroute} \\
& u_j \geq u_i + v_i - (1-x_{i,j}) M, \qquad \forall i,j \in V_c,  \label{Con:VF_MTZ2}  \\
& u_j \leq Q, \qquad  \forall j \in V_c \label{Con:VFcap}\\
& x_{i,j} \in \{0,1\}, \quad \forall (i,j) \in A, \nonumber\\
& v_i, u_i, w_i \geq 0 \qquad  \forall i \in V_c \label{Con:VFvariables}
\end{align}

{The model starts with the two objective functions of our bi-objective problem, where objective~1 represents total
traveling time, while objective~2 is the objective function expressing inequity of demand coverage. The inequity measure ${\cal I} = {\cal I}(v)$ will be defined and explained in Section~\ref{sec:equityMeasure}.}
Constraints (\ref{Con:VFsinglevisit}) ensure that each shelter will be visited once. Constraints (\ref{Con:VFbalance}) impose {\mm
ensure that a vehicle leaves a shelter exactly once. Constraints (\ref{Con:VFdepotout}) and (\ref{Con:VFdepotin}) make sure that not more than~$m$ vehicles are in use.} Constraints (\ref{Con:VFdelivery}) and (\ref{Con:VFSupply}) guarantee that the delivered amount to 
{a shelter does not exceed the shelter's demand, and that the sum of the delivered amounts is less or equal to} 
the total available relief supply at the depot, respectively. Constraints (\ref{Con:VF_MTZ1}) calculate the visiting time of each shelter $i$, represented by variable $u_i$.  Constraints (\ref{Con:VFMaxroute}) make sure that all shelters are visited before {time} $\Psi$ given that vehicles start at time~0. Constraints (\ref{Con:VF_MTZ2}) calculate the vehicle load at the time of visiting shelter $i$, and Constraints (\ref{Con:VFcap}) guarantee that the vehicle load does not exceed the vehicle capacity. Finally, Constraints (\ref{Con:VFvariables}) define the variables' domains. 


\subsection{Representation of inequity}\label{sec:equityMeasure}
{Unsatisfied demand in at least one shelter is an inevitable consequence of the situation considered in this paper, that of insufficient overall supply.
A straightforward approach (called ``utilitarian'' in the literature) would simply minimize the total amount of unsatisfied demand. However, it is well-known that this approach often leads to very unbalanced distributions of the relief commodities, privileging parts of the population and grossly discriminating against other parts. Thus,}
in a situation of scarce resources, balancing the amount of relief supply delivered to shelter locations in a fair way plays a crucial role for achieving an allocation of supply that is ethically tenable and well-accepted by the population. Therefore, we do not restrict our 
second, coverage-related objective function to an evaluation of total coverage, but complement it by a second term: an equity measure expressing the degree to which the unsatisfied demand in the different shelters is fairly distributed. 
{The well-known {\em Gini index} seems to be the most popular equity measure in the economic literature, so we have decided to build on this index.}

In this way, we get two components for our second objective function:
(i) 
an effectiveness term, representing the total (or, equivalently, average) unsatisfied demand 
and
(ii)
an inequity term, quantified by means of the Gini index concept, which judges the extent of inequity of the unsatisfied demand values across the population.
The two terms are aggregated in the form of a weighted sum. This produces a social welfare function belonging to a class called {\em inequity-averse aggregation functions} {(IAAFs)} in~\cite{karsu2015inequity}.
For previous applications of the same measure, see \cite{gutjahr2018equity}, \cite{mostajabdaveh2018inequity} or \cite{gutjahr2023fair}. Concerning a motivation of the way how our IAAF is constructed, we restrict us here to the remark that a weighted sum is obviously the most natural approach to combine the two basic terms. 
{This approach has a long tradition in the literature:
readers interested in an axiomatic justification of the chosen IAAF are referred to \cite{porath1994linear} and \cite{ARGYRIS2022560}. Moreover, the weighted sum formula can also be derived from game-theoretic models, as shown in \cite{schmidt2019inequity}.
}

{\wg In a minimization problem concerning $N$ individual persons, indexed $\nu = 1,\ldots,N$,
let $c^{(\nu)}$ represent the cost or any alternative form of disutility encountered by individual $\nu$. In this study, $c^{(\nu)}$ quantifies the unsatisfied demand (detailed further below). We define our IAAF as ${\cal I} = \mu + \lambda \cdot \Delta$, where $\mu$ denotes the average cost, calculated as $\mu = \frac{1}{N} \sum_{\nu=1}^N c^{(\nu)}$, and $\Delta$ represents Gini's mean absolute difference, defined as $\Delta = \frac{1}{N^2} \sum^N\limits_{\nu=1} \sum^N\limits_{\ell=1}|c^{(\nu)}-c^{(\ell)}|$, to measure the dispersion of individual costs.}
{\wg It is well-known that $\Delta= 2 \mu {\cal G}$, where ${\cal G}$ denotes the Gini index. 
To use Gini's mean absolute difference~$\Delta$ instead of the Gini index~${\cal G}$ as the second term has the advantages that $\Delta$ can be linearized, and that $\Delta$ is expressed in the same units as~$\mu$, such that the weight factor~$\lambda$ is a dimension-less constant.

For $0 \leq \lambda \leq 1/2$, the measure ${\cal I}$ is guaranteed to be monotonous in the individual costs $c^{(\nu)}$ (this is Axiom~4 in~ \cite{porath1994linear}), which excludes implausible solutions where the value of ${\cal I}$ could be improved by increasing the costs of one or several individuals while keeping the costs of the other individuals unchanged. In our computational results, we choose $\lambda = 1/2$.

In our study, we assume that the demand of each beneficiary is one unit, that is, $d_i$  represents the number of beneficiaries at shelter~$i$. We define the unsatisfied demand {of an} individual~$\nu$ at shelter~$i$ as $c^{(\nu)}=1-v_i/d_i$, assuming equal distribution {\em within} each shelter. In particular, $c^{(\nu)}$ is identical for all individuals~$\nu$ at shelter~$i$ and can therefore be written as $c_i$ $(i \in V_c)$.  Note that this term is always non-negative since the delivery amount~$v_i$ to shelter $i$ is less {or equal to} the demand $d_i$. Moreover, the total demand $D = \sum_id_i$ is identical to the number~$N$ of beneficiaries.

The quantities $\mu$ and $\Delta$ can be immediately calculated as follows. 
\begin{equation}\label{def:mu}
\mu=\frac{1}{D} \sum_{i \in V_c} d_i \, c_i =\frac{1}{D} \sum_{i \in V_c} d_i \cdot (1-v_i/d_i) =
\frac{1}{D}\sum\limits_{i \in V_c}(d_i-v_i)=1-\frac{1}{D}\sum\limits_{i \in V_c} v_i,
\end{equation}
\begin{equation}\label{def:Delta}
\Delta=\frac{1}{D^2} \sum\limits_{i \in V_c} \sum\limits_{j \in V_c} d_i  d_j \, |c_i - c_j| =
\frac{1}{D^2} \sum\limits_{i \in V_c} \sum\limits_{j \in V_c} d_i  d_j \, |v_j/d_j - v_i/d_i| =
\frac{1}{D^2}\sum\limits_{i \in V_c} \sum\limits_{j \in V_c} |d_iv_j-d_j v_i|.
\end{equation}
}
Please note that our inequity measure does not refer to inequity {in demand satisfaction ratios} between {\em shelters}, but between {\em individuals}.\footnote{This is not the same: Imagine the situation where there are three shelters with 10, 100 and 1000 individuals, respectively. Consider variant A where shelters~2 and~3 have a demand satisfaction ratio of $v_i/d_i = 0.5$ while shelter~1 has only a ratio of $v_1/d_1 = 0.4$, and compare it to variant B where shelters~1 and~2 have a demand satisfaction ratio of 0.5 while shelter~3 has only 0.4. Then the inequity between the three demand satisfaction ratios of the shelters is the same in both variants, but one would probably prefer variant A where the number of {\em individuals} getting a share deviating from 50 \% is much smaller than in variant~B. Our formula~(\ref{def:Delta}) is able to represent this preference since it weighs the absolute difference between the shelter-related ratios $v_i/d_i$ and $v_j/d_j$ by~$d_i d_j$, the number of pairs of individuals affected by the difference.}
We multiply $\cal I$ by {the constant} $D$ and adopt the resulting expression as the inequity objective function in our problem. 
\begin{align} \label{P2:Giniobj}
D \cdot {\cal I} = \sum\limits_{i\in V_c} (d_i-v_i)+ \frac{\lambda}{D}\sum\limits_{i \in V_c}\sum\limits_{j \in V_c}|d_iv_j-d_jv_i|.
\end{align}

{\mm
To linearize the non-linear term in the objective function, we introduce $2 \cdot |V_c|^2$ new continuous variables alongside two sets of corresponding constraints. Specifically, we define auxiliary variables $\tau^+_{i,j}$ and $\tau^-_{i,j}$, where $\tau^+_{i,j}=\max\{0, (d_i v_j-d_j v_i)\}$ and $\tau^-_{i,j}=\max\{0, -(d_i v_j-d_j v_i)\}$ for every pair of $i,j \in V_c$. These variables allow us to express the objective function~(\ref{P2:Giniobj}) as follows:
\[
\sum_{i\in V_c} (d_i-v_i)+ \frac{\lambda}{D}\sum_{i \in V_c}\sum_{j \in V_c} ( \tau^+_{i,j}+\tau^-_{i,j} ),
\]
where
\begin{align}
\tau^+_{i,j} - \tau^-_{i,j} &= d_iv_j-d_jv_i, &\forall i,j \in V_c, \label{linearization_1}\\
\tau^+_{i,j}, \tau^-_{i,j} &\ge 0, &\forall i,j \in V_c.
\end{align}

Note that in view of constraint (\ref{linearization_1}) and the fact that both $\tau^+_{i,j}$ and $\tau^-_{i,j}$ have positive coefficients in our minimization objective, at most one of these variables will assume a positive value in the optimal solution.
}
\subsection{Characteristics of the optimal solution}\label{sec:char}

A feasible solution $X$ to our problem consists of two components: (i) a set of $m$ vehicle routes $K$ that meet the maximum route length constraint, and (ii) a specification of delivery quantities~$v_i$ to the shelter locations that meet the vehicle capacity constraint and the limitation imposed by the amount of prepositioned relief items. Our first objective function (total traveling time) is already determined by the first component, while the second objective function (the IAAF of the uncovered demand) is exclusively determined by the second component. 
{In this section, we assume the vehicle routes in solution $X$ are fixed, resulting in a constant total travel time. Our focus is on optimizing delivery quantities $v_i$ according to the second objective function~(\ref{P2:Giniobj}).}
The set of all nodes that are visited by route~$k$ will be denoted by~$V_k$ 
$(k \in K)$. Then we have the problem
\begin{align} \label{Giniobj}
\min \quad & D \cdot {\cal I} = \sum\limits_{i\in V_c} (d_i-v_i)+ \frac{\lambda}{D}\sum\limits_{i \in V_c}\sum\limits_{j \in V_c}|d_iv_j-d_jv_i|\\
& \sum_{i \in V_k} v_i \le Q \quad \forall k \in K \label{vcon1}\\ 
& \sum_{i \in V_c} v_i \le C  \label{vcon2}\\
& 0 \le v_i \le d_i \quad \forall i \in V_c \label{vcon3}
\end{align}
For each route $k \in K$, set $D_k = \sum_{i \in V_k} d_i$ and introduce variables $\xi_k = \sum_{i \in V_k} v_i$. 
\vspace{1ex}

The proofs of the following results can be found in the Online Supplement.
\vspace{1ex}

\noindent
{\bf Lemma 1.}
{\em There is an optimal solution $v = (v_i)$ of (\ref{Giniobj})~--~(\ref{vcon3}) for which within each set $V_k$, the $v_i$ are distributed demand-proportional, i.e.~as
$v_i = \xi_k \cdot \frac{d_i}{D_k}$ $(i \in V_k, \, k \in K)$.}
\vspace{0.5ex}


By Lemma~1, when looking for optimal solutions, we can restrict ourselves to solutions in which deliveries $v_i$ within each $V_k$ are given by $v_i = \xi_k \cdot \frac{d_i}{D_k}$ $(i \in V_k, \, k \in K)$. Insertion into (\ref{Giniobj})~--~(\ref{vcon3}) yields
the following optimization problem in the non-negative variables $\xi_k$:
\begin{equation}\label{xObj}
\min_\xi \; \psi(\xi) = D \, {\cal I} = D - \sum_k \xi_k + \frac{\lambda}{D} \, \sum_{k \in K} \sum_{\ell \in K}  
| D_\ell \xi_k - D_k \xi_\ell|.
\end{equation}
\begin{equation}\label{xCon}
\mbox{s.t.} \; \: 0 \le \xi_k \le \min (Q, D_k) \; \: \forall k, \quad \sum_{k \in K} \xi_k \le C.
\end{equation}

The following two propositions will allow a solution of (\ref{xObj})~--~(\ref{xCon}) by simple arithmetic calculations: 
\begin{prop}\label{prop:opt_1}
If $\frac{D_k}{D} \le \frac{Q}{C}$ for all $k \in K$, then the optimal solution of (\ref{xObj})~--~(\ref{xCon}) is given by
\begin{equation}\label{xiprop1}
\xi_k = \frac{C}{D} \cdot D_k \quad (k \in K),
\end{equation}
which constitutes perfect equity.
\end{prop}


\begin{prop}\label{prop:opt_2new}
If $\frac{D_k}{D} > \frac{Q}{C}$ for some $k \in K$, then in an optimal solution $\xi = (\xi_1,\ldots,\xi_m)$ {\mm of (\ref{xObj})~--~(\ref{xCon})},
it holds that $\xi_k = Q$.
\end{prop}


Note that taken together, Propositions \ref{prop:opt_1} and \ref{prop:opt_2new} cover all possible cases. Therefore, we can apply them alternatively in a step-by-step manner, starting with a feasible solution for fixed $x_{i,j}$ variables, going through the sets $V_k$, and computing the optimal values for the $v_i$ variables. This is detailed in the Algorithm~\ref{Alg:Optimal_quantity} 
presented in the following.

\begin{algorithm}[h]
\small
\caption{Optimal delivery quantities' calculation procedure}
\label{Alg:Optimal_quantity}
\begin{algorithmic}
\STATE Input: Feasible solution X
\STATE Output: The optimal $v_i$ given the fixed $x_{i,j}$ 
\STATE Set $\bar{K} = K$, $K^{fixed} = \emptyset$
\WHILE {$\bar{K} \neq \emptyset$}
\STATE $K^{vio} = \emptyset$
\STATE Set $D' = \sum_{k \in \bar{K}} D_k$
\STATE Set $C' = \min \{ C - \sum_{k \in K^{fixed}} \sum_{i \in V_k} v_i, \: D' \}$
\FOR {$k \in \bar{K}$}
\IF {$D_k/D' > Q/C'$}
\STATE $K^{vio} =  K^{vio} \cup \{k\}$
\ENDIF
\ENDFOR
\IF {$K^{vio} == \emptyset$}
\STATE $v_i =  (C'/D') \cdot d_i, \quad \forall i \in V_k, \forall k \in \bar{K}$
\STATE $\bar{K} = \emptyset$
\ELSE
\STATE $v_i = (Q/D_k) \cdot d_i, \quad \forall i \in V_k, \forall k \in K^{vio}$
\STATE $\bar{K} =  \bar{K} \setminus K^{vio}$ 
\STATE $K^{fixed} =  K^{fixed} \cup K^{vio}$
\ENDIF
\ENDWHILE
\end{algorithmic}
\end{algorithm}

The algorithm determines the set~$K^{vio}$ of that $ k \in K$ to which Prop.~\ref{prop:opt_2new} is applicable, and sets the total delivery to the corresponding sets of nodes equal to $Q$. If there should not be such a~$k$, then Prop.~\ref{prop:opt_1} can be applied; otherwise, the instance can be reduced by removing the sets corresponding to the $k \in K^{vio}$, reducing $D$ and $C$ correspondingly, and repeating the search. 
\vspace{0.5ex}
\noindent
{\bf Corollary 1.}
{\em
For a given vehicle route $k$, in the optimal solution {\mm of (\ref{xObj})~--~(\ref{xCon})}, only one of the two equations $\xi_k = Q$ or $\frac{C}{D} D_k \leq \xi_k \leq D_k$ is valid for the total delivery amount $\xi_k$.
}
\vspace{1ex}

We enhance the solution method by leveraging the above results. Corollary 1 is utilized to generate valid inequalities to enhance the lower bound provided by the Dantzing-Wolfe decomposition. Algorithm~\ref{Alg:Optimal_quantity} calculates the optimal supply distribution in our heuristics approaches. With the result of Lemma~1, we enforce demand-proportional deliveries in the set partitioning formulation of the problem, leading to a reduction in the number of columns.

\section{Exact solution method}\label{sec:solution_method}
{The model presented in Section \ref{sec:P2Math} can be solved using a MIP solver. However, the computation times for this approach are prohibitive for medium and large instances commonly encountered in practice. To address this, we introduce a Branch-and-Price (B\&P) solution method. Our solution approach incorporates several problem-specific components and improvements to tackle the unique challenges of equitable relief supply distribution. First, in Section~\ref{sec:SPformulation}, we introduce route-deliveries as columns within the column generation framework, enabling the simultaneous optimization of vehicle routes and delivery quantities. Unlike most VRP literature, our pricing problem cannot be reduced to a resource-constrained shortest path problem due to the specific characteristics of our model, including the linearization constraints of the Gini absolute difference objective, the total travel time constraint imposed by the $\epsilon$-constraint method, and the delivery quantity decisions. To overcome this, in Section~\ref{sec:Subproblem}, we develop a novel MIP model for the pricing problem and design a customized GRASP algorithm (detailed in the Online Supplement) for efficient optimization.

Additionally, in Section~\ref{sec:Subproblem}, we propose two valid inequalities to accelerate the solution of the pricing problem, derived from optimal solution properties presented in two propositions discussed in Section 3.3. Finally, in Section~\ref{sec:InitialHeuristic}, we develop an initial column generation heuristic based on a Tabu Search algorithm, tailored to respect branching decisions at each node of the branch-and-bound tree. This heuristic leverages our optimal delivery quantity allocation algorithm (Algorithm 1) to generate high-quality solutions within reasonable computational time, ensuring the exclusion or inclusion of specific edges as required.}

\subsection{Multi-objective approach}
To determine the set of Pareto-optimal solutions, we use the $\epsilon$-constraint method, which transforms the bi-objective problem into a series of single-objective problems by introducing an additional constraint. Specifically, we adopt the augmented $\epsilon$-constraint method proposed by \cite{mavrotas2013improved}, which enhances the main objective function by subtracting the normalized slack variable of the added constraint. The augmented $\epsilon$-constraint method transforms the problem to
\begin{equation}\label{exampleEpsilon}
\min\{f_1(x) - \gamma \cdot \frac{s}{\theta}: \:  f_2(x) +s = \epsilon, \: x \in {\cal X} \}    
\end{equation}
where $\gamma$ is a {small machine precision constant}, 
$\theta=f^{max}_2-f^{min}_2$ is the range of $f_2(x)$ values, and $s \ge 0$ is a nonnegative slack.

{\wg To adopt this approach, we choose our inequity measure as the main objective function and use the total traveling time {objective}
{for the} $\epsilon$-constraint, {i.e., we interchange the two objective functions in Subsection~\ref{sec:P2Math}, such that now $f_1 = z_2$ and $f_2 = z_1$.} The reason for the interchange is that it is computationally easier to handle the more complicated fairness-related objective function under formally simple constraints than, vice versa, a formally simple objective function under more complicated constraints. Note that this has only computational relevance and does not have any impact on the mutual rank of the two objective functions.

The optimal (i.e., minimal) value of the problem considering only the total travelling time objective yields $f^{min}_2$. To save runtime, we combine the augmented $\epsilon$-constraint method with the adaptive $\epsilon$-constraint approach: We start with the computation of the minimal solution~$x^*_1$ of the problem with $f_1$ as the objective function and set $f^{max}_2$ and $\epsilon$ equal to $f_2(x^*_1)$. To find the Pareto front, we then iteratively decrease the value of $\epsilon$ from $f^{max}_2$ to $f^{min}_2$. At each iteration, we solve the model (\ref{exampleEpsilon}) with a fixed bound $\epsilon$ to obtain the optimal solution $\hat{x}_1$, and set then $\epsilon$ equal to $f_2(\hat{x}_1) - \delta$, where $\delta$ is a small step size, ensuring in this way that the previous solution is excluded.}

In the remainder of this section, we focus on solving a single instance of the problem with a fixed $\epsilon$. In Section~\ref{sec:pareto_front}, we vary $\epsilon$ to determine a set of Pareto-optimal solutions and approximate the Pareto front.

\subsection{Route-delivery based formulation}\label{sec:SPformulation}
A \textit{route-delivery} is defined {as} a sequence of nodes visited by a vehicle, {respecting the route length constraint, together with} delivery quantities to the visited nodes that {do not exceed} the demands and {respect} the vehicle capacity. 
Let $R$ be the set of all {feasible} route-deliveries starting from the depot. For a given route-delivery $r\in R$, the binary {indicator} variable $y_{r}$ is {set to} 1 if route-delivery $r$ belongs to the {overall} solution, $q_i^r \, \forall i \in V_c$ {denotes} the delivery amount to node $i$ {by route-delivery~$r$}, and indicator parameter $a_{ir}$ is set to 1 if node $i$ is visited by route-delivery $r$. {Note that if $a_{ir} = 0$ then $q^r_i$ will be zero.} The total length of the vehicle route in the route-delivery $r$ is {denoted} by $t^r$. We define $N_r$ to be the set of nodes that are visited in a route-delivery $r$. {Using these definitions,} we can present an {\mm RDB} formulation of the problem.
\begin{align}
\min \:\: &  \sum\limits_{i\in V_c} (d_i-\sum_{r \in R} q_{i}^r y_{r})+ \frac{\lambda}{D}\sum\limits_{i \in V_c}\sum\limits_{j \in V_c} (\tau^+_{ij} + \tau^-_{ij})  - \gamma \frac{\epsilon-\sum_{r \in R} t^r y_{r} }{\theta}\label{SP:obj}\\
\mbox{s.t. } 
& \tau^+_{ij} - \tau^-_{ij} = (d_i \cdot\sum_{r \in R} q_{j}^r y_{r})-(d_j \cdot \sum_{r \in R} q_{i}^r y_{r}) \qquad \forall i,j \in V_c \label{SP:Linear_Gini}\\
& \sum_{r \in R} t^r y_{r}  \leq \epsilon  \label{SP:Total_time}\\
& \sum_{r \in R}{a_{ir}y_{r} }= 1,  \quad \forall i \in V_c, \label{SP:One_Visit}\\
& \sum_{r \in R} y_{r}  \leq m  \label{SP:Vehicle_No}\\
& \sum_{r \in R} \sum_{i \in N_r} q_{i}^r y_{r} \leq C \label{SP:Supply_Limit} \\
& \tau^+_{ij}, \, \tau^-_{ij}\geq 0, \quad \forall i,j \in V_c \label{SP:variablesBound}\\
& y_{r} \in \{0,1\}, \quad \forall r \in R \label{SP:ybinary}
\end{align}

The objective function (\ref{SP:obj}) is composed of two parts: the first part is the equity measure of the delivery amount, and the second part is {derived from} the total traveling time objective, {our other objective, which is incorporated here to define the slack variable~$s$ of} the augmented $\epsilon$-constraint method \citep{mavrotas2013improved}.
Constraints (\ref{SP:Linear_Gini}) {\mm are} designed to linearize the absolute value term in the objective function.
The total traveling time is restricted by $\epsilon$ in constraint (\ref{SP:Total_time}). Each node should be visited only once, which is {\mm ensured} by Equation (\ref{SP:One_Visit}). Constraints (\ref{SP:Vehicle_No}) and (\ref{SP:Supply_Limit}) limit the number of vehicles and the relief item supply, respectively. 


\subsection{Column generation} \label{sec:ColGen}

To solve the linear relaxation of the previous formulation, we use Column Generation (CG), an iterative method that starts with a subset of route-deliveries and dynamically adds new route-deliveries until optimality. Each CG iteration solves the Restricted Master Problem (RMP) for a limited set of route-deliveries $\bar{R}$, corresponding to the linear relaxation of the RDB formulation (\ref{SP:obj})--(\ref{SP:ybinary}) with $R$ replaced by $\bar{R} \subset R$. CG identifies new route-deliveries from $R \setminus \bar{R}$ by solving a pricing problem, adding deliveries with negative reduced cost to $\bar{R}$ and re-optimizing the RMP. This process repeats until no negative reduced-cost deliveries remain, with the final RMP solution providing a lower bound for the model (\ref{SP:obj})--(\ref{SP:ybinary}). 

\subsubsection{Pricing subproblem} \label{sec:Subproblem}
{The pricing problem identifies a single vehicle route starting from the depot, visiting some nodes, and delivering certain amounts of the relief item to each visited node while adhering to maximum route length and vehicle capacity constraints. Its objective minimizes the reduced cost of the variables~$y_r$. The full mathematical model is provided in the Online Supplement.

By Lemma 1, the optimal solution of the original problem has demand-proportional deliveries for each route. Enforcing this property in the pricing problem does not compromise the completeness of the CG approach, while significantly reducing the number of columns and improving CG convergence. To enforce demand-proportionality, we add the following valid inequality:
\begin{align}
q_i d_j \leq q_j d_i + d_i d_j (1- a_j), \qquad \forall i\in V_c, j \in V_c, i \ne j \label{Sub:demand_porortionality}
\end{align}
where $a_i$ indicates whether node $i$ is visited, and $q_i$ is the delivery amount to node~$i$.

Based on Corollary~1, we can also add valid inequalities to constrain total deliveries:
\begin{align}
& ({C}/{D}) \sum_{i \in V_c} d_i a_i \leq \sum_{i \in V_c} q_i + C z \\
& ({C}/{D}) \sum_{i \in V_c} d_i a_i \leq Q + C z \\
& \sum_{i \in V_c} q_i \geq Q z \label{Sub:valid_inequality_total_delivery}
\end{align}
where $z$ is a binary variable.

We solve the pricing problem using Gurobi 9.5, which efficiently handles large instances within reasonable time limits. However, since the pricing problem is the most time-intensive part of B\&P algorithms \citep{pecin2017improved, costa2019exact}, we accelerate the process using a GRASP heuristic. GRASP constructs initial solutions using a greedy randomized approach and refines them with a local search \citep{resende2010greedy}. Details of the GRASP heuristic are provided in the Online Supplement.

In each CG iteration, we first run GRASP to generate route-deliveries with negative reduced costs. If GRASP fails, we solve the pricing problem using Gurobi.} To accelerate the CG process while managing the size of the column set, we add a maximum of 10 columns using GRASP and 2 columns when employing Gurobi. We utilize Gurobi's 'PoolSolutions' parameter to obtain two solutions with negative reduced costs. We only terminate the CG when the optimal solution of the pricing model solved by Gurobi is non-negative.

\subsubsection{Generating initial columns} \label{sec:InitialHeuristic}

{To enhance the efficiency of our algorithm, we populate set $\bar{R}$ with feasible and high-quality initial columns, as this significantly improves the performance of CG \citep{feillet2010tutorial}. To achieve this, we first construct initial routes using a parallel Clarke and Wright savings heuristic (CW) and then refine these routes with a Tabu Search (TS) algorithm.}

In our Tabu Search, we use a \textit{Relocate} move, which removes a node from one route and inserts it into the best possible position in another route. To restrict the number of possible moves and to accelerate the TS, we employ heuristic pruning (\cite{toth2003granular}). Initially, for each node, we define the set of $n^{\text{near}}$ nearest neighbors and only insert a node before or after one of its neighbors. In every iteration, we randomly select a subset of $\eta$ nodes, using a uniform probability distribution, and evaluate all candidate moves. A move with the best fitness value will be accepted if it is not prohibited. We also use the \textit{Aspiration Criterion} (\cite{glover1990tabu}), meaning that we \textit{do} accept a prohibited move if it improves the best solution found so far. The accepted move is then considered prohibited and will be kept in the tabu list for a certain number of iterations, randomly selected between $\omega_{\text{min}}$ and $\omega_{\text{max}}$. The algorithm continues until the number of consecutive iterations without improvement exceeds $it^{\text{NO}}$. To evaluate the fitness of a new solution, we first decide on the delivery quantities to each node using Algorithm~\ref{Alg:Optimal_quantity}. Next, we calculate the first two terms of the objective (\ref{SP:obj}) and add penalty costs for the violation of constraints on maximum route length and the number of available vehicles to obtain the fitness value. 

Algorithm \ref{Alg:Tabu}, which uses Algorithm~\ref{Alg:Optimal_quantity} as a subprocedure, presents the steps of our TS heuristic. In the description of Algorithm \ref{Alg:Tabu}, we denote the best solution (e.g., a solution with the smallest fitness value) found so far by $s^{BS}$, and the best feasible solution by $s^{BFS}$. We define $s^{IN}$ as the incumbent solution and $\mathcal{S}^{CA}$ as the set of candidate solutions for the next incumbent solution; this set consists of neighbor solutions of $s^{IN}$ generated by the \textit{Relocate} move. A move $(i, k)$ is defined by the node $i$ that is relocated from its current route to the best position in the destination route $k$. Please refer to the Online Supplement for detailed information on our Tabu Search algorithm.

\subsection{Branch and Price} \label{sec:B&P}
\begin{figure}[htb]
\centering
  \includegraphics[width=1\textwidth]{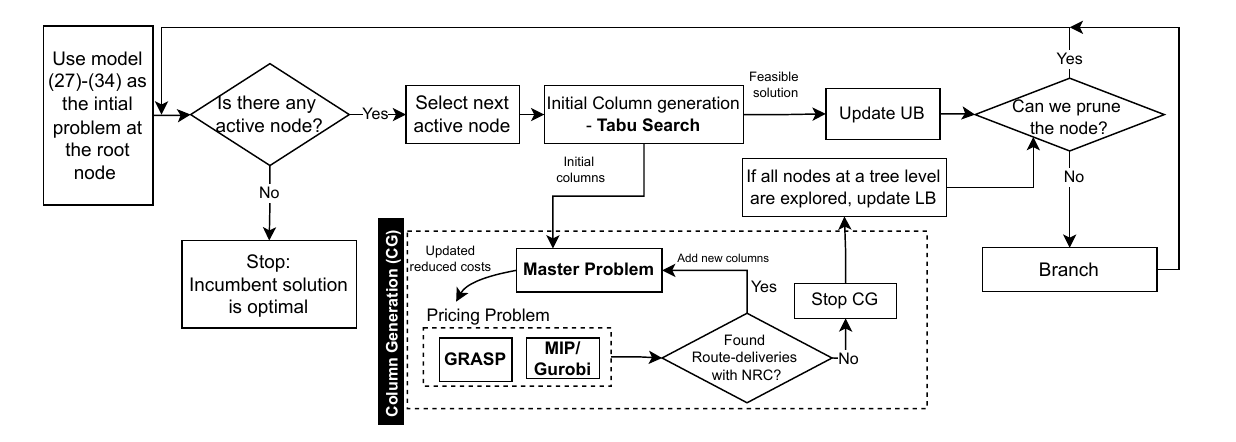}
  \caption{\centering A flowchart representing the main components of our branch-and-price algorithm and their interactions.}
  \label{fig:BnP}
\end{figure}

We propose a branch-and-price (B$\&$P) algorithm, which is essentially a branch-and-bound algorithm {where the lower bound computation in the nodes is performed} by solving linear relaxations of the {\mm RDB} formulation {using} CG. 
If CG provides a fractional solution, a {branching step is carried out} by dividing the feasible {set} of integer solutions in a way {where} the two resulting {subsets} exclude the fractional solution. Each of these new problems will be solved by CG. The algorithm continues to branch on nodes with fractional solutions until the lower bound and the upper bound {coincide. In three cases} we do not explore a node further: (\romannumeral 1) if the solution produced by CG is integer, which provides an upper bound on the problem,(\romannumeral 2) {if} the problem {assigned to the considered node} is infeasible, (\romannumeral 3) {if} the lower bound provided by CG is larger or equal to the {currently best found} upper bound. 
{Figure~\ref{fig:BnP} illustrate the general framework of our B\&P algorithm.}

Given that the common branching on arc variables in arc-based formulations, like our VF model, does not efficiently yield integer solutions quickly \citep{dayarian2015column}, we opt to branch on edges instead, as recommended by  \cite{desaulniers2015branch}. One of the advantages of this branching scheme is that it does not require changing the RMP and can be imposed only by updating the pricing problem and the column set.

We calculate the values $x_{i,j}$ in an optimal solution of the RMP by 
$x_{i,j} = \sum_r y_r b_{i,j}^r $ 
where $b_{i,j}^r$ {is the binary {\mm parameter} indicating whether} route-delivery $r$ uses arc $(i,j)$ or not.  
To accelerate the branching variable selection, we use a simple heuristic rule by branching on the variable that has the closest value to 0.5. Our computational results show the effectiveness of this approach.

Branching creates two new nodes; one with the implicit constraints $x_{i,j} = 0$ and $x_{j,i} = 0$, which we call the left node,
and one with implicit constraint $x_{i,j} = 1$ or $x_{j,i} = 1$, which we refer to as the right node. 
In the left {successor} node, we remove the columns whose routes traverse arc $(i,j)$ or $(j,i)$ from $\bar{R}$. Both $(i,j)$ and $(j,i)$ are removed from the graph $G$ and the subproblem is updated accordingly. 
In the right {successor} node, we eliminate all route-deliveries from $\bar{R}$ that visit node $i$ (node $j$) but do not directly go to the node $j$ (node $i$), {respectively}. To avoid generating routes that do not visit $i$ and $j$ consecutively during CG, we add the following constraints to the subproblem. Let {us} define $\bar{E}$ as the set of edges {that} should be visited. 
\begin{align}
& x_{i,j} = a_j, \quad \forall i,j \in \bar{E}, \, i=0, \\
& x_{i,j}+x_{j,i}=a_i, \quad \forall i,j \in \bar{E}, \, i,j \neq 0, \\
& a_i=a_j   \quad \forall i,j \in \bar{E}, \, i,j \neq 0.
\end{align}
To ensure at least one feasible solution for RMP, we run the TS to find some feasible solutions prior to solving RMP at each node of the B$\&$P tree. 
We regard a node in $B\&P$ tree as infeasible {if} its corresponding graph becomes disconnected as a result of {the omission of edges}, or {if} the subproblem cannot find a feasible route-delivery. Infeasibility also arises when edges in $\bar{E}$  form a cycle that does not contain the depot. 
{To update the upper bound more frequently, we solve the {\mm route-delivery based} formulation (\ref{SP:obj})-(\ref{SP:ybinary}) at every $\chi$ nodes of the search tree with the current set of columns using Gurobi.} {\mm Here, $\chi$ is defined as a parameter of our B\&P approach which controls the frequency of which we solve the route-delivery based model.}

\begin{algorithm}[h]
\footnotesize
\caption{Tabu search algorithm}
\label{Alg:Tabu}
\begin{algorithmic}
\STATE Input: Algorithm parameters, Initial solution $s$.
\STATE Output: A set of feasible solutions, $\mathcal{S}^{F}$
\STATE $s^{BS}, s^{IN} \leftarrow s$ 
\STATE $it \leftarrow 0$
\STATE $\mathcal{S}^{F} =  \emptyset$
\WHILE {$it \leq it^{NO}$}
\STATE  Create set $V^{S}$ which contains $\eta$ nodes randomly selected from $V_c$.
\FOR{ $i$ in $V^{S}$}
\STATE Determine set $V^{near}(i)$ as the set of $n^{near}$ nearest nodes to node $i$
\STATE Define $K^{near}(i)$ as the set of routes visit at least one node in $V^{near}(i)$
\FOR {route $k$ in $K^{near}(i)$}
\STATE \textit{Creating neighbor solution $s'$}
\STATE Insert node $i$ in the best position in route $k$
\STATE Calculate delivery quantities for the updated routes using Algorithm \ref{Alg:Optimal_quantity}
\STATE Calculate $F(s')$
\IF {move ($i$,$k$) is not in Tabu list}
\STATE Add $s'$ to $\mathcal{S}^{CA}$ set
\ELSIF{$F(s') < F(s^{BS})$ or ($s'$ is feasible and $F(s') < F(s^{BFS})$) }
\STATE Add $s'$ to $\mathcal{S}^{CA}$ set
\ENDIF
\ENDFOR
\ENDFOR
\STATE $s_{min}$ $\leftarrow$ the solution with minimum fitness in $\mathcal{S}^{CA}$
\IF {$F(s_{min}) < F(s^{BS})$}
\STATE $s^{BS} \leftarrow s_{min}$
\ENDIF
\STATE $s^{F}_{min}$ $\leftarrow$ the feasible solution with minimum fitness in $\mathcal{S}^{CA}$
\IF {$F(s^{F}_{min}) < F(s^{BFS})$}
\STATE $s^{BFS} \leftarrow s^{F}_{min}$ 
\STATE $\mathcal{S}^{F} =  \mathcal{S}^{F} \cup \{s^{BFS}\}$
\ENDIF
\IF {$F(s_{min}) \leq F(s^{IN})$}
\STATE $s^{IN} \leftarrow s_{min}$
\STATE $it  = 0$
\STATE Generate $\omega_{(i, k)}$ randomly between $\omega_{min}$ and $\omega_{max}$
\STATE Keep the move ($i, k$) for $\omega_{(i, k)}$ in Tabu List. 
\ELSE 
\STATE $it = it + 1$
\ENDIF
\STATE Remove the moves that stayed more than $\omega_{(i, k)}$ iterations in Tabu List.  
\ENDWHILE
\end{algorithmic}
\end{algorithm}

\section{Computational results}\label{sec:computational_test}

We perform a computational evaluation of the proposed exact solution method using two data sets from the literature. The first data set, adopted from \cite{noyan2015stochastic}, is related to the 2011 Van province earthquake, encompassing 60 instances. The second data set, drawn from \cite{kilci2015locating} and \cite{mostajabdaveh2018inequity}, is based on real data from Istanbul's Kartal district and comprises 40 instances. 
{Both datasets are available at GitHub\footnote{\url{https://github.com/mahdims/ReliefAid_Distribution}}.}


Computational experiments were performed on a workstation equipped with an i7-9700k Intel processor and 16 GB RAM, running the Ubuntu 20.04.4 operating system. All mathematical models, including the pricing problem and the VF model, are solved using the Gurobi 9.5 solver. In practical scenarios, our approach addresses problems that arise in the immediate aftermath of a disaster, when time is of the essence. As such, computational solution methods must provide solutions within a reasonable timeframe. To accommodate this constraint, we limit the execution time for both the B\&P algorithm and the Gurobi application of the VF model to 2 hours for all instances. {This is a reasonable time limit as the alternative manual approach would take significantly more time, at least several hours, as it requires solving the combined problem of determining vehicle routes and delivery amounts to shelters.}

For the single-objective minimization problem, we define the percentage gap between the best-known solution value (UB) and the best lower bound (LB) as $Gap = ((UB-LB)/{UB}) \cdot 100$. When solving the problem with both the VF model and B\&P, we employ a relative gap drop below 0.01\% as an additional termination criterion. Based on our preliminary tests, we set the parameter $\chi$ equal to 2 to achieve a balance between execution time and B\&P algorithm convergence.

To determine the RMP objective function, we need to find the value $\theta$ for each considered instance. This parameter represents the difference between the minimum and maximum of $f_2$, i.e., the total traveling time of the vehicles. To save runtime in the context of bi-objective optimization, we replace the computation of minimal and maximal total traveling time for the original problem by solving the corresponding linear relaxations of these two problems with CG. This still yields a valid $\theta$. {\mm We use these $\theta$ values in experiments with both the VF model and the B\&P method so we can directly compare the objective values of the solutions yielded by these methods.}


\subsection{Van earthquake case study}\label{Sec:Van}
\begin{figure}[htb]
\centering
  \includegraphics[width=0.5\textwidth]{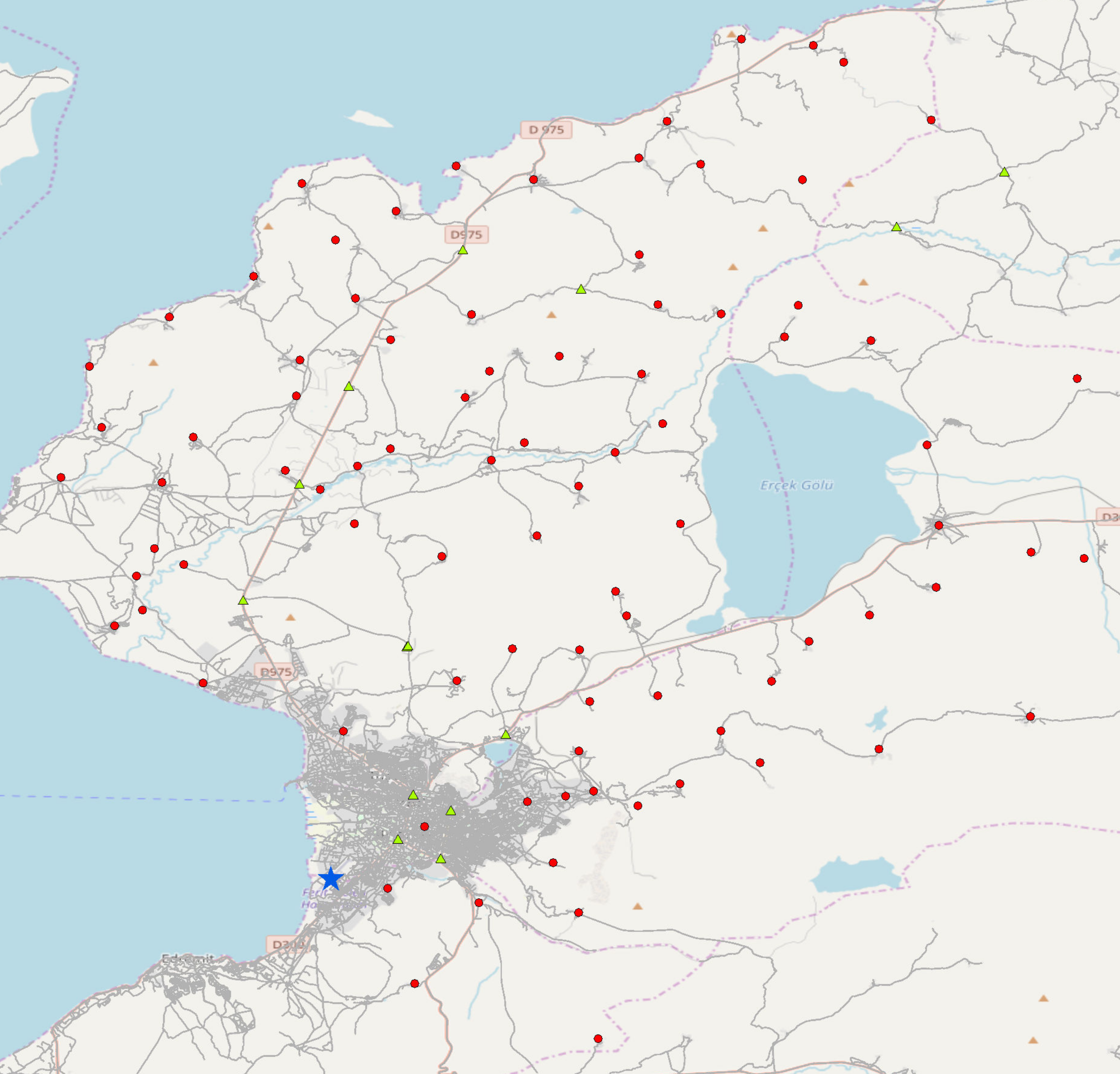}
  \caption{\centering \small Map of Van province: the demand nodes are represented by red dots, the road junction points by green triangles, and the depot by a blue star.}
  \label{fig:van_map}
\end{figure}

Turkey has experienced several severe earthquakes over the past few decades. One of the most destructive earthquakes struck eastern Turkey in Van province on October 23, 2011. The death toll rose to 600 people, while 4,100 individuals were injured. The 7.2 magnitude earthquake and its aftershocks left approximately 60,000 people homeless. One month after the event, about 120,000 people continued to receive warm food daily from the Van Food Assistance Center (\cite{mehdi2013van}). 
{\mm Fig.~\ref{fig:van_map} shows the map of Van province, including the locations of demand nodes and junction points, as well as the depot.}
For more details about this case and how we prepared data for our analysis, refer to the Online Supplement. 

Next, we present and analyze the computational results for the 60 instances derived from the Van province data set. We categorize the instances into four types based on vehicle capacity and maximum route length constraints. Type $A$ instances have abundant vehicle capacity and regular maximum route length. Type $T$ and $VT$ instances feature regular maximum route lengths, with tight and very tight vehicle capacities, respectively. Type $VTL$ instances exhibit both very tight vehicle capacity and maximum route length constraints. {\mm In these experiments, we set the value of $\epsilon$ to $0.85 \, m \cdot \Psi$, recalling that $\Psi$ represents the maximum tour length and $m$ is the number of available vehicles.}

{
Table \ref{Table:Van_results} summarizes computational results for the Van earthquake case study, comparing the VF model solved with Gurobi to our B\&P algorithm across various instance types, with metrics averaged by instance type. The \textit{Gap\%} column shows the optimality gap, and \textit{Time} lists computational time in seconds for both methods. For the B\&P algorithm, we include \textit{\# Nodes} (nodes explored), \textit{Avg. Columns} (average columns per node), and the total columns generated by the MIP and GRASP methods for the pricing problem, labeled as \textit{\# Columns MIP} and \textit{\# Columns GRASP}. The \textit{LB-Gap\%} column indicates the percentage gap between lower bounds from Gurobi and B\&P, with positive values showing B\&P achieved a superior bound. The detailed computational results are reported in the Online Supplement.}

As evident from Table~\ref{Table:Van_results}, the B\&P algorithm consistently finds a better or equal lower bound than the MIP solver across all instances. On average, the B\&P algorithm provides a 1.347\%, 1.185\%, and 1.427\% better lower bound for instances with 15, 30, and 60 nodes, respectively.
Although the MIP solver may converge to the optimal solution faster than the B\&P algorithm in ``easy" instances (i.e., instances of types A and T), the B\&P algorithm outperforms the MIP solver in all challenging instances (i.e., instances of types VT and VTL) in terms of execution time and optimality gap. 
\begin{table}[H]
\mm
\small
\centering
\caption{\small Summary of computational results from the Van earthquake case study}
\vspace{-8pt}
\label{Table:Van_results}
\begin{tabular}{l|cc|cccccc|c}
\hline
\multirow{2}{*}{\begin{tabular}[c]{@{}l@{}}Instance\\ type\end{tabular}} & \multicolumn{2}{c|}{Model} & \multicolumn{6}{c|}{B\&P} & \multirow{2}{*}{LB-Gap\%} \\
 & Gap\% & Time & Gap\% & Time & \begin{tabular}[c]{@{}c@{}}\#\\ Nodes\end{tabular} & \begin{tabular}[c]{@{}c@{}}Avg.\\ Columns\end{tabular} & \begin{tabular}[c]{@{}c@{}}\# \\ Columns\\ MIP\end{tabular} & \begin{tabular}[c]{@{}c@{}}\# \\ Columns\\ GRASP\end{tabular} &  \\ \hline
Van15\_A & 0.000 & 0.6 & 0.001 & 13.7 & 1 & 84 & 42 & 39 & 0.000 \\
Van15\_T & 0.005 & 0.5 & 0.000 & 14.3 & 1 & 102 & 44 & 56 & 0.001 \\
Van15\_VT & 2.699 & 7200.4 & 0.007 & 54.9 & 4 & 310 & 136 & 274 & 2.698 \\
Van15\_VTL & 2.691 & 7200.1 & 0.006 & 109.3 & 7 & 245 & 222 & 325 & 2.690 \\
Van30\_A & 0.000 & 2.1 & 0.004 & 388.4 & 1 & 178 & 71 & 102 & 0.000 \\
Van30\_T & 0.004 & 3.4 & 0.004 & 305.3 & 1 & 285 & 82 & 198 & 0.001 \\
Van30\_VT & 2.395 & 7201.0 & 0.011 & 6273.6 & 15 & 912 & 430 & 1524 & 2.369 \\
Van30\_VTL & 2.394 & 7201.0 & 0.011 & 6064.8 & 18 & 455 & 276 & 1003 & 2.371 \\
Van60\_A & 0.000 & 27.7 & 0.008 & 3403.8 & 8 & 529 & 233 & 424 & 0.000 \\
Van60\_T & 0.018 & 3691.7 & 0.005 & 1195.3 & 1 & 573 & 111 & 454 & 0.001 \\
Van60\_VT & 1.733 & 7202.1 & 0.014 & 7229.9 & 3 & 4107 & 8 & 4470 & 1.639 \\
Van60\_VTL & 4.292 & 7201.6 & 0.011 & 3709.5 & 10 & 841 & 19 & 1327 & 4.066 \\ \hline
Average & 1.352 & 3911.0 & 0.007 & 2396.9 & 6 & 718 & 139 & 849 & 1.320
\end{tabular}
\end{table}

Solving instances of types VT and VTL proves to be highly challenging, with the MIP solver unable to converge to the optimal solution within the 2-hour time limit for any of these instances. In contrast, the B\&P algorithm finds high-quality solutions for all these instances, with an average gap of only 0.01\%. {For all 60 tested instances, our proposed algorithm either provides an optimal solution and verifies its optimality or finds a very high-quality solution with an average gap of 0.007\% within the specified time limit. 

From our computational results, it is evident that as the problem becomes more restrictive, regardless of its size, the optimality gap reported by the MIP-solver drastically increases.} In contrast, the optimality gap for the B\&P algorithm remains relatively stable. Remarkably, despite the B\&P algorithm's shorter runtime, its relative gap is consistently smaller than the one reported by the model, particularly for challenging instances. 

In summary, our computational results on the Van data set suggest that the B\&P algorithm is {distinctly more efficient} than solving the VF model by a MIP-solver. It can provide the optimal solution or a very high-quality solution in a reasonable time. Therefore, the proposed B\&P algorithm is reliable to be utilized for decision making in post-disaster situations when both time and solution quality are of the utmost importance.

\subsection{Istanbul-Kartal real data} \label{Sec:Kartal}
\begin{figure}[htb]
\centering
  \includegraphics[width=0.45\textwidth]{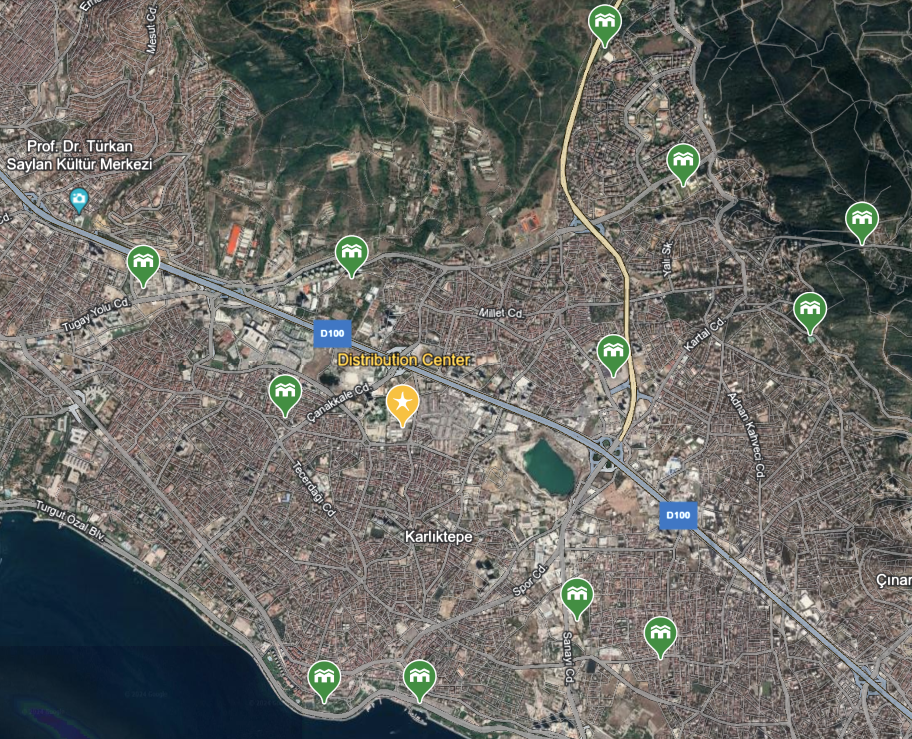}
  \caption{\small \centering Map of Istanbul's Kartal district highlighting selected shelter locations (marked in green) and the central relief aid distribution center (marked in yellow). }
  \label{fig:kartal_map}
\end{figure}

In this section, we test our approach on the data of the Kartal district of Istanbul provided by \cite{kilci2015locating},
which were also used in~\cite{mostajabdaveh2018inequity} for a location-allocation problem rather than a routing problem. {\mm  As illustrated in Fig.~\ref{fig:kartal_map}, we adopt the twelve shelter locations proposed by \cite{mostajabdaveh2018inequity} and select a depot location based on the actual location of a relief supply distribution center in the region. We assume that $m = 3$ identical vehicles are available for distributing relief supplies among the shelter locations. Additionally, we have fixed the value of $\epsilon$ to $0.85m \cdot \Psi$.
For more information about the data preparation process, please refer to the Online Supplement.}

{\mm Table \ref{Table:kartal_results} summarizes our computational results for the Kartal case study, comparing the performance of the VF model solved with Gurobi and our B\&P algorithm across various instance types. The meaning of the columns is the same as in Table \ref{Table:Van_results}. We report the detailed computational results in the Online Supplement. }

\begin{table}[H]
\mm
\small
\centering
\caption{\small Summary of computational results for the Kartal case study}
\vspace{-8pt}
\label{Table:kartal_results}
\begin{tabular}{l|cc|cccccc|c}
\hline
\multirow{2}{*}{\begin{tabular}[c]{@{}l@{}}Instance\\ type\end{tabular}} & \multicolumn{2}{c|}{Model} & \multicolumn{6}{c|}{B\&P} & \multirow{2}{*}{LB-Gap\%} \\
 & Gap\% & Time & Gap\% & Time & \begin{tabular}[c]{@{}c@{}}\#\\ Nodes\end{tabular} & \begin{tabular}[c]{@{}c@{}}Avg.\\ Columns\end{tabular} & \begin{tabular}[c]{@{}c@{}}\# \\ Columns\\ MIP\end{tabular} & \begin{tabular}[c]{@{}c@{}}\# \\ Columns\\ GRASP\end{tabular} &  \\ \hline
Kartal\_A & 0.000 & 0.2 & 0.004 & 250.6 & 68 & 90 & 661 & 496 & -0.003 \\
Kartal\_T & 5.148 & 7200.0 & 0.003 & 931.6 & 308 & 291 & 2310 & 2128 & 5.566 \\
Kartal\_VT & 3.834 & 7200.0 & 0.005 & 915.8 & 341 & 230 & 2116 & 1958 & 4.055 \\
Kartal\_VTL & 5.179 & 7200.0 & 0.007 & 783.0 & 70 & 140 & 790 & 512 & 5.563 \\ \hline
Average & 3.540 & 5400.1 & 0.005 & 720.3 & 197 & 188 & 1469 & 1274 & 3.795
\end{tabular}
\end{table}

In these instances, the average optimality gap obtained by the B\&P algorithm {turned out to be only about 1/775 of} the average gap reported by the VF model solver, {which is especially significant in view of the much shorter runtime of B\&P (only 11/100), compared to the model solver.} {Remarkably, the B\&P algorithm reaches the desired optimality gap in all 40 instances except one.} {Contrary to that}, the {model solver} acts unpredictably, {obtaining gaps of largely varying size}, {and only converges to the desired gap in type A instances. This confirms  the superiority of the B\&P approach again.}

{\mm
\subsection{Contributions of the algorithmic components}
Our B\&P algorithm benefits from several enhancements, including an initial column generation heuristic, GRASP heuristics for the pricing problem, and valid inequalities for the mathematical pricing model. To evaluate the impact of these components on the performance of our B\&P algorithm, we conducted a comprehensive analysis by introducing three distinct variations: (i) B\&P without GRASP, (ii) B\&P without valid inequalities, and (iii) B\&P without initial route enhancements via Tabu Search. For this experiment, we utilized challenging instances from the Van earthquake dataset.
{Our findings indicate that the inclusion of Tabu Search for initial column enhancement is essential for achieving superior algorithm performance, as omitting it results in a 54.4\% increase in runtime and gaps that are 59 times larger. Similarly, excluding valid inequalities in the pricing model leads to a 9.3\% longer runtime and a 12.5\% increase in the average gap. In contrast, while the GRASP algorithm improves efficiency for small and medium-sized instances, its effectiveness diminishes for larger problem instances.}
For a detailed exposition of our methodology, experimental setup, and an in-depth discussion of the results, we invite readers to consult the online supplement accompanying this paper.
}

{
\subsection{Pareto front approximation}\label{sec:pareto_front}

In this section, we aim to provide non-dominated Pareto solutions by changing the value of $\epsilon$ between $f^{max}_2$ and $f^{min}_2$. We obtain $f^{min}_2$ by solving the problem with minimization of the total traveling time as the objective. Similarly, $f^{max}_2$ is equal to the total traveling time of the optimal solution of the single objective problem with IAAF as the objective. 
{To approximate the Pareto front, we solve a series of problem instances with varying $\epsilon$ values, starting from $f^{\max}_2$ and decrementing $\epsilon$ until reaching $f^{\min}_2$. At each step $t$, $\epsilon^t$ denotes the current $\epsilon$ value, and $f_2^t$ represents the second objective value of the optimal solution. Based on the augmented $\epsilon$-constraint method \citep{mavrotas2013improved} and our preliminary tests, we set the step size for decreasing $\epsilon$ to 1. We also apply the bypass mechanism from \cite{mavrotas2013improved}, updating $\epsilon^t$ as $\min\{\epsilon^{t-1} - 1, f_2^{t-1} - 1\}$ to skip some $\epsilon$ values.}

We select 20 instances with different characteristics from both case studies, including at least one instance from each instance type and size. We use the B\&P algorithm or the MIP Solver to find the non-dominated solutions, selecting the approach that could find the optimal solution faster based on our experiments in the previous sections\footnote{ Note that, in practice, the precise selection of an approach is not possible without conducting some prior experiments. We recommend conducting preliminary tests with both algorithms on a fixed $\epsilon$ value and selecting the algorithm that converges faster. However, we generally suggest using the MIP Solver for smaller instances and B\&P for larger ones.}. We restrict the runtime for each instance of the auxiliary problem (i.e., the problem for a fixed $\epsilon$ value) to 600 seconds. Based on our preliminary experiments—presented in the Online Supplement—this time limit is sufficient for our solver to find solutions that are very close to optimal. Table~\ref{Tab:pareto_sol} presents the results of these experiments. The first column lists the names of the instances. The subsequent columns report the selected solver, total number of instances solved, number of non-dominated solutions found, total runtime, $f^{max}_2$ and $f^{min}_2$, respectively.

\begin{table}[ht]
\footnotesize
\centering
\caption{Pareto solutions}
\label{Tab:pareto_sol}
\begin{tabular}{@{}l|lcclll@{}}
\toprule
Instance     & Solver     & \# Solved & \# Non-dominated & Time   & {\mm $f^{max}_2$} & {\mm $f^{min}_2$} \\ \midrule
Van15\_A1    & MIP-Solver & 8            & 4           & 1.3      & {\mm 41888}   & {\mm 21692}\\
Van15\_T3    & MIP-Solver & 12           & 6           & 5.3      & {\mm 43766}   & {\mm 21692}\\
Van15\_T5    & MIP-Solver & 8            & 6           & 2.2      & {\mm 39872}   & {\mm 21694}\\
Van15\_VT5   & B\&P       & 16           & 7           & 458.5    & {\mm 44615}   & {\mm 21696}\\
Van15\_VTL1  & B\&P       & 8            & 6           & 1218.5   & {\mm 30708}   & {\mm 21696}\\
Van30\_A5    & MIP-Solver & 12           & 6           & 361.0    & {\mm 95335}   & {\mm 32440}\\
Van30\_T1    & MIP-Solver & 24           & 15          & 829.9    & {\mm 83740}   & {\mm 32524}\\
Van30\_T2    & MIP-Solver & 12           & 8           & 385.3    & {\mm 91901}   & {\mm 32628}\\
Van30\_VT5   & B\&P       & 26           & 23          & 8557.2   & {\mm 88200}   & {\mm 32524}\\
Van30\_VTL3  & B\&P       & 29           & 27          & 7274.0   & {\mm 71425}   & {\mm 32628}\\
Van60\_A2    & MIP-Solver & 5            & 2           & 124.8    & {\mm 158612}  &{\mm 47336}\\
Van60\_T2    & B\&P       & 35           & 28          & 18527.8  & {\mm 179760}  &{\mm 47349}\\
Van60\_T5    & B\&P       & 25           & 18          & 13023.6  & {\mm 142704}  &{\mm 47349}\\
Van60\_VT1   & B\&P       & 91           & 53          & 44807.7  & {\mm 166134}  &{\mm 47189}\\
Van60\_VTL1  & B\&P       & 107          & 61          & 53109.5  & {\mm 138648}  &{\mm 47189}\\
Kartal\_A2   & MIP-Solver & 7            & 5           & 1.5      & {\mm 46794 }  &{\mm 16341}\\
Kartal\_T3   & B\&P       & 12           & 11          & 289.3    & {\mm 42577 }  &{\mm 17664}\\
Kartal\_T6   & B\&P       & 10           & 9           & 1957.7   & {\mm 38032 }  &{\mm 17236}\\
Kartal\_VT7  & B\&P       & 8            & 6           & 231.5    & {\mm 34268 }  &{\mm 17664}\\
Kartal\_VTL5 & B\&P       & 10           & 9           & 147.8    & {\mm 20406 }  &{\mm 16698}\\ \bottomrule
\end{tabular}
\end{table}
Although our step size to decrease $\epsilon$ was one, the total number of solved instances stayed relatively small, even for large instances. On average, 69\% of the solved problem instances resulted in a non-dominated solution. In Figure~\ref{fig:pareto_front}, we approximate the Pareto fronts of four instances Van15\_VTL1, Van30\_VT5, Van60\_T2 and Kartal\_VTL5 by connecting the non-dominated solutions. The figures show the conflicting nature of the two objectives. 

\begin{figure}[ht] 
  \begin{subfigure}[b]{0.5\linewidth}
    \centering
    \includegraphics[width=0.9\linewidth]{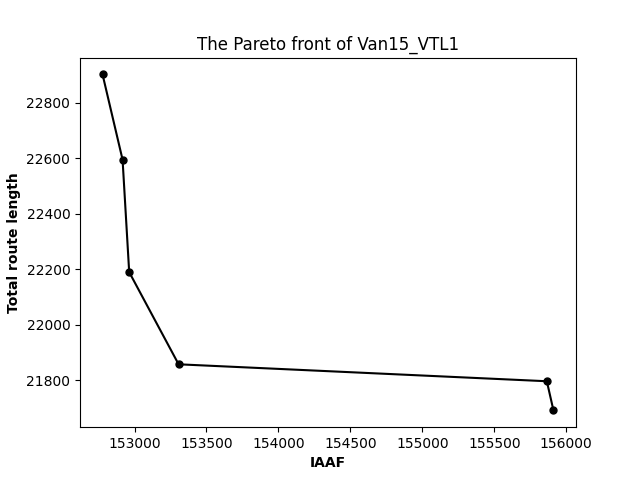} 
    \caption{Van15\_VTL1 instance} 
    \label{fig:pareto_Van15_VTL1} 
    \vspace{6pt}
  \end{subfigure}
  \begin{subfigure}[b]{0.5\linewidth}
    \centering
    \includegraphics[width=0.9\linewidth]{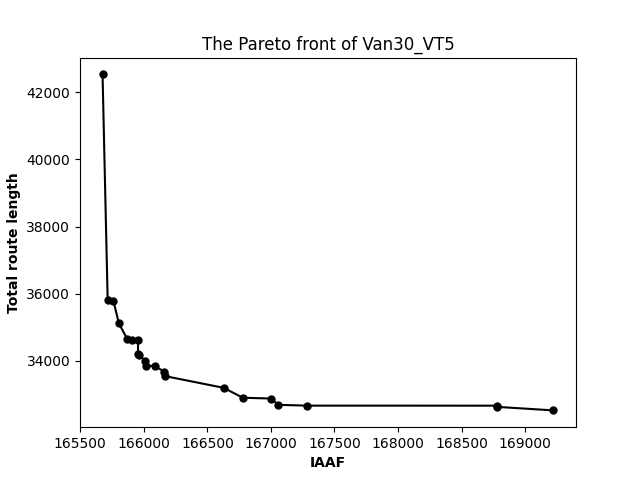} 
    \caption{Van30\_VT5 instance} 
    \label{fig:pareto_Van30_VT5} 
    \vspace{6pt}
  \end{subfigure} 
  \begin{subfigure}[b]{0.5\linewidth}
    \centering
    \includegraphics[width=0.9\linewidth]{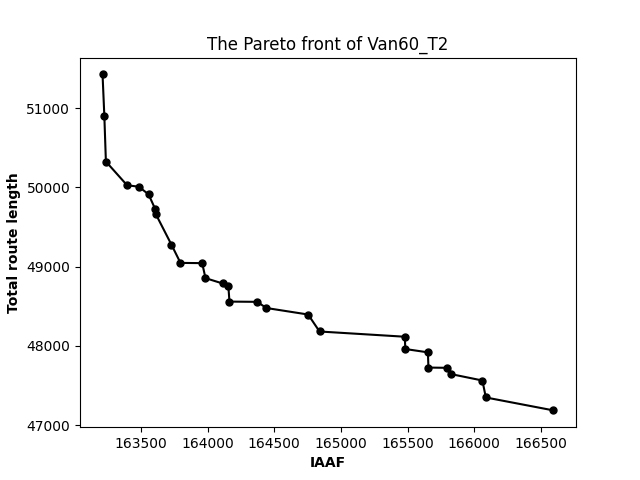} 
    \caption{Van60\_T2 instance} 
    \label{fig:pareto_Van60_T2} 
  \end{subfigure}
  \begin{subfigure}[b]{0.5\linewidth}
    \centering
    \includegraphics[width=0.9\linewidth]{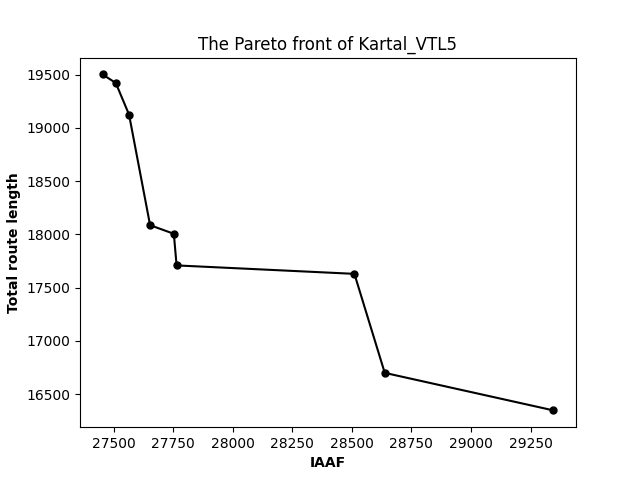} 
    \caption{Kartal\_VTL5 instance} 
    \label{fig:pareto_Kartal_VTL5} 
  \end{subfigure} 
  \caption{Pareto front of four selected instances}
  \label{fig:pareto_front} 
\end{figure}

\subsection{Price of Fairness}\label{Sec:price_fairness}
{\mm
In this section, we analyze the consequences of decision-making processes that exclusively emphasize either fairness or efficiency and compare them to the use of the IAAF objective.
The focus will be mainly on cases where the total traveling time is not restrictive. 
Three objectives will be investigated:
(i)  minimization of total unsatisfied demand:
     $\min D \mu = \min \sum_{i \in V_c} (d_i-v_i)$, with solution MinUnD;
(ii) minimization of Gini's mean absolute difference multiplied by~$D$:
    $\min D \Delta  = \min \frac{1}{D} \sum_{i \in V_c}\sum_{j \in V_c} |d_iv_j - d_jv_i|$, with solution MinGini; and
(iii) minimization of the IAAF objective multiplied by~$D$:
    $\min D {\cal I} = \min D (\mu + \lambda \Delta)$, with solution MinIAAF.
In addition to Gini's mean absolute difference, we look at the Gini index ${\cal G} = \Delta/(2\mu)$. The parameter~$\lambda$ is chosen as $\lambda = 1/2$ (see Section~\ref{sec:problem_defination}). 



We define the {\em price of fairness} (PoF) as the percentage increase in total unsatisfied demand caused by some ``fair'' solution (such as MinGini or MinIAAF), compared to perfect efficiency as achieved using the objective UnD, denoting total unsatisfied demand:
$$
\mbox{PoF(MinGini)} = ({\mbox{UnD(MinGini)}}/{\mbox{UnD(MinUnD)}}-1) \cdot 100
$$
for MinGini, and analogously for MinIAAF.
Fig.~\ref{fig:UnD_obj_comparision} presents the average price of fairness for each instance type when employing the solutions MinGini and MinIAAF, respectively. It turns out that the 
MinGini objective produces solutions with perfect equity (i.e., the Gini index equals zero) for all problem instances, and these perfectly equitable solutions exhibit an average 50\% increase in total unsatisfied demand compared to the solutions with perfect efficiency. On the other hand, solutions obtained by using the IAAF objective turn out to be able to achieve perfect efficiency, resulting in a zero price of fairness. Still, they are not able to reduce the Gini index to zero, such that the MinGini solutions are superior to them in terms of fairness. 
Nevertheless, the MinIAAF solutions are equally efficient as those achieved by the MinUnD objective while simultaneously decreasing the Gini index by 33.68\% relative to the UnD objective on average.
}

\begin{figure}[!htb]
\centering
  \includegraphics[width=75ex]{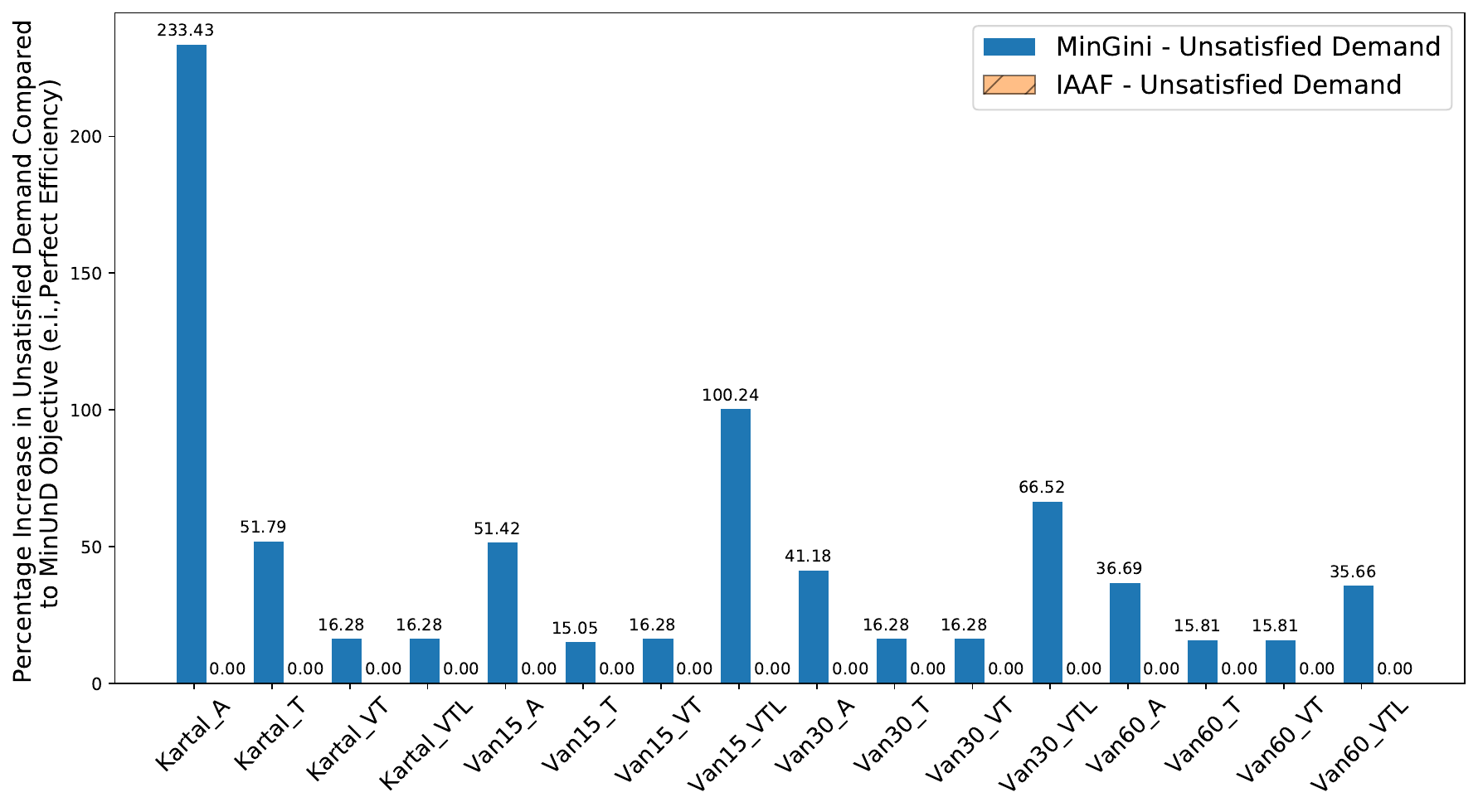}
  \vspace{-15pt}
  \caption{\centering Evaluating percentage increases in Unsatisfied Demand for MinGini and MinIAAF objectives relative to the perfect efficiency (MinUnD Objective)}
  \label{fig:UnD_obj_comparision}
\end{figure}

In summary, for scenarios without tight travel time constraints and with a moderate inequity-aversion parameter~$\lambda$, we found that IAAF optimization produces lexicographically optimal solutions across all real-world test instances, prioritizing total demand coverage (primary objective) over minimizing Gini’s mean absolute difference (secondary objective), resulting in a PoF of zero. However, this is no longer the case when tighter total travel time limits are imposed; in this scenario, an essential trade-off between efficiency and fairness emerges.

The following table shows the PoF for the Kartal~A2 test instance, varying according to ten values of the upper bound (UTTT) on total travel time:
$$
\begin{array}{r|rrrrrrrrrr}
\mbox{UTTT}&42635&21593&20560&20262&19661&19284&18775&17781&17702&17269\\
\hline
\mbox{PoF (\%)}&0&0&0&0&10.66&4.46&14.50&0&0&0\\
\end{array}
$$
We can see that for intermediate bounds on the total traveling time, a nonzero price of fairness has to be taken into account.
Of course, the range of scenarios where the price of fairness is larger than zero depends on the inequity aversion parameter~$\lambda$: as~$\lambda$ increases, indicating a higher ambition for fairness, the resulting PoF from the optimization model also grows.

{It can be observed that prioritizing efficiency over fairness is advisable only when time constraints are either abundant or excessively restrictive.} For practitioners, this suggests that in such conditions, lexicographic optimization with demand coverage as the more important criterion and distributional fairness as the second criterion may yield preferable plans —similar to those generated by our approach. However, if time constraints are moderately restrictive, this approach may prove insufficient. In such cases, a balanced consideration of coverage and fairness is essential to prevent significantly unfair plans.
}

\section{Conclusion}\label{sec:conclution}
We proposed a bi-objective optimization problem and an efficient algorithm for its exact solution to plan the {   allocation} and distribution of scarce relief supplies via a fleet of vehicles to a set of shelters, originating from a single depot, in disaster response.  {   Unlike previous studies, we optimized both an objective function measuring effectiveness and another accounting for fair supply allocation.} To assess fairness, we utilized the widely recognized Gini coefficient. 
 {   We focused on the early post-disaster stage, where insufficient supply at the depot makes full coverage unattainable within the specified time frame, necessitating the allocation of delivery amounts to each shelter. }

{  
We minimized both the total travel time of vehicles and an inequity-averse aggregation function (IAAF), which combines total unsatisfied demand with Gini's absolute differences in unsatisfied demand. We established conditions for achieving perfect equity and derived key characteristics of the optimal solution with regard to the IAAF. Leveraging these insights, we proposed an algorithm that computes the optimal delivery amounts for {\em given} feasible vehicle routes using only simple arithmetic operators. 
Since a straightforward application of standard MIP solvers proved inefficient for tackling even the single-objective subproblems, we developed a branch-and-price (B\&P) algorithm capable of handling real-world instance sizes. Our B\&P method integrates several problem-specific enhancements, including route-deliveries as columns, a customized GRASP algorithm for the pricing problem, valid inequalities for the pricing problem, and an initial column generation heuristic based on the Tabu Search algorithm.

Our computational tests on realistic data from a past earthquake in Van, Turkey, and predicted data for the Kartal region of Istanbul, where a major earthquake is expected, demonstrate the superior performance of the B\&P algorithm compared to commercial MIP solvers. Our bi-objective method can reduce the inequity of aid distribution by 34\% without compromising efficiency.
The results suggest that in cases where time constraints are either very loose or extremely tight, lexicographic optimization with demand coverage as the more important criterion and distributional fairness as the second criterion may yield good plans. However, if time constraints are moderately restrictive, a balanced approach is necessary to prevent inequitable outcomes.}

Our problem setting currently restricts each demand point to a single-vehicle service, omitting potential efficiency gains from split deliveries, which could also enhance supply reliability through visit redundancy. Extending the problem to incorporate split deliveries is a promising avenue for future research. Additionally, exploring alternative frameworks for inequity aversion in optimization, such as the one presented in \cite{chen2022combining}, would be valuable. Finally, alternative mathematical formulations of the problem could be explored. We conducted preliminary experiments with alternative models based on the two-commodity flow formulation by \cite{baldacci2004exact} and the multi-commodity formulation by \cite{letchford2015stronger}. However, MIP solvers demonstrated better convergence with the model presented in this study. A deeper analysis would be worthwhile in future research.

\vspace*{1ex}
{\bf Acknowledgements.}
While working on most parts of the study, the first author was a Ph.D. candidate at the College of Engineering, Ko{\c c} University, Istanbul, Turkey. 
\vspace*{1ex}

{\small
\begin{spacing}{0.7}
\bibliography{refd}
\end{spacing}
\bibliographystyle{apalike}
}
\end{document}


\maketitle
\addappheadtotoc
\counterwithin{figure}{section}
\counterwithin{table}{section}
\counterwithin{equation}{section}

\section{Proof of Lemma 1} \label{appx:lemma1_proof}
Suppose $v=(v_i)_{i \in V_c}$ is an optimal solution, and consider an arbitrary $k_0 \in K$. We will show that after changing the values $v_i$ for all $i \in V_{k_0}$ to demand-proportional values $v_i^*$ with the same sum $\sum_{i \in V_{k_0}} v_i^* = \xi_{k_0}$, the solution remains feasible, and the objective function value does not increase, i.e., $v^*$ is optimal again. As one can apply this consideration consecutively to all $k_0 \in K$, it will prove the assertion of the Lemma. 

Without loss of generality, assume $k_0=1$. Define
$$
v_i^* = \xi_1 \cdot \frac{d_i}{D_1} \; \: (i \in V_1), \quad v_i^* = v_i \; \: (i \in V_k, \: k \neq 1).
$$
First, the feasibility of $v^* = (v_i^*)$ has to be shown. The feasibility of $v = (v_i)$ implies 
$$
0 \le \xi_k \le D_k \; \: (k \in K), \quad \xi_k \le Q \; \: (k \in K), \quad \sum_{k \in K} \xi_k \le C,
$$
and hence
$$
v_i^* \ge 0 \: \; (i \in V_1), \quad v_i^* \le D_1 \cdot \frac{d_i}{D_1} = d_i \: \; (i \in V_1),
$$
as well as
$$
\sum_{i \in V_1} v_i^* = \xi_1 \cdot \frac{D_1}{D_1} = \xi_1 \le Q
$$
and
$$
\sum_{i \in V_c} v_i^* = \xi_1 + \sum_{i \in V_c \setminus V_1} v_i = \xi_1 + \sum_{k \neq 1} \xi_k = \sum_{k \in K} \xi_k \le C.
$$
Therefore $v^*$ is feasible. Now consider the objective function. The first term of
$$
\sum\limits_{i\in V_c} (d_i-v_i)+ \frac{\lambda}{D}\sum\limits_{i \in V_c}\sum\limits_{j \in V_c}|d_iv_j-d_jv_i|
$$
is identical for $v$ and $v^*$, since
$$
\sum_{i \in V_c} v_i = \sum_{k \in K} \xi_k = \sum_{i \in V_c} v_i^*.
$$
It remains to show that
\begin{equation}\label{AgeAstar}
\sum_{i \in V_c} \sum_{j \in V_c} |d_i \, v_j - d_j \, v_i| \ge 
\sum_{i \in V_c} \sum_{j \in V_c} |d_i \, v_j^* - d_j \, v_i^*|.
\end{equation}
Set $\bar{V}_1 = V_c \setminus V_1$, decompose the l.h.s.~of (\ref{AgeAstar}) into three terms $A_1 + A_2 + A_3$ with
$$
A_1= \sum_{i \in V_1} \sum_{j \in V_1} |d_i \, v_j - d_j \, v_i|,
\quad
A_2 = 2 \cdot \sum_{i \in V_1} \sum_{j \in \bar{V}_1} |d_i \, v_j - d_j \, v_i|,
\quad
A_3 = \sum_{i \in \bar{V}_1} \sum_{j \in \bar{V}_1} |d_i \, v_j - d_j \, v_i|,
$$
and decompose the r.h.s.~of (\ref{AgeAstar}) into analogously defined terms $A_1^* + A_2^* + A_3^*$, replacing $v_i$ by $v_i^*$ $(i \in V_c)$. Because of $v_i = v_i^*$ for $i \in \bar{V}_1$, it is obvious that $A_3 = A_3^*$.
Furthermore,
$$
A_1^* = \sum_{i \in V_1} \sum_{j \in V_1}  |d_i \cdot \xi_1 \cdot \frac{d_j}{D_1} - d_j \cdot \xi_1 \cdot \frac{d_i}{D_1}| = 0,
$$
so $A_1 \ge A_1^*$.
It remains to show that $A_2 \ge A_2^*$, i.e., that
\begin{equation}\label{stilltoshow}
\sum_{i \in V_1} \sum_{j \in \bar{V}_1} |d_i \, v_j - d_j \, v_i| \ge 
\sum_{i \in V_1} \sum_{j \in \bar{V}_1} |d_i \, v_j - d_j \, v_i^*|
\end{equation}
(note that $v_j^* = v_j$ for $j \in \bar{V}_1$). It suffices to show that for each fixed $j \in \bar{V}_1$, the inequality
\begin{equation}\label{sufftoshow}
\sum_{i \in V_1} |d_i \, v_j - d_j \, v_i| \ge 
\sum_{i \in V_1} |d_i \, v_j - d_j \, v_i^*|
\end{equation}
holds. Division of (\ref{sufftoshow}) by $d_j$ and setting $\delta_{i,j} = d_i \cdot \frac{v_j}{d_j}$ transforms (\ref{sufftoshow}) into the equivalent representation
\begin{equation}\label{eqtoshow}
\sum_{i \in V_1} |\delta_{i,j} - v_i| \ge 
\sum_{i \in V_1} |\delta_{i,j} - v_i^*|.
\end{equation}
For all~$i \in V_1$,
$$
\frac{v_i^*}{\delta_{i,j}} = \xi_1 \, \frac{d_i}{D_1} \cdot \frac{d_j}{d_i \, v_j} = \xi_1 \, \frac{d_j}{D_1 \, v_j}
= \xi_1 / \sum_{s \in V_1} \delta_{s,j},
$$
an expression which is actually independent of~$i$. Substituting for $v_i^*$ by means of the last equation, we get
$$
\sum_{i \in V_1} |\delta_{i,j} - v_i^*| = \sum_{i \in V_1} \left|
\delta_{i,j} - \delta_{i,j} \cdot \frac{\xi_1}{\sum_{s \in V_1} \delta_{s,j}} \right|
= \sum_{i \in V_1} \delta_{i,j} \left| 1 - \frac{\xi_1}{\sum_{s \in V_1} \delta_{s,j}} \right|
$$
$$
= \left| 1 - \frac{\xi_1}{\sum_{s \in V_1} \delta_{s,j}} \right| \: \left( \sum_{i \in V_1} \delta_{i,j} \right)
= \left| \sum_{i \in V_1} \delta_{i,j} - \xi_1 \right| = \left| \sum_{i \in V_1} (\delta_{i,j} - v_i) \right| 
\le \sum_{i \in V_1} |\delta_{i,j} - v_i|.
$$
Thus (\ref{eqtoshow}) is shown, and the proof is complete.

\section{Proof of Proposition 1} \label{appx:prop1_proof}

For each $k$, by the assumption $D_k/D \le Q/C$, solution $\xi = (\xi_k)$ given by
$$
\xi_k = \frac{C}{D} \cdot D_k \quad (k \in K)
$$
satisfies  $\xi_k = C \cdot (D_k/D) \le Q$, and because of $C \le D$, also $\xi_k \le D_k$ holds. Moreover, $\sum_{k \in K} \xi_k = C \le C$. Together this shows that $\xi$ is feasible w.r.t.~the constraint
$$
0 \le \xi_k \le \min (Q, D_k) \; \: \forall k, \quad \sum_{k \in K} \xi_k \le C.
$$
Furthermore, $\xi$ clearly minimizes the part $D - \sum_k \xi_k$ of the objective function
\begin{equation}\label{xObj}
D - \sum_k \xi_k + \frac{\lambda}{D} \, \sum_{k \in K} \sum_{\ell \in K}  
| D_\ell \xi_k - D_k \xi_\ell|
\end{equation}
and the second part $\frac{\lambda}{D} \, \sum_{k \in K} \sum_{\ell \in K}  
| D_\ell \xi_k - D_k \xi_\ell|$ of this objective function becomes zero, which is minimal as well. This completes the proof.

\section{Proof of Proposition 2}\label{appx:prop2_proof}
Without loss of generality, suppose that the condition $\frac{D_k}{D} > \frac{Q}{C}$ is satisfied for $k=1$, i.e., 
$\frac{D_1}{D} > \frac{Q}{C}$. Let $\xi$ be an optimal solution of (22) - (23), presented in the main text. Assume that the statement of the proposition is false, i.e., $\xi_1 < Q$. We show that a contradiction follows.

First of all, it is shown that $\xi_1 < C$: Suppose $\xi_1 = C$. Then, since $\sum_{k \in K} \xi_k \le C$, it follows that $\xi_k = 0$ for all $k > 1$, so we are dealing with the solution $\xi= (C,0,\ldots,0)$. Define $\xi' = (\epsilon, C - \epsilon, 0,\ldots,0)$. For sufficiently small~$\epsilon > 0$, solution $\xi'$ is feasible again, and it is better than solution~$\xi$: With $\bar{D} = \sum_{\ell > 2} D_\ell$,
$$
\psi(\xi') = D - C + \frac{2 \lambda}{D} \cdot \left\{ |D_1 (C - \epsilon) - D_2 \epsilon| + \sum_{\ell >2} |D_\ell \, \epsilon| + \sum_{\ell >2} |D_\ell (C - \epsilon)| \right\}
$$
$$
= D - C +  \frac{2 \lambda}{D}\, [ (D_1 + \bar{D}) C - (D_1 + D_2) \epsilon],
$$
which is strictly decreasing with growing $\epsilon$, so $\psi(\xi') < \psi(\xi)$ in contradiction to the optimality of $\xi$. This shows $\xi_1 < C$.
Distinguish now two cases:

 {\em Case (i):} $\sum_{k \in K} \xi_k < C$ (total delivery amount is less than the total supply).
Set $\xi' = (\xi_1+\epsilon,\xi_2,\xi_3,\ldots,\xi_m)$. Because of $\xi_1 < C$, this is a feasible solution, provided that $\epsilon > 0$ is sufficiently small, and that $\xi_1$ is not already identical to the upper bound $\min(Q, D_1)$. However, the latter can be excluded:
because of $\xi_1 < Q$ by assumption, $\xi_1 = \min(Q, D_1)$ could only be the case if $\min(Q, D_1) = D_1$ and hence $D_1 \le Q$. By ${D_1}/{D} > {Q}/{C}$, however, we have $D_1 > (D/C) \, Q \ge Q$. So $\xi'$ is feasible for small~$\epsilon$. 

We use the following general property (expressed in costs rather than benefits) of the Gini-based social welfare function ${\cal I}$ adopted in this paper: 
Suppose that ${\cal I}$ is applied to the costs $c_i$ of individuals $i \in I = \{1,\ldots,N\}$, and consider two cost vectors $c =(c_1,\ldots,c_N)$ and $c' = (c'_1,\ldots,c'_N)$ with $c_i = c'_i$ $(i \neq i_0)$ and $c_{i_0} < c'_{i_0}$. Then ${\cal I}[c']$ is strictly smaller (i.e., better)  than ${\cal I}[c]$.

Note that {\em strict} monotonicity is asserted here, which requires a justification. As an OWA, measure ${\cal I}$ can be represented in the form
${\cal I}[c] = \sum_{i=1}^N q_i \, c^{(i)}$,
where $c^{(1)} \le c^{(2)} \le \ldots \le c^{(N)}$ is the sequence of cost values $c_i$ sorted in non-descending order, and the weights $q_1 \le q_2 \le \ldots \le q_N$ are non-descending constants specific to the OWA function. For our generalized Gini measure, 
$$
q_i = \frac{2}{N^2} \left\{ \frac{N}{2} + \lambda(2i - N - 1) \right\} \quad (i=1,\ldots,N)
$$
(see the Online Supplement to \cite{mostajabdaveh2018inequity}). It is easy to see that if $q_i > 0$ for all~$i$, then the claimed strict monotonicity property holds. Now, for $0 \le \lambda \le 1/2$,
$$
q_1 =\frac{1}{N} - 2 \lambda \cdot \frac{N-1}{N^2} = \frac{1}{N} \, (1 - 2 \lambda + \frac{2 \lambda}{N}) \ge \frac{2 \lambda}{N^2} > 0,
$$
and hence also $q_i > 0$ for all $i$. 

Applied to benefits $v_i$ instead of costs $c_i = d_i - v_i$, the above result says that if the benefit of some individual is increased, leaving the benefits of the other individuals unchanged, then the evaluation by ${\cal I}$ is strictly improved (i.e., decreased). Successive application to each individual in $V_1$ immediately yields $\psi(\xi') < \psi(\xi)$, in contradiction to the optimality of~$\xi$.

{\em Case (ii):} $\sum_{k \in K} \xi_k = C$ (total delivery amount is equal to the total supply).
Since $\xi_1 < C$, there must be some $k \in K \setminus \{1\}$ for which $\xi_k > 0$. We choose that $k \in K \setminus \{1\}$ for which ${\xi_k}/{D_k}$ is maximal; in other words, we look for the ``best off'' area $V_k$ for which the delivered supply per individual is at a maximum. Without loss of generality, assume that the
route with maximal delivery-to-demand ratio is $k=2$.
Set $\xi' = (\xi_1+\epsilon,\xi_2-\epsilon,\xi_3,\xi_4,\ldots,\xi_m)$. For $\epsilon$ sufficiently small, it is ensured that the solution $\xi'$ is feasible; this would only be wrong if we would have $\xi_1 = \min(Q, D_1)$, such that $\xi_1$ could not be increased anymore. However, $\xi_1 = \min(Q, D_1)$ can be excluded as in case~(i) above. So, again, $\xi'$ is feasible for small~$\epsilon$.

Observe now that ${\xi_1/}{D_1} < {\xi_2}/{D_2}$: Suppose this would not be true. Then from  ${\xi_1}/{D_1} \ge {\xi_2}/{D_2}$, even  ${\xi_1}/{D_1} \ge {\xi_k}/{D_k}$ for all~$k$ would follow. However, because of ${D_1}/{D} > {Q}/{C}$,
this would entail
$$
\frac{\xi_k}{D_k} \le \frac{\xi_1}{D_1}  < \frac{Q}{D_1} < \frac{C}{D} \quad \forall k \in K,
$$
and therefore
$$
\xi_k < \frac{C}{D} \, D_k \quad \forall k \in K, \quad \sum_{k \in K} \xi_k < \frac{C}{D} \cdot D = C,
$$
in contradiction to the assumption of case (ii). 

Now, we apply the Pigou-Dalton Principle of Transfers (PDPT) (\cite{moulin2004fair}). It states that for a proper inequity-averse evaluation function, a transfer of a sufficiently small amount $e > 0$ from a richer individual to a poorer individual results in a strict improvement of the inequity-averse evaluation function (cf.~\cite{weymark2003generalized}, \cite{fleurbaey2006social}). Because of $\xi_2/D_2 > \xi_1/D_1$, the individuals in $V_2$ are richer than those in~$V_1$. The generalized Gini function used by us satisfies the PDPT. Therefore, the transfer of the quantity~$\epsilon$ from $V_2$ to $V_1$ improves the objective function, i.e., $\psi(\xi') < \psi(\xi)$, so $\xi$ cannot be optimal in case~(ii) either. This completes the proof.

\section{Proof of Corollary 1}\label{appx:cor1_proof}

According to Prop.~2,
if $\frac{D_k}{D} \geq \frac{Q}{C}$ then $\xi_k = Q$ and $\frac{C}{D} D_k  \leq \xi_k \leq D_k$ does not hold. In case $\frac{D_k}{D} < \frac{Q}{C}$, according to Algorithm~A.1,
$\xi_k$ will be equal to $ \frac{C}{D} D_k$ or $\frac{C'}{D'} D_k$. We can observe that any value of $\frac{C'}{D'}$ calculated during the algorithm is larger than $\frac{C}{D}$. This shows the assertion.

{
\section{Multi-objective approach}
We use the augmented $\epsilon$-constraint method by \cite{mavrotas2013improved}, which augments the main objective function by {subtracting} the normalized slack variable of the added constraint. The augmented $\epsilon$-constraint method transforms the problem to
\begin{equation}\label{exampleEpsilon}
\min\{f_1(x) - \gamma \cdot \frac{s}{\theta}: \:  f_2(x) +s = \epsilon, \: x \in {\cal X} \}    
\end{equation}
where $\gamma$ is a {small machine precision constant}, 
$\theta=f^{max}_2-f^{min}_2$ is the range of $f_2(x)$ values, and $s \ge 0$ is a nonnegative slack.

To adopt this approach, we choose our inequity measure as the main objective function and use the 
total traveling time {objective}
{for the} $\epsilon$-constraint, {i.e., we interchange the two objective functions in the main text, Subsection 3.1, such that now $f_1 = z_2$ and $f_2 = z_1$.} The reason for the interchange is that it is computationally easier to handle the more complicated fairness-related objective function under formally simple constraints than, vice versa, a formally simple objective function under more complicated constraints. Note that this has only computational relevance and does not have any impact on the mutual rank of the two objective functions.

The optimal (i.e., minimal) value of the problem considering only the total travelling time objective yields $f^{min}_2$. To save runtime, we combine the augmented $\epsilon$-constraint method with the adaptive $\epsilon$-constraint approach: We start with the computation of the minimal solution~$x^*_1$ of the problem with $f_1$ as the objective function and set $f^{max}_2$ and $\epsilon$ equal to $f_2(x^*_1)$. To find the Pareto front, we then iteratively decrease the value of $\epsilon$ from $f^{max}_2$ to $f^{min}_2$. At each iteration, we solve the model (\ref{exampleEpsilon}) with a fixed bound $\epsilon$ to obtain the optimal solution $\hat{x}_1$, and set then $\epsilon$ equal to $f_2(\hat{x}_1) - \delta$, where $\delta$ is a small step size, ensuring in this way that the previous solution is excluded.}

\section{Mathematical model of the pricing problem}\label{appx:pricing_model}

Let us define the dual variables $\pi_{i,j}^1$, $\pi^2$, $\pi_i^3$, $\pi^4$ and $\pi^5$ corresponding to the constraints (28), (29), (30), (31) and (32)
in the main paper, respectively. 
The reduced cost for variable $y_{r}$ can be calculated as follows: 
$$
c_{r} = -\sum_{i\in V_c}\sum_{j\in V_c}{(d_jq_i^r-d_iq_j^r)\pi^1_{i,j}} - t^r \pi^2 - \sum_{i\in V_c} a_{ir}\pi_i^3 - \pi^4 - \sum_{i \in V} (\pi^5+1) q_i^r + \frac{\gamma}{\theta}t^r.
$$

The following variables describe a solution to the pricing problem: The binary variable $x_{i,j}$ {attains the value one} if and only if edge $(i,j)$ is used by the vehicle. {The binary variable} $a_i$ {indicates whether} demand point $i$ is visited by the vehicle or not, and $q_i$ is the amount of delivery to the node~$i$. The mathematical formulation of the subproblem is presented below. 
\begin{align}
\min \: & -\sum_{i\in V_c}\sum_{j\in V_c}{\bar{\pi}^1_{i,j} (d_jq_i-d_iq_j) } - \sum_{(i,j) \in A} (\bar{\pi}^2 - \frac{\gamma}{\theta} )t_{i,j}x_{i,j}  - \sum_{i\in V_c} \bar{\pi}^3_i a_{i} - \bar{\pi}^4 - \sum_{i \in V_c} (\bar{\pi}^5 + 1) q_i \label{Sub:obj}\\
\mbox{s.t. } 
& \sum_{(i,j) \in A}t_{i,j}x_{i,j} \leq \Psi \label{Sub:maxtour}\\
& \sum_{i \in V_c} x_{0,i}=1 \label{Sub:depotout}\\
& \sum_{i \in V_c} x_{i,n+1}=1 \label{Sub:depotin}\\
& \sum_{j \in V_c} x_{i,j} - \sum_{j \in V_c} x_{j,i}=0, \qquad \forall i \in V_c \label{Sub:balance}\\
&\sum_{j \in V_c} x_{i,j}=a_{i} \qquad \forall i \in V_c \label{Sub:Ax_relation}\\ 
& \sum_{i\in S, j\in S} x_{i,j} \leq \sum_{i \in S, i\neq k}a_i, \qquad \forall S \subset V_c, \forall k \in S 
\label{Sub:subtour}\\
& q_i \leq d_i a_i, \qquad \forall i \in V_c \label{Sub:delivery}\\
& \sum_{i \in V_c} q_i \leq Q \label{Sub:capacity}\\
& x_{i,j} \in \{0,1\}, \quad \forall i,j \in V_c \nonumber \\
&a_i\in \{0,1 \},\quad \forall i \in V_c \nonumber \\
& q_i \geq 0, \quad \forall i \in V_c \label{Sub:vars}
\end{align}

Objective function (\ref{Sub:obj}) is the reduced cost of a route-delivery (column) in RMP. Note that  $\bar{\pi}_{i,j}^1$, $\bar{\pi}^2$, $\bar{\pi}_i^3$, $\bar{\pi}^4$ and $\bar{\pi}^5$ are constant values {given by the dual solution of the RMP.} 
Constraint~(\ref{Sub:maxtour}) ensures that the route length does not exceed the specified maximum route length.
Constraints (\ref{Sub:depotin}) and (\ref{Sub:depotout}) make sure that the route starts at the depot and ends at the depot. The balance between incoming and outgoing flow for each shelter is {ensured} by Eq.~(\ref{Sub:balance}). Constraint (\ref{Sub:Ax_relation}) specifies the relation between $a_i$ and $x_{i,j}$ variables such that if node $i$ is not visited in the route, then all variables $x_{i,j}, \, (j \in V_c)$ should be zero. Subtour elimination is enforced by inequality (\ref{Sub:subtour}) which is adopted from \cite{feillet2005traveling} for {the TSP with profit problem}. The delivery amount to node $i$ is limited by its demand $d_i$, which {is guaranteed} by constraint (\ref{Sub:delivery}). Constraint (\ref{Sub:capacity}) restricts the total delivery amounts in a route by the vehicle capacity.

\section{Impact of total traveling time on equity}

\begin{figure}[h] 
  \begin{subfigure}[b]{0.5\linewidth}
    \centering
    \includegraphics[width=0.9\linewidth]{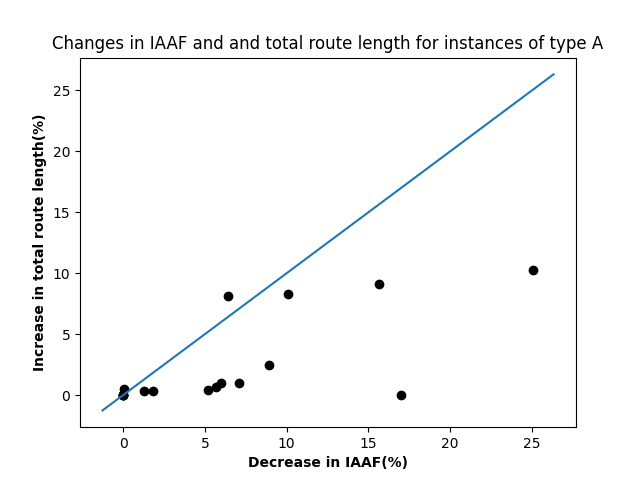} 
    \caption{Instances of type A} 
    \label{fig:two_obj_TypeA} 
    \vspace{7pt}
  \end{subfigure}
  \begin{subfigure}[b]{0.5\linewidth}
    \centering
    \includegraphics[width=0.9\linewidth]{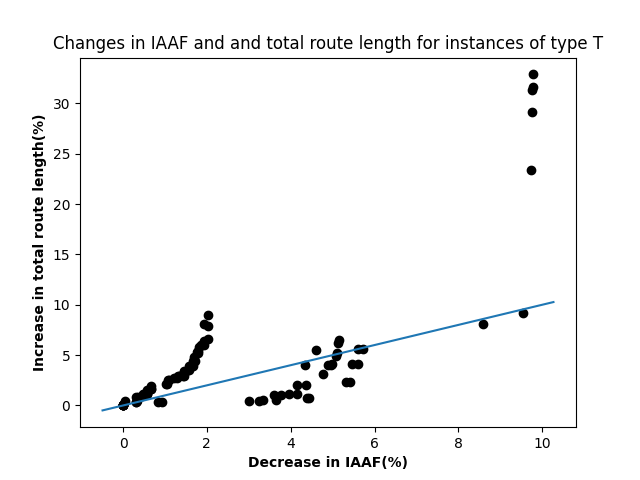} 
    \caption{Instances of type T} 
    \label{fig:two_obj_TypeT} 
    \vspace{7pt}
  \end{subfigure} 
  \begin{subfigure}[b]{0.5\linewidth}
    \centering
    \includegraphics[width=0.9\linewidth]{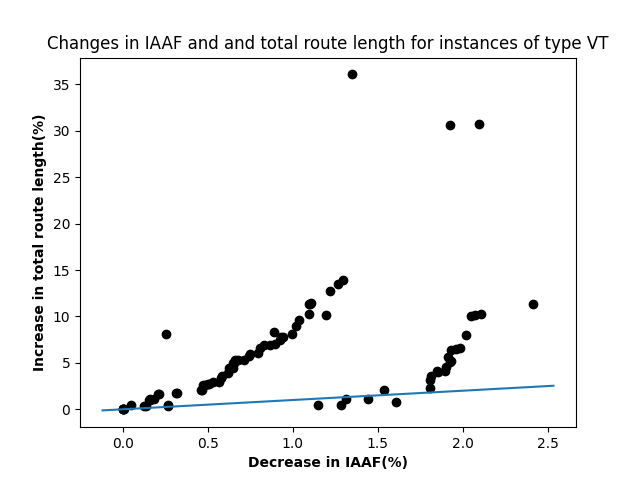} 
    \caption{Instances of type VT} 
    \label{fig:two_obj_TypeVT} 
  \end{subfigure}
  \begin{subfigure}[b]{0.5\linewidth}
    \centering
    \includegraphics[width=0.9\linewidth]{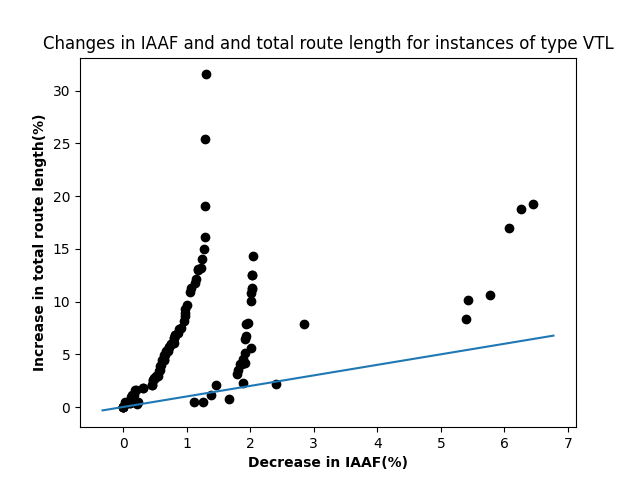} 
    \caption{Instances of type VTL} 
    \label{fig:two_obj_TypeVTL} 
  \end{subfigure} 
  
  \caption{Improvement of IAAF by increasing the total traveling time}
  \label{fig:route_length_IAAF} 
\end{figure}

We select the solution of only minimizing the total traveling time as the reference solution.  For each non-dominated solution, we calculate the percentage of total traveling time increase and IAAF decrease compared to the instance's reference solution. To draw a general insight, we combine all data points obtained from instances of the same type, which produces the four plots presented in Fig~\ref{fig:route_length_IAAF}. Points below the continuous line show the non-dominated solutions where the percentage of increase in total traveling time is smaller than the percentage of improvement in IAAF.

In type A instances,  IAAF is significantly improved by a slight increase in total traveling time. This does not hold in instances of type T but many points are below the straight line.
On the contrary, for instances of types  VT and VTL, most of the points are above the straight line (see Fig.~\ref{fig:two_obj_TypeVT} and Fig.~\ref{fig:two_obj_TypeVTL}), which means that  a larger amount of
the total traveling time should be sacrificed. 
Finally, as the instances become more restricted regarding vehicle capacity and maximum route length, the improvement in IAAF becomes smaller and harder to achieve.


\section{GRASP heuristic for the pricing problem}\label{appx:GRASP}
\setcounter{algorithm}{0}
\renewcommand{\thealgorithm}{3.\arabic{algorithm}}

A solution to the pricing problem, {and thus also the output of our GRASP algorithm,} consists of two parts, {a single} vehicle route that visits a subset of nodes, and {a vector of} delivery quantities to the visited nodes. 
The constructive phase {of our GRASP adaptation} starts with sorting the nodes according to a {score} and selecting one node randomly from a truncated list~RCL of nodes. Then a route that only visits the chosen node is created. 

{
To create the node selection measure, we use a proxy on the contribution of nodes on the pricing problem objective function (B.1). Since we cannot decide on the $q_i$ variables before knowing which nodes are selected, we consider the objective value change for a unit increase in delivery quantity $q_i$ in our node score measure. Following this approach, we set $q_i = 1, \, \forall i \in V_c$ and try to separate the pricing problem objective (B.1) for every node $i$. Note that the second term in the objective cannot be written for each node separately, thus we ignore it in our heuristic measure.} The node {scores} are calculated as follows:
$\sigma(i) = - \sum_{j \in V_c} \left(\bar{\pi}^1_{ij} - \bar{\pi}^1_{ji}\right) d_j - \bar{\pi}^3_{i} - (\bar{\pi}^5 +1 ) , \quad \forall i \in V_c$.
The list~RCL is created such that RCL$= \{i \, | \, \sigma(i) \leq \alpha \cdot \sigma^{min} \}$, 
where $\alpha$ is a parameter that controls the size of the list, and $\sigma^{min}$ is the minimum node {score}. The heuristic selects one node from RCL {uniformly at random}. Each possible insertion {of a node} is evaluated according to the reduced cost of the route-delivery {(see below)}.
The node will be inserted to the best position {for which} the insertion does not violate the vehicle route length constraint. To calculate the reduced cost of any insertion, we obtain the delivery amounts {by} using Algorithm \ref{Alg:q_heuristic}. The heuristic updates the RCL list by removing the previously selected nodes and repeating the above steps. This is continued until none of the remaining nodes can be inserted feasibly into the route, or all of them have a {non-negative} value. {As a result, we obtain an incumbent solution which will be improved in the second phase.} Algorithm \ref{Alg:GRASP-initial} shows the detailed steps of the constructive phase of our GRASP heuristic.  
\begin{algorithm}[H]
\srcsize
\caption{GRASP to solve pricing problem: Constructive phase}
\label{Alg:GRASP-initial}
\begin{algorithmic}
\STATE Input: $V_c$, $\Psi$, Dual variables, $\sigma(i)$.
\STATE Output: One route-delivery 
\STATE Create an empty route, $\bar{r} \leftarrow [0,n+1]$.
\WHILE {$any(\sigma(i) < 0 , \forall i \in I^{un})$  and $I^{un} \neq \emptyset$}
\STATE $RCL= \{i \in I^{un}| \sigma(i) \leq \alpha \times \sigma^{min} \}$.
\STATE Select $\bar{i}$ randomly from RCL.
\STATE $CL = \{\}$
\FOR {p in [all possible insertion positions of $\bar{r}$]}
\STATE Build route $r'$ by inserting $\bar{i}$ into $\bar{r}$ at $p$.
\IF {length $r'$ is less than or equal to $\Psi$}
\STATE Calculate delivery amounts for nodes served by $r'$ with Algorithm \ref{Alg:q_heuristic}. 
\STATE Calculate the reduced cost of the $r'$, $RC(r')$, and store in CL.
\ENDIF
\ENDFOR
\IF {$CL \neq \emptyset $}
\STATE Find the route-delivery in CL with minimum reduced cost, $r''$.
\IF { $RC(r'') \leq RC(r)$}
\STATE Replace $\bar{r}$ with $r''$.
\ENDIF
\ENDIF
\STATE Remove $\bar{i}$ from $I^{un}$
\ENDWHILE
\STATE Return route delivery $\bar{r}$
\end{algorithmic}
\end{algorithm}

In the second phase, we 
{improve} the incumbent solution {by a Local Search procedure based on neighborhoods defined by two types of moves:} Let us denote by $I^{un}$ the set of nodes that are not visited by the current route. The first move, \textit{Exchange}, exchanges one node from $I^{un}$ with one node from the current vehicle route. The other move, \textit{Insertion}, inserts one node from $I^{un}$ to the best possible position in the current route, given that the insertion is feasible regarding the maximum route length {constraint}. 
At each iteration, our heuristic explores all possible Exchange moves and {accepts} the one with the maximum improvement. With the updated incumbent solution, the heuristic evaluates all possible Insertion moves, {accepts} the best one, and moves on to the next iteration. 
We terminate the procedure when no further improvement can be found {anymore}. During the algorithm, we store every solution with a negative reduced cost and use these solutions as new columns in the column generation. The pseudocode of the improvement phase of our GRASP heuristic is presented by Algorithm~\ref{Alg:GRASP-improve}.
\begin{algorithm}[H]
\srcsize
\caption{GRASP to solve pricing problem: Improvement phase}
\label{Alg:GRASP-improve}
\begin{algorithmic}
\STATE Input: $V_c$, $\Psi$, Dual variables, An initial route-delivery $\bar{r}$.
\STATE Output: A set of route-deliveries with negative reduced cost $\bar{R}$. 
\STATE Define $I(\bar{r})$ as the set of nodes (except depot) visited by $\bar{r}$.
\WHILE {Improvements are found in previous iteration}
\STATE $I^{un} = V_c \setminus I(\bar{r})$, $CL = \{\}$. 
\FOR {$i \in I^{un}$}
\FOR {$j \in I(\bar{r})$}
\STATE Build $r'$ by replacing $i$ with $j$ in $\bar{r}$.
\IF  {length $r'$ is less than or equal to $\Psi$}
\STATE Calculate the reduced cost of $r'$, $RC(r')$, and store in CL if $RC(r') \leq 0$.
\ENDIF
\ENDFOR
\ENDFOR
\IF {$CL \neq \emptyset $}
\STATE $\bar{R} = \bar{R} \bigcup CL$.
\STATE Replace $\bar{r}$ with the route-delivery with minimum reduced cost in CL. 
\ENDIF
\STATE $I^{un} = V_c \setminus I(\bar{r})$,  $CL = \{\}$.
\FOR {$i \in I^{un}$}
\FOR {p in [all possible insertion positions of $\bar{r}$]}
\STATE Build route $r'$ by inserting $\bar{i}$ into $\bar{r}$ at $p$.
\IF {length $r'$ is less than or equal to $\Psi$}
\STATE Calculate the reduced cost of $r'$ $RC(r')$, and store in CL if $RC(r') \leq 0$.
\ENDIF
\ENDFOR
\ENDFOR
\IF {$CL \neq \emptyset $}
\STATE $\bar{R} = \bar{R} \bigcup CL$.
\STATE Replace $\bar{r}$ with the route-delivery with minimum reduced cost in CL. 
\ENDIF
\ENDWHILE
\STATE Return $\bar{R}$.
\end{algorithmic}
\end{algorithm}

To evaluate a considered move during the GRASP heuristic, we decide first on the delivery quantities to each node visited by the route, which creates a route-delivery. Then we calculate the reduced cost associated with a move using the pricing problem objective function. We use a simple algorithm to calculate delivery amounts. 
The algorithm presented in Algorithm \ref{Alg:q_heuristic} is guaranteed to find the optimal delivery amounts in the pricing problem for a given route {so that only the routing part is solved heuristically. Let us assume the vehicle route $\bar{r}$ is given, and $V_{\bar{r}}$ is the set of nodes that this route serves. Optimality of the deliveries with respect to the objective function~(B.1) is seen as follows:} 

For a given route $\bar{r}$ that visits the nodes in $V_{\bar{r}}$, the pricing problem reduces to 
\begin{align}
    \min \: & -\sum_{i\in V_c}\sum_{j\in V_c}{ \bar{\pi}^1_{i,j} (d_jq_i-d_iq_j) } - (\bar{\pi}^5 + 1) \sum_{i \in V_{\bar{r} }} q_i \label{RSub:obj}\\ 
    & q_j = \frac{d_j}{d_i} q_i, \quad \forall i \in V_c, j \in V_c \label{RSub:poropostional_demand}\\ 
    & \frac{C}{D} \sum_{i \in V_{\bar{r}}} d_i \leq \sum_{i \in V_{\bar{r}}} q_i + C z\\
    & \frac{C}{D} \sum_{i \in V_{\bar{r}}} d_i \leq Q + C z\\
    & \sum_{i \in V_{\bar{r}}} q_i \geq Q z \\
    & \sum_{i \in V_{\bar{r}}} q_i \leq Q  \\
    & 0 \leq q_i \leq d_i, \quad \forall i \in V_{\bar{r}}  \\
    & q_i =0, \quad \forall i \in V_c \setminus V_{\bar{r}} \label{RSub:last} 
\end{align}
The objective function can be rewritten as
$$
-\sum_{i\in V_{\bar{r}}}\sum_{j\in V_c}{ \bar{\pi}^1_{i,j} d_j q_i} + \sum_{i\in V_c} \sum_{ j \in V_{\bar{r}} }  \bar{\pi}^1_{i,j}  d_i q_j  - (\bar{\pi}^5 + 1) \sum_{i \in V_{\bar{r} }} q_i. 
$$

We denote $\kappa_i$ as the coefficient of variable $q_{\hat{i}}$ in the above objective which for a specific $\hat{i} \in V_{\bar{r}}$ computes as
\begin{equation}\label{equ:q_coefficient_pricing}
    \kappa_{\hat{i}} =  - \sum_{j\in V_c} { \bar{\pi}^1_{\hat{i},j} d_j  + \sum_{i\in V_c}  \bar{\pi}^1_{i,\hat{i}}  d_i - (\bar{\pi}^5 + 1)} = -(\bar{\pi}^5 + 1) - \sum_{j\in V_c} ( \bar{\pi}^1_{\hat{i},j} - \bar{\pi}^1_{j,\hat{i}}) d_j  ,  
\end{equation}
Using equation (\ref{RSub:poropostional_demand}) we can transform the problem (\ref{RSub:obj})-(\ref{RSub:last}) by substituting $q_j$, $j \in V_{\bar{r}} \setminus \{\hat{i}\}$ with $q_{\hat{i}} d_j/d_{\hat{i}}$.
This turns the above optimization problem into a continuous optimization problem with one bounded variable. Variable $q_{\hat{i}}$ has its optimal value on its lower or upper bound. Thus, we can solve the problem to optimality by the greedy algorithm presented in Algorithm~\ref{Alg:q_heuristic}. 
\begin{algorithm}[H]
\srcsize
\caption{Algorithm to obtain the delivery amounts}
\label{Alg:q_heuristic}
\begin{algorithmic}
\STATE Input: A route-delivery $\bar{r}$, $d_i$, $\kappa_i$ (computed by (\ref{equ:q_coefficient_pricing})), $C$, $Q$.
\STATE Output: Route-delivery $\bar{r}$ with new delivery amounts $q_i, \forall i \in V_{\bar{r}}$.
\STATE $D_{\bar{r}}= \sum_{i \in V_{\bar{r}}} d_i$.
\STATE $\hat{i} = argmin\{ d_i | i \in V_{\bar{r}}\}$. 
\IF {$\frac{D_{\bar{r} }}{D} \geq \frac{Q}{C} $}
\STATE $q_{\hat{i}} = d_{\hat{i}} {Q}/{D_{\bar{r}}}$.
\ELSE
\STATE Calculate the coefficient of $q_{\hat{i}}$ in the objective function, $\hat{c} = \sum_{i \in V_{\bar{r}}} \kappa_i {d_i}/{d_{\hat{i}}}$.
\IF {$\hat{c} < 0$}
\STATE $q_{\hat{i}} = \min\{ d_{\hat{i}}, d_{\hat{i}} {Q}/{D_{\bar{r}}} \}$.
\ELSE
\STATE $q_{\hat{i}} = d_{\hat{i}} {C}/{D} $.
\ENDIF 
\ENDIF
\FOR {other nodes i in $V_{\bar{r}}\setminus \{\hat{i}\} $}
\STATE $q_{i} = q_{\hat{i}} {d_i}/{d_{\hat{i}}}$
\ENDFOR
\STATE Return route-delivery $\bar{r}$ with new delivery amounts $q_i$.
\end{algorithmic}
\end{algorithm}

\section{Tabu Search algorithm to create initial columns}\label{appx:TabuSearch}
\setcounter{algorithm}{0}
\renewcommand{\thealgorithm}{4.\arabic{algorithm}}

The parallel {variant of the} CW algorithm starts with vehicle routes containing {only} the depot and one customer. {Routes are merged to larger routes in a step-by-step manner.}
At each step, the cost savings of merging all pairs of routes are calculated, and a feasible merger with the highest savings is implemented. The algorithm is repeated until none of the remaining mergers is feasible. A merger is {considered} feasible {only if} the length of the resulting route is less than $\Psi$. We adapt the CW heuristic to create {solutions} with 
load-balanced routes. We abandon the vehicle capacity constraint and assume {that} route feasibility only depends on maximum route length.
We adapted the CW algorithm to our problem by an extension of the saving cost computation: instead of considering only the changes in route lengths, we also take the effect of joining two routes on Gini’s mean absolute difference into account. For the sake of brevity, we omit the details.


We continue merging routes until we can not find any feasible {merger}, even if the saving has a negative value. Thus, the CW heuristic {tries} to minimize the number of vehicles. The CW heuristic always ensures the feasibility of the routes regarding the maximum length $\Psi$, but the number of created routes may {by larger than} $m$, or the total traveling time {may exceed} $\epsilon$. As the next step, we apply the Lin-Kernighan heuristic (see \cite{lin1973effective}), one of the most effective heuristics to solve the TSP, to each route in order to reduce the route length. We use a recent implementation of the algorithm, LKH-2.0.9, accessible at \hyperlink{http://webhotel4.ruc.dk/~keld/research/LKH/}{http://webhotel4.ruc.dk/~keld/research/LKH/}. 
To create route-deliveries from the set of routes, we employ Algorithm A.1 to find the optimal delivery quantities.

Next, we develop a Tabu Search (TS) heuristic to improve the solution {quality while retaining} the solution feasibility. For many years, TS has remained among the most popular algorithms to solve VRPs, due to its efficiency, flexibility, and simplicity. Please refer to \cite{osman1993metastrategy}, \cite{gendreau1994tabu} and \cite{toth2003granular} for successful TS designs. 
Tabu search is an iterative algorithm that starts with an initial solution $s$ and tries to find a better solution by moving to a {neighbor} solution in each iteration. The implemented moves are considered forbidden for a number of iterations to avoid returning to previous solutions. 

We use the neighborhood {definition} that is generated by a \textit{Relocate} move, which removes a node {from} a route and inserts it to the best possible position in another route. 
To restrict the number of possible moves and {to} accelerate the TS, we use heuristic punning~(\cite{toth2003granular}). First, for each node, we define the set of $n^{near}$ nearest neighbors and only insert a node before or after one of its neighbors.  
In every iteration, we randomly select a subset of $\eta$ nodes, {using a uniform probability distribution}, and evaluate all candidate moves. A move with the best fitness value will be accepted if it is not prohibited. We also use the \textit{Aspiration Criterion} (\cite{glover1990tabu}), meaning that we {\em do} accept a prohibited move if it improves the best solution found so far. The accepted move is {then} considered prohibited and will be kept in the tabu list for a certain number of iterations, randomly selected between $\omega_{min}$ and $\omega_{max}$.
The algorithm continues until the number of consecutive iterations without improvement exceeds $it^{NO}$.  

To evaluate the fitness of a new solution~$s'$, first, we decide on the delivery quantities to each node using Algorithm~A.1. 
{Denote by $r$} a route-delivery in set $RD(s')$, {where the latter is the set of all route-deliveries belonging} to solution $s'$. Then we calculate the fitness~$F(s')$ of solution $s'$ as follows: 
\begin{align}
F(s')= D \cdot {\cal I} (s')  & +  \frac{\gamma}{\theta} \max(0, \sum_{r\in RD(s')} t^r -\epsilon) \nonumber \\
& + \nu^{MLR} \sum_{r\in RD(s')} \max(0,t^r -\Psi)  + \nu^{NV} \max(0,|RD(s')| - m)
\end{align}

The first two terms of the above fitness function are calculated according to the SP formulation's objective function. The {third and forth} terms act as a penalty cost for a violation {of the constraints on} maximum route length and the number of available vehicles, respectively. {The parameters} $\nu^{MLR}$ and $\nu^{NV}$ are the {penalty} coefficients for these violations. {These parameters} have to be tuned before using the algorithm. Since the primary aim of our TS heuristic is to find a set of feasible solutions, we {use} large values for $\nu^{MLR}$ and $\nu^{NV}$. 

Algorithm \ref{Alg:Tabu}, {which uses Algorithm~A.1 as a subprocedure,} presents the steps of our TS heuristic. In {the description of} Algorithm \ref{Alg:Tabu}, we denote the best solution (e.g., a solution with the smallest fitness value) found so far by $s^{BS}$, and the best feasible solution by $s^{BFS}$. We define $s^{IN}$ as the incumbent solution and $\mathcal{S}^{CA}$ as the set of candidate solutions for the next incumbent solution; {this set consists of neighbor solutions} of $s^{IN}$ generated by the \textit{Relocate} move. A move $(i, k)$ is defined by the node $i$ that is relocated from its current route to the best position in the destination route $k$, {and by the index~$k$ of the destination route.}
Algorithm~\ref{Alg:Tabu} presents the pseudocode of our Tabu Search algorithm. 
\begin{algorithm}[H]
\srcsize
\caption{Tabu search algorithm}
\label{Alg:Tabu}
\begin{algorithmic}
\STATE Input: Algorithm parameters, Initial solution $s$.
\STATE Output: A set of feasible solutions, $\mathcal{S}^{F}$
\STATE $s^{BS}, s^{IN} \leftarrow s$ 
\STATE $it \leftarrow 0$
\STATE $\mathcal{S}^{F} =  \emptyset$
\WHILE {$it \leq it^{NO}$}
\STATE  Create set $V^{S}$ which contains $\eta$ nodes randomly selected from $V_c$.
\FOR{ $i$ in $V^{S}$}
\STATE Determine set $V^{near}(i)$ as the set of $n^{near}$ nearest nodes to node $i$
\STATE Define $K^{near}(i)$ as the set of routes visit at least one node in $V^{near}(i)$
\FOR {route $k$ in $K^{near}(i)$}
\STATE \textit{Creating neighbor solution $s'$}
\STATE Insert node $i$ in the best position in route $k$
\STATE Calculate delivery quantities for the updated routes using Algorithm \ref{Alg:Optimal_quantity}
\STATE Calculate $F(s')$
\IF {move ($i$,$k$) is not in Tabu list}
\STATE Add $s'$ to $\mathcal{S}^{CA}$ set
\ELSIF{$F(s') < F(s^{BS})$ or ($s'$ is feasible and $F(s') < F(s^{BFS})$) }
\STATE Add $s'$ to $\mathcal{S}^{CA}$ set
\ENDIF
\ENDFOR
\ENDFOR
\STATE $s_{min}$ $\leftarrow$ the solution with minimum fitness in $\mathcal{S}^{CA}$
\IF {$F(s_{min}) < F(s^{BS})$}
\STATE $s^{BS} \leftarrow s_{min}$
\ENDIF
\STATE $s^{F}_{min}$ $\leftarrow$ the feasible solution with minimum fitness in $\mathcal{S}^{CA}$
\IF {$F(s^{F}_{min}) < F(s^{BFS})$}
\STATE $s^{BFS} \leftarrow s^{F}_{min}$ 
\STATE $\mathcal{S}^{F} =  \mathcal{S}^{F} \cup \{s^{BFS}\}$
\ENDIF
\IF {$F(s_{min}) \leq F(s^{IN})$}
\STATE $s^{IN} \leftarrow s_{min}$
\STATE $it  = 0$
\STATE Generate $\omega_{(i, k)}$ randomly between $\omega_{min}$ and $\omega_{max}$
\STATE Keep the move ($i, k$) for $\omega_{(i, k)}$ in Tabu List. 
\ELSE 
\STATE $it = it + 1$
\ENDIF
\STATE Remove the moves that stayed more than $\omega_{(i, k)}$ iterations in Tabu List.  
\ENDWHILE
\end{algorithmic}
\end{algorithm} 

For each feasible solution in $\mathcal{S}^{F}$, we run the LKH heuristic on each vehicle route separately as a post-optimization step. 
Table~\ref{Table:Tabu} {lists the parameters of} the algorithm {together with their values as used in the experimental part.}

\begin{table}[]
\center
\Tablesize
\caption{{Parameters for the} Tabu Search heuristic}
\label{Table:Tabu}
\begin{tabular}{@{}lll@{}}
\toprule
Parameter & Description & Value \\ \midrule
 $\eta$  & Number of nodes selected for relocation in each iteration & $\ceil{\frac{n}{5}}$ \\ 
  $n^{near}$ & Number of nodes considered near to a node &  $\ceil{\frac{n}{3}}$\\ 
 $[\omega_{min} , \omega_{max}]$ & \begin{tabular}{@{}l@{}} Bounds on the number of iterations \\ \qquad for which a move is prohibited \end{tabular} & $[2,6]$ \\
$it^{NO}$ & \begin{tabular}{@{}l@{}} Maximum number of consecutive iterations \\ \qquad without improvement before termination \end{tabular} & 20 \\
$\nu^{MLR}$  & Coefficient of maximum route length violation in fitness function& $10 \max\limits_{(i,j) \in A} t_{i,j}$\\
$\nu^{NV}$  & Coefficient of  maximum number of vehicles violation in fitness function & $10 \max\limits_{(i,j) \in A} t_{i,j}$\\
\bottomrule
\end{tabular}
\end{table}

\section{Van earthquake data preparation}\label{Sec:Data_gen_Van}
As our first case study, we {build on real-world} data from the 2011 Van earthquake, made available to us by \cite{noyan2015stochastic}. {We suitably adapted the data to our problem formulation in order to obtain test instances.} There are 94 settlements that have been {affected} by the earthquake; one local distribution center {was responsible for distributing} the relief items immediately after the disaster. \cite{noyan2015stochastic} studied a problem {of deciding on the location} of the relief aid distribution points (PODs) in a limited number of districts, {on} their capacity, {on the allocation of} supply to these distribution centers from the LDC, and {on the} assignment of demand points to PODs as a preparedness {measure}.


In our study, we focus on relief aid distribution in the post-disaster phase where the demands and {the (possibly damaged) state of the} road network are known. Therefore, instead of constructing different scenarios, we generate instances with different characteristics and solve our relief routing problem for each instance separately. To generate demands, we exactly follow the approach of \cite{noyan2015stochastic}. However, to generate realistic post-disaster distances between road network nodes, {taking possible damages into account,} we {apply} the network sampling algorithm described in \cite{yucel2018improving}, {which has also been used in} \cite{mostajabdaveh2018inequity} {(see below).}
The network has 94 demand points which can also act as intermediate points {for the routes}, and one LDC, the depot {where} all prepositioned supply {is stored}. The vehicles {start and end their routes} at the depot.
{In addition to the} 94 demand points, we added 14 road junction points and {built} graph $G = (V,E)$, where $V$ is the set of all nodes (junctions, demand points and LDC) and $E$ is the set of edges that connect nodes. We assume that there is a direct edge between every two nodes that have a distance less than 3 kilometres. The traveling times on the edges were calculated by the ArcGIS Pro $2.3.3$ software. 

To consider post disaster road damage when generating distances, we use the link failure dependency model presented in \cite{yucel2018improving}. We {partition the set of edges} into three sets $E_d$, $E_a$, and $E_s$, according to the damage intensity category of their endpoints. The edges that have one or two endpoints with damage intensity \textit{destructive} are placed in set $E_d$. From the remaining edges, {those} having one or two endpoints with intensity \textit{damaging} are assigned to set $E_a$. Set $E_s$ contains the edges {that do not belong to either of the two other sets}. Each edge $e \in E$ has a survival probability {of} $p_e$, which is the probability that the edge remains intact after the disaster. For each of our sample instances, we assign the initial $p_e$ values to the edges in the sets $E_d$, $E_a$ and $E_s$ by {uniformly} selecting a {random} value from the interval $[0.7, 0.8]$, $[0.8, 0.9]$ and $[0.9, 1]$, respectively. The link failure dependency model determines which edges are going to fail and which will remain functional in the post-disaster network, {introducing local correlations between the failure events.} We use this information to calculate the shortest paths between the demand nodes and the depot {in the damaged post-disaster network.}

In order to reduce the size of the problem,  \cite{noyan2015stochastic} clustered the demand nodes and created new networks with 15, 30, 60 nodes. The demand {within} each {cluster is} aggregated and assigned to {a selected node inside the cluster, cluster centriod}, which represents the cluster. Relief distribution in \cite{noyan2015stochastic} was only considered {in the form of distribution of} the supply from the depot to selected PODs. However, here we consider the case {where the supply has to be transported} directly from the depot to the demand points. It is {sufficient for practice} to solve the problem {only on the} 15-, 30- and 60-node-networks instead of the original 94-node-network since the {beneficiaries} can reach {the} cluster centroids easily. We set the number of vehicles equal to 3, 5 and 9 for instances with 15, 30 and 60 shelters, respectively. 
{To calculate the distances between two nodes in the reduced networks, we first generate the distances in the original network with 94 nodes according to the procedure described above. Then we only keep the edges that their endpoints are the clusters' centroids.}

{In total}, we generate {60} random instances, 20 for each of the problem sizes. We divide the number of nodes by the number of vehicles and multiply it by the average traveling time between any two nodes to obtain the maximum route length $\Psi$, such that $\Psi= \zeta^1 \ceil{n/m} \cdot \bar{t}$ where $\bar{t}= \sum\limits^n_{i=1}\sum\limits^n_{j=1}t_{i,j}/n(n-1)$ and $\zeta^1$ is a coefficient that controls maximum route length. 

For instances where we aim at generating a {regular} maximum route length, we {set}  $\zeta^1=2$, {while} for instances with {very tight} maximum route length, we set $\zeta^1=1$. 
We assume that the total available supply at the depot is equal to {70 \%} of the total demand, {i.e.,} $C=\zeta^2 \sum_id_i $
{with $\zeta^2=0.7$.}
The capacity of a vehicle is calculated as $Q=\zeta^3 \frac{C}{m} $ where 
for a {tight, but not very tight} capacity, we select $\zeta^3=0.5$, while for a very tight capacity, we set $\zeta^3=0.2$.  
The {purpose of picking} comparably tight capacities is to {observe} the performance of the algorithm under {stress, considering that generous vehicle capacities make the problem computationally easy}. However, for the sake of completeness, we generate a few instances with abundant vehicle capacity, where $\zeta^3 = 1.2$. Finally, the value $\epsilon$ is fixed to $0.85 m \cdot\Psi$.

\section{Detailed results of Van data set}\label{appx:Van_results}

Tables \ref{Table:Van_results15},  \ref{Table:Van_results30} and  \ref{Table:Van_results60} report the computational results obtained by {executing} both {the MIP-solver for} the VF model and the B\&P algorithm on these instances. 
The first column in each of these tables shows the instance name. 
The first number in instances names indicates the number of shelters that are to be visited by the vehicles. The second number indicates the instance identifier within each group of instances. The letters after ``\_" in the instance name indicate the instance type. {\mm The next column presents the computational time for $\theta$ values using the column generation approach, as described in Section 5.}
The next four columns in Table \ref{Table:Van_results15} present the results obtained by solving the VF model with the MIP-solver. The first two columns, \textit{UB} and \textit{LB}, report the best feasible solution value and the lower bound found until termination, respectively. The column \textit{Time} shows the execution time (in seconds) of the MIP solver. The percentage gap between the lower bound and the upper bound (i.e., the optimality gap) is reported in the column~\textit{Gap}. The next four columns under \textit{B$\&$P} provide analogous information to the four columns for \textit{Model}, but now refer to our B\&P algorithm. {\mm For the B$\&$P algorithm, we also report the number of branch-and-bound tree nodes explored before termination under \textit{\# Nodes}, the average number of columns in the column set at each node under \textit{Avg. Columns}, and the total number of columns with negative reduced cost generated by \textit{MIP} and \textit{GRASP} when solving the pricing problem.} Finally, the last column, \textit{LB-Gap}, represents the percentage gap between the lower bound reported by the model solver and that of the B$\&$P algorithm. Note that a positive value for \textit{LB-Gap} indicates that the B$\&$P algorithm was able to find a better lower bound than the MIP solver.

\begin{landscape}

\begin{table}[]
\mm
\Tablesize
\centering
\caption{Van earthquake case study results on the 15 node-network with 3 vehicles}
\vspace{-8pt}
\label{Table:Van_results15}
\begin{tabular}{c|c|cccc|cccccccc|c}
\hline
\multirow{2}{*}{Instance} & \multirow{2}{*}{\begin{tabular}[c]{@{}c@{}}Time\\ to $\theta$ \end{tabular}} & \multicolumn{4}{c|}{Model} & \multicolumn{8}{c|}{B\&P} & \multirow{2}{*}{LB-Gap\%} \\
 &  & UB & LB & Gap\% & Time & UB & LB & Gap\% & Time & \begin{tabular}[c]{@{}c@{}}\#\\ Nodes\end{tabular} & \begin{tabular}[c]{@{}c@{}}Avg.\\ Columns\end{tabular} & MIP & GRASP &  \\ \hline
Van15\_A1 & 17.8 & 111016.0 & 111015.3 & 0.000 & 0.7 & 111015.8 & 111015.7 & 0.000 & 14.0 & 1 & 77 & 44 & 30 & 0.000 \\
Van15\_A2 & 15.1 & 112810.6 & 112809.7 & 0.000 & 0.5 & 112812.1 & 112809.8 & 0.002 & 13.7 & 1 & 93 & 41 & 49 & 0.000 \\
Van15\_A3 & 17.7 & 101100.7 & 101100.0 & 0.000 & 0.3 & 101100.4 & 101100.4 & 0.000 & 13.9 & 1 & 89 & 42 & 44 & 0.000 \\
Van15\_A4 & 16.0 & 102262.4 & 102261.9 & 0.000 & 0.7 & 102264.7 & 102262.3 & 0.002 & 13.2 & 1 & 79 & 41 & 35 & 0.000 \\
Van15\_A5 & 16.9 & 101806.8 & 101805.9 & 0.000 & 0.9 & 101808.5 & 101806.1 & 0.002 & 13.7 & 1 & 80 & 41 & 36 & 0.000 \\
Van15\_T1 & 13.3 & 142638.7 & 142631.3 & 0.005 & 1.7 & 142633.2 & 142633.1 & 0.000 & 13.7 & 1 & 99 & 46 & 50 & 0.001 \\
Van15\_T2 & 21.1 & 143815.7 & 143809.5 & 0.004 & 0.3 & 143810.4 & 143810.4 & 0.000 & 16.9 & 1 & 93 & 48 & 42 & 0.001 \\
Van15\_T3 & 17.0 & 139494.3 & 139487.4 & 0.005 & 0.3 & 139490.4 & 139489.1 & 0.001 & 12.3 & 1 & 107 & 48 & 56 & 0.001 \\
Van15\_T4 & 17.1 & 140654.7 & 140646.7 & 0.006 & 0.2 & 140649.1 & 140649.1 & 0.000 & 15.2 & 1 & 106 & 33 & 70 & 0.002 \\
Van15\_T5 & 15.1 & 140200.7 & 140192.4 & 0.006 & 0.2 & 140194.8 & 140194.8 & 0.000 & 13.5 & 1 & 107 & 43 & 61 & 0.002 \\
Van15\_VT1 & 19.5 & 157263.6 & 152717.0 & 2.977 & 7200.1 & 157274.8 & 157261.9 & 0.008 & 47.9 & 3 & 294 & 125 & 238 & 2.976 \\
Van15\_VT2 & 18.3 & 169210.6 & 165388.9 & 2.311 & 7200.1 & 169221.8 & 169208.1 & 0.008 & 73.0 & 5 & 332 & 168 & 308 & 2.309 \\
Van15\_VT3 & 19.6 & 155805.1 & 151188.6 & 3.053 & 7200.1 & 155808.5 & 155805.0 & 0.002 & 47.3 & 3 & 282 & 108 & 269 & 3.053 \\
Van15\_VT4 & 14.0 & 159072.6 & 155027.0 & 2.610 & 7200.8 & 159082.3 & 159071.0 & 0.007 & 66.9 & 5 & 291 & 161 & 254 & 2.609 \\
Van15\_VT5 & 14.3 & 163207.4 & 159160.9 & 2.542 & 7200.8 & 163216.2 & 163204.9 & 0.007 & 39.6 & 5 & 349 & 118 & 301 & 2.541 \\
Van15\_VTL1 & 23.9 & 152773.2 & 148468.4 & 2.899 & 7200.1 & 152785.2 & 152773.2 & 0.008 & 134.1 & 5 & 259 & 149 & 299 & 2.899 \\
Van15\_VTL2 & 14.3 & 168460.0 & 164469.6 & 2.426 & 7200.1 & 168472.8 & 168459.1 & 0.008 & 32.4 & 11 & 193 & 272 & 419 & 2.426 \\
Van15\_VTL3 & 11.0 & 159873.3 & 155736.2 & 2.656 & 7200.1 & 159883.6 & 159873.0 & 0.007 & 126.3 & 15 & 252 & 470 & 436 & 2.656 \\
Van15\_VTL4 & 38.3 & 162461.3 & 158327.8 & 2.611 & 7200.1 & 162464.0 & 162460.7 & 0.002 & 112.0 & 3 & 257 & 110 & 254 & 2.610 \\
Van15\_VTL5 & 59.8 & 160383.7 & 155922.3 & 2.861 & 7200.1 & 160388.5 & 160380.9 & 0.005 & 141.9 & 3 & 263 & 109 & 216 & 2.859 \\ \hline
Average & 20.0 &  &  & 1.349 & 3600.4 &  &  & 0.003 & 48.1 & 3 & 185 & 111 & 173 & 1.347
\end{tabular}
\end{table}

\begin{table}[H]
\mm
\Tablesize
\centering
\caption{Van earthquake case study results on the 30 node-network with 5 vehicles}
\vspace{-8pt}
\label{Table:Van_results30}
\begin{tabular}{c|c|cccc|cccccccc|c}
\hline
\multirow{2}{*}{Instance} & \multirow{2}{*}{\begin{tabular}[c]{@{}c@{}}Time\\ to $\theta$ \end{tabular}} & \multicolumn{4}{c|}{Model} & \multicolumn{8}{c|}{B\&P} & \multirow{2}{*}{LB-Gap\%} \\
 &  & UB & LB & Gap\% & Time & UB & LB & Gap\% & Time & \begin{tabular}[c]{@{}c@{}}\# \\ Nodes\end{tabular} & \begin{tabular}[c]{@{}c@{}}Avg.\\ Columns\end{tabular} & MIP & GRASP &  \\ \hline
Van30\_A1 & 423.0 & 130565.3 & 130564.5 & 0.000 & 1.3 & 130569.0 & 130564.9 & 0.003 & 341.0 & 1 & 206 & 87 & 114 & 0.000 \\
Van30\_A2 & 818.2 & 134712.9 & 134712.1 & 0.000 & 1.2 & 134717.8 & 134712.4 & 0.004 & 596.7 & 1 & 187 & 93 & 89 & 0.000 \\
Van30\_A3 & 526.1 & 134699.4 & 134698.9 & 0.000 & 3.4 & 134705.1 & 134699.2 & 0.004 & 402.2 & 1 & 193 & 93 & 95 & 0.000 \\
Van30\_A4 & 409.7 & 132003.8 & 132003.1 & 0.000 & 0.8 & 132010.0 & 132003.4 & 0.005 & 406.3 & 1 & 200 & 82 & 113 & 0.000 \\
Van30\_A5 & 230.5 & 135583.7 & 135582.9 & 0.000 & 3.9 & 135588.9 & 135583.2 & 0.004 & 195.8 & 1 & 103 & 1 & 97 & 0.000 \\
Van30\_T1 & 411.3 & 145083.0 & 145077.0 & 0.004 & 2.1 & 145086.1 & 145080.3 & 0.004 & 344.6 & 1 & 272 & 75 & 192 & 0.002 \\
Van30\_T2 & 251.6 & 156988.1 & 156979.9 & 0.005 & 2.6 & 156989.8 & 156983.8 & 0.004 & 231.2 & 1 & 299 & 94 & 200 & 0.002 \\
Van30\_T3 & 311.8 & 155593.5 & 155592.0 & 0.001 & 7.6 & 155598.6 & 155592.9 & 0.004 & 287.2 & 1 & 322 & 63 & 254 & 0.001 \\
Van30\_T4 & 274.0 & 148049.9 & 148043.4 & 0.004 & 2.3 & 148050.9 & 148044.9 & 0.004 & 255.5 & 1 & 281 & 91 & 185 & 0.001 \\
Van30\_T5 & 546.2 & 159615.6 & 159609.4 & 0.004 & 2.2 & 159617.2 & 159610.8 & 0.004 & 408.2 & 1 & 250 & 87 & 158 & 0.001 \\
Van30\_VT1 & 872.4 & 170444.3 & 166837.1 & 2.162 & 7201.1 & 170451.7 & 170432.4 & 0.011 & 7223.7 & 11 & 1133 & 407 & 1470 & 2.155 \\
Van30\_VT2 & 594.4 & 166735.1 & 163137.0 & 2.206 & 7201.0 & 166720.7 & 166702.9 & 0.011 & 7217.1 & 15 & 826 & 456 & 1455 & 2.186 \\
Van30\_VT3 & 486.6 & 157852.6 & 153696.6 & 2.704 & 7201.3 & 157796.7 & 157773.9 & 0.014 & 7205.1 & 19 & 895 & 398 & 1368 & 2.653 \\
Van30\_VT4 & 574.5 & 165529.6 & 161425.1 & 2.543 & 7201.0 & 165538.8 & 165521.4 & 0.011 & 7219.5 & 17 & 804 & 466 & 1803 & 2.538 \\
Van30\_VT5 & 169.6 & 165752.5 & 161929.7 & 2.361 & 7200.9 & 165689.2 & 165673.6 & 0.009 & 2162.8 & 15 & 903 & 421 & 1523 & 2.312 \\
Van30\_VTL1 & 319.4 & 160276.4 & 156403.4 & 2.476 & 7200.4 & 160264.0 & 160251.0 & 0.008 & 6751.0 & 23 & 344 & 436 & 1110 & 2.460 \\
Van30\_VTL2 & 478.0 & 158186.7 & 154002.8 & 2.717 & 7200.9 & 158203.3 & 158179.8 & 0.015 & 7210.6 & 19 & 404 & 352 & 921 & 2.712 \\
Van30\_VTL3 & 519.2 & 169682.4 & 166147.2 & 2.128 & 7201.3 & 169681.2 & 169666.0 & 0.009 & 1880.3 & 5 & 458 & 19 & 650 & 2.118 \\
Van30\_VTL4 & 478.6 & 165363.0 & 161384.4 & 2.465 & 7200.9 & 165355.8 & 165335.9 & 0.012 & 7209.8 & 17 & 467 & 215 & 1230 & 2.449 \\
Van30\_VTL5 & 391.6 & 168168.6 & 164576.7 & 2.182 & 7201.2 & 168079.1 & 168059.7 & 0.012 & 7212.2 & 25 & 600 & 359 & 1104 & 2.116 \\ \hline
Average & 454.3 &  &  & 1.349 & 3600.4 &  &  & 0.003 & 48.1 & 9 & 457 & 215 & 707 & 1.185
\end{tabular}
\end{table}

\begin{table}[H]
\mm
\scriptsize
\centering
\caption{Van earthquake case study results on the 60 node-network with 9 vehicles}
\vspace{-8pt}
\label{Table:Van_results60}
\begin{tabular}{c|c|cccc|cccccccc|c}
\hline
\multirow{2}{*}{Instance} & \multirow{2}{*}{\begin{tabular}[c]{@{}c@{}}Time\\ to $\theta$\end{tabular}} & \multicolumn{4}{c|}{Model} & \multicolumn{8}{c|}{B\&P} & \multirow{2}{*}{LB-Gap\%} \\
 &  & UB & LB & Gap\% & Time & UB & LB & Gap\% & Time & \begin{tabular}[c]{@{}c@{}}\#\\ Nodes\end{tabular} & \begin{tabular}[c]{@{}c@{}}Avg. \\ Columns\end{tabular} & MIP & GRASP &  \\ \hline
Van60\_A1 & 222.7 & 145093.6 & 145092.3 & 0.000 & 20.7 & 145105.6 & 145093.1 & 0.009 & 1979.1 & 9 & 577 & 225 & 452 & -0.007 \\
Van60\_A2 & 481.4 & 146713.6 & 146712.1 & 0.000 & 32.0 & 146726.0 & 146712.7 & 0.009 & 3280.6 & 9 & 558 & 244 & 463 & -0.001 \\
Van60\_A3 & 1026.3 & 144145.9 & 144144.6 & 0.000 & 53.1 & 144158.9 & 144145.3 & 0.009 & 3107.0 & 3 & 578 & 206 & 388 & 0.000 \\
Van60\_A4 & 955.8 & 145964.1 & 145962.9 & 0.000 & 13.1 & 145977.3 & 145963.5 & 0.009 & 5008.0 & 7 & 450 & 235 & 395 & -0.004 \\
Van60\_A5 & 354.9 & 141773.5 & 141772.2 & 0.000 & 19.8 & 141778.9 & 141772.2 & 0.005 & 3644.6 & 11 & 483 & 254 & 421 & -0.003 \\
Van60\_T1 & 1297.6 & 161777.2 & 161775.4 & 0.001 & 496.1 & 161783.8 & 161777.2 & 0.004 & 1225.6 & 1 & 541 & 72 & 460 & -0.005 \\
Van60\_T2 & 1475.6 & 163284.9 & 163216.0 & 0.042 & 7201.2 & 163225.5 & 163217.6 & 0.005 & 1063.0 & 1 & 527 & 117 & 401 & -0.001 \\
Van60\_T3 & 1166.2 & 153801.8 & 153800.9 & 0.001 & 780.0 & 153810.3 & 153801.6 & 0.006 & 1089.4 & 1 & 615 & 117 & 489 & -0.002 \\
Van60\_T4 & 1355.2 & 150207.4 & 150143.1 & 0.043 & 7201.6 & 150150.3 & 150143.5 & 0.005 & 1176.8 & 1 & 603 & 108 & 486 & -0.003 \\
Van60\_T5 & 1581.0 & 163182.1 & 163180.3 & 0.001 & 2779.5 & 163190.1 & 163182.1 & 0.005 & 1421.8 & 1 & 580 & 139 & 432 & -0.001 \\
Van60\_VT1 & 2521.4 & 166162.4 & 163186.4 & 1.824 & 7201.6 & 166040.5 & 166015.1 & 0.015 & 7205.4 & 3 & 4414 & 6 & 4399 & 0.240 \\
Van60\_VT2 & 1573.5 & 173435.8 & 170810.3 & 1.537 & 7202.3 & 173379.0 & 173358.7 & 0.012 & 7308.6 & 5 & 2176 & 10 & 4060 & 2.112 \\
Van60\_VT3 & 3014.7 & 171395.8 & 168632.3 & 1.639 & 7202.6 & 171253.0 & 171233.7 & 0.011 & 7208.7 & 3 & 4572 & 12 & 4551 & 4.535 \\
Van60\_VT4 & 3299.7 & 164363.2 & 161348.3 & 1.869 & 7202.3 & 164219.5 & 164191.1 & 0.017 & 7206.6 & 3 & 4887 & 4 & 4874 & 12.765 \\
Van60\_VT5 & 3081.7 & 169090.9 & 166109.3 & 1.795 & 7201.6 & 168899.4 & 168878.3 & 0.012 & 7220.1 & 3 & 4484 & 10 & 4465 & 3.511 \\
Van60\_VTL1 & 84.5 & 171590.5 & 164652.0 & 4.214 & 7201.2 & 171307.4 & 171291.7 & 0.009 & 766.3 & 11 & 799 & 18 & 1268 & 10.334 \\
Van60\_VTL2 & 825.9 & 176006.9 & 168925.1 & 4.192 & 7202.7 & 175768.4 & 175745.9 & 0.013 & 7265.2 & 11 & 919 & 22 & 1297 & 6.600 \\
Van60\_VTL3 & 128.3 & 173486.7 & 166606.7 & 4.129 & 7201.8 & 173249.9 & 173238.4 & 0.007 & 1294.3 & 11 & 871 & 29 & 1766 & 6.670 \\
Van60\_VTL4 & 1900.3 & 164269.1 & 157480.2 & 4.311 & 7201.4 & 163609.4 & 163585.6 & 0.015 & 7282.3 & 5 & 852 & 12 & 1026 & 8.093 \\
Van60\_VTL5 & 239.0 & 161385.5 & 154271.8 & 4.611 & 7201.1 & 161078.0 & 161063.9 & 0.009 & 1939.5 & 11 & 765 & 16 & 1279 & 0.803 \\ \hline
Average & 1329.3 &  &  & 1.510 & 4530.8 &  &  & 0.009 & 3884.6 & 6 & 1512 & 88 & 1669 & 8.201
\end{tabular}
\end{table}

\end{landscape}

\section{Kartal case study data preparation}\label{Sec:Data_gen_Kartal}
{To} make {the} Kartal case study data compatible with our problem {formulation}, 
we adopted the solution of an instance with 2000 scenarios, presented in \cite{mostajabdaveh2018inequity}, to 
determine shelter locations. Twelve locations {were} selected for shelter establishment. The depot location {was} selected based on the real location of a relief supply distribution center in the region. We generated 40 instances with different parameter {values}. For each instance,
we randomly selected one scenario {of} post-disaster demands and {of post-disaster} distances. We assumed that all individuals assigned to a shelter will reside at that shelter after the disaster. Therefore, {the demand occurring in a shelter was supposed to be} equal to the sum of {the demands in the} assigned population nodes. {Furthermore, we assumed} without loss of generality {that} each individual needs one unit of the relief supply.
We supposed that $m = 3$ identical vehicles are available to distribute the relief supply among the shelter locations.
 
We calculated $\Psi$ by an approach {analogous} to the one in Section \ref{Sec:Data_gen_Van}. 
The total available supply at the depot {was set} equal to $C=\zeta^2 \sum_id_i $, where $\zeta^2=0.85$. The vehicle capacity 
was calculated by the same  
procedure {as used} in Section~\ref{Sec:Data_gen_Van} for the Van instances. Finally, the value for $\epsilon$ {was} fixed to $0.85 m \times\Psi$.

\section{Detailed results on Kartal data set}\label{appx:Kartal_results}
Table \ref{Table:KartalP2} reports the computational results provided by both solving the VF model and B\&P algorithm on {the Kartal} instances. {The meaning of the columns is the same as in Table \ref{Table:Van_results15}.}

\begin{landscape}
\begin{table}[H]
\mm
\scriptsize
\centering
\caption{Kartal case study results with 13 nodes and 3 vehicles}
\vspace{-8pt}
\label{Table:KartalP2}
\begin{tabular}{c|c|cccc|cccccccc|c}
\hline
\multirow{2}{*}{Instance} & \multirow{2}{*}{\begin{tabular}[c]{@{}c@{}}Time\\ to $\theta$ \end{tabular}} & \multicolumn{4}{c|}{Model} & \multicolumn{8}{c|}{B\&P} & \multirow{2}{*}{LB-Gap\%} \\
 &  & UB & LB & Gap\% & Time & UB & LB & Gap\% & Time & \begin{tabular}[c]{@{}c@{}}\# \\ Nodes\end{tabular} & \begin{tabular}[c]{@{}c@{}}Avg.\\ Columns\end{tabular} & MIP & GRASP &  \\ \hline
Kartal\_A1 & 2.0 & 14031.3 & 14031.2 & 0.001 & 0.1 & 14031.3 & 14030.2 & 0.008 & 42.1 & 17 & 142 & 384 & 166 & -0.007 \\
Kartal\_A2 & 3.3 & 33521.2 & 33521.0 & 0.001 & 0.8 & 33521.8 & 33520.7 & 0.003 & 257.2 & 49 & 79 & 449 & 522 & -0.001 \\
Kartal\_A3 & 2.8 & 24921.9 & 24921.9 & 0.000 & 0.2 & 24922.9 & 24921.8 & 0.004 & 707.8 & 199 & 39 & 936 & 1050 & 0.000 \\
Kartal\_A4 & 1.9 & 14107.9 & 14107.9 & 0.000 & 0.2 & 14107.9 & 14107.3 & 0.004 & 193.0 & 81 & 58 & 879 & 682 & -0.004 \\
Kartal\_A5 & 5.4 & 27867.6 & 27867.4 & 0.001 & 0.2 & 27867.6 & 27866.6 & 0.004 & 81.2 & 11 & 154 & 315 & 108 & -0.003 \\
Kartal\_A6 & 2.6 & 11272.9 & 11272.9 & 0.000 & 0.1 & 11273.3 & 11272.4 & 0.008 & 164.3 & 57 & 39 & 423 & 245 & -0.005 \\
Kartal\_A7 & 2.1 & 37870.2 & 37870.2 & 0.000 & 0.1 & 37870.2 & 37869.8 & 0.001 & 254.4 & 89 & 82 & 879 & 705 & -0.001 \\
Kartal\_A8 & 5.0 & 21379.2 & 21379.2 & 0.000 & 0.1 & 21379.2 & 21378.7 & 0.002 & 347.6 & 63 & 78 & 830 & 605 & -0.002 \\
Kartal\_A9 & 2.9 & 21457.9 & 21457.9 & 0.000 & 0.1 & 21457.9 & 21457.4 & 0.003 & 243.1 & 75 & 99 & 941 & 545 & -0.003 \\
Kartal\_A10 & 3.8 & 41481.2 & 41481.2 & 0.000 & 0.1 & 41483.3 & 41480.8 & 0.006 & 214.9 & 37 & 132 & 572 & 337 & -0.001 \\
Kartal\_T1 & 18.1 & 52538.2 & 52412.4 & 0.240 & 7200.0 & 52538.2 & 52538.2 & 0.000 & 29.8 & 1 & 209 & 69 & 137 & 0.240 \\
Kartal\_T2 & 38.4 & 94017.4 & 92072.9 & 2.068 & 7200.0 & 94017.4 & 94017.4 & 0.000 & 40.6 & 1 & 428 & 129 & 296 & 2.112 \\
Kartal\_T3 & 16.0 & 57949.4 & 55433.9 & 4.341 & 7200.0 & 57949.5 & 57947.7 & 0.003 & 25.0 & 1 & 172 & 54 & 115 & 4.535 \\
Kartal\_T4 & 2.6 & 71345.4 & 63263.1 & 11.328 & 7200.0 & 71345.4 & 71338.6 & 0.009 & 2177.7 & 651 & 61 & 6521 & 4987 & 12.765 \\
Kartal\_T5 & 68.3 & 32889.2 & 31772.6 & 3.395 & 7200.0 & 32889.2 & 32888.0 & 0.004 & 84.3 & 1 & 280 & 59 & 218 & 3.511 \\
Kartal\_T6 & 1.7 & 97702.3 & 88543.0 & 9.375 & 7200.0 & 97702.3 & 97693.0 & 0.009 & 6773.6 & 2425 & 54 & 15915 & 14183 & 10.334 \\
Kartal\_T7 & 48.6 & 69758.8 & 65439.8 & 6.191 & 7200.0 & 69758.8 & 69758.8 & 0.000 & 60.2 & 1 & 842 & 161 & 678 & 6.600 \\
Kartal\_T8 & 31.5 & 52495.8 & 49213.4 & 6.253 & 7200.0 & 52495.8 & 52495.8 & 0.000 & 39.8 & 1 & 302 & 61 & 238 & 6.670 \\
Kartal\_T9 & 38.3 & 47349.3 & 43804.1 & 7.487 & 7200.0 & 47349.3 & 47349.2 & 0.000 & 49.0 & 1 & 259 & 51 & 205 & 8.093 \\
Kartal\_T10 & 22.8 & 67350.6 & 66814.0 & 0.797 & 7200.0 & 67350.6 & 67350.6 & 0.000 & 36.3 & 1 & 300 & 75 & 222 & 0.803 \\
Kartal\_VT1 & 3.4 & 32690.5 & 30210.6 & 7.586 & 7200.0 & 32690.5 & 32688.0 & 0.008 & 513.9 & 135 & 87 & 1415 & 836 & 8.201 \\
Kartal\_VT2 & 3.3 & 58758.2 & 54544.3 & 7.172 & 7200.0 & 58759.3 & 58754.3 & 0.008 & 912.2 & 195 & 73 & 1866 & 1316 & 7.718 \\
Kartal\_VT3 & 8.5 & 108071.6 & 103984.4 & 3.782 & 7200.0 & 108071.6 & 108064.1 & 0.007 & 13.3 & 1 & 140 & 67 & 70 & 3.923 \\
Kartal\_VT4 & 64.0 & 81271.1 & 79932.0 & 1.648 & 7200.0 & 81271.1 & 81271.1 & 0.000 & 85.5 & 1 & 332 & 99 & 230 & 1.675 \\
Kartal\_VT5 & 4.7 & 117299.7 & 112925.2 & 3.729 & 7200.0 & 117301.4 & 117298.3 & 0.003 & 230.1 & 33 & 122 & 486 & 308 & 3.873 \\
Kartal\_VT6 & 2.9 & 36238.1 & 35791.2 & 1.233 & 7200.0 & 36231.3 & 36228.8 & 0.007 & 38.9 & 11 & 161 & 154 & 247 & 1.223 \\
Kartal\_VT7 & 53.3 & 104849.4 & 102241.7 & 2.487 & 7200.0 & 104851.0 & 104849.4 & 0.001 & 71.5 & 1 & 506 & 121 & 382 & 2.551 \\
Kartal\_VT8 & 27.7 & 83575.2 & 83075.1 & 0.598 & 7200.0 & 83576.3 & 83572.1 & 0.005 & 28.0 & 1 & 232 & 42 & 187 & 0.598 \\
Kartal\_VT9 & 1.5 & 102261.9 & 94445.4 & 7.644 & 7200.0 & 102261.9 & 102250.2 & 0.011 & 7204.8 & 3029 & 60 & 16708 & 15622 & 8.264 \\
Kartal\_VT10 & 52.5 & 86892.1 & 84750.2 & 2.465 & 7200.0 & 86892.3 & 86892.1 & 0.000 & 60.0 & 1 & 585 & 200 & 382 & 2.527 \\
Kartal\_VTL1 & 29.3 & 55568.7 & 51807.9 & 6.768 & 7200.0 & 55568.7 & 55563.5 & 0.009 & 515.1 & 143 & 114 & 1719 & 912 & 10.882 \\
Kartal\_VTL2 & 5.1 & 55870.0 & 54534.8 & 2.390 & 7200.0 & 55870.0 & 55865.6 & 0.008 & 40.8 & 5 & 147 & 69 & 105 & 2.440 \\
Kartal\_VTL3 & 4.3 & 57308.1 & 54867.7 & 4.258 & 7200.0 & 57309.0 & 57305.1 & 0.007 & 258.7 & 37 & 158 & 540 & 309 & 4.442 \\
Kartal\_VTL4 & 4.2 & 45769.0 & 41953.5 & 8.336 & 7200.0 & 45768.4 & 45764.6 & 0.008 & 1343.4 & 281 & 91 & 3120 & 1703 & 9.084 \\
Kartal\_VTL5 & 13.9 & 27559.5 & 26799.9 & 2.756 & 7200.0 & 27559.5 & 27559.3 & 0.001 & 16.6 & 1 & 136 & 35 & 98 & 2.834 \\
Kartal\_VTL6 & 3.3 & 97726.1 & 94801.9 & 2.992 & 7200.0 & 97726.5 & 97717.3 & 0.009 & 217.8 & 57 & 86 & 412 & 576 & 3.075 \\
Kartal\_VTL7 & 5.1 & 46543.1 & 43833.9 & 5.821 & 7200.0 & 46542.5 & 46540.7 & 0.004 & 822.2 & 117 & 81 & 1115 & 654 & 6.175 \\
Kartal\_VTL8 & 6.1 & 53445.4 & 47922.1 & 10.334 & 7200.0 & 53442.9 & 53437.8 & 0.009 & 419.2 & 55 & 117 & 791 & 409 & 11.510 \\
Kartal\_VTL9 & 18.3 & 42008.0 & 41110.8 & 2.136 & 7200.0 & 42008.0 & 42007.0 & 0.002 & 19.3 & 1 & 180 & 45 & 132 & 2.180 \\
Kartal\_VTL10 & 25.2 & 74876.3 & 72688.0 & 2.923 & 7200.0 & 74871.8 & 74871.8 & 0.000 & 38.9 & 1 & 286 & 56 & 227 & 3.004 \\ \hline
Average & 16.4 &  &  & 3.540 & 5400.1 &  &  & 0.005 & 720.3 & 197 & 188 & 1469 & 1274 & 3.795
\end{tabular}
\end{table}
\end{landscape}



\section{Algorithm components analysis}
In this section, we evaluate the impact of various components on our B\&P algorithm. The objective of this analysis is to determine the effects of the enhancements we developed and added to the traditional B\&P algorithm. Specifically, we aim to identify which component has played a pivotal role in the success of our method. For this analysis, we introduced three distinct variations of our B\&P algorithm. The first variation is the \textit{B\&P W/O valid inequalities}, where we removed valid inequalities (37) - (40) from the pricing mathematical model. The second variation is the \textit{B\&P W/O GRASP}, where we rely solely on the MIP of the pricing problem to find columns with negative reduced costs. The third variation is the \textit{B\&P W/O initial routes}, where we depend only on our version of the CW heuristic to generate initial columns and omit the Tabu search that was used to enhance these initial columns.

To conduct the analysis, we utilized our Van earthquake dataset, specifically using all instances of types T, VT, and VTL, with 15, 30, and 60 nodes. This selection was made because these instances represent the most challenging scenarios within the dataset. The time limit is set to two hours for each experiment. Tables \ref{tab:base_grasp} and \ref{tab:valid_tabu} aggregate the average performance metrics across instances of identical type. The evaluated metrics include the optimality gap, the number of nodes explored within the branch-and-bound tree, the average number of columns at each node, and the total computation time. The Full Version refers to our B\&P algorithm with all of its components, as described in Section 4.

\begin{table}[ht]
\small
\centering
\caption{Experiment results for the Full Version and B\&P W/O GRASP}
\label{tab:base_grasp}
\begin{tabular}{l|cccc|cccc}
 & \multicolumn{4}{c|}{Full Version} & \multicolumn{4}{c}{B\&P W/O GRASP} \\ \hline
Instance Type & Gap & \# Node & \begin{tabular}[c]{@{}c@{}}Avg.\\ columns\end{tabular} & Time & Gap & \# Node & \begin{tabular}[c]{@{}c@{}}Avg.\\ columns\end{tabular} & Time \\
\midrule
Van15\_T & 0.000 & 1 & 102 & 14.3 & 0.000 & 1 & 63 & 18.2 \\
Van15\_VT & 0.007 & 4 & 310 & 54.9 & 0.008 & 5 & 155 & 535.0 \\
Van15\_VTL & 0.006 & 7 & 245 & 109.3 & 0.006 & 9 & 124 & 251.7 \\
Van30\_T & 0.004 & 1 & 285 & 305.3 & 0.005 & 1 & 170 & 1271.2 \\
Van30\_VT & 0.011 & 15 & 912 & 6273.6 & 0.013 & 16 & 250 & 7308.3 \\
Van30\_VTL & 0.011 & 18 & 455 & 6064.8 & 0.017 & 18 & 206 & 7247.2 \\
Van60\_T & 0.005 & 1 & 573 & 1195.3 & 0.005 & 1 & 479 & 3044.5 \\
Van60\_TL & 0.014 & 3 & 4107 & 7229.9 & 0.008 & 3 & 231 & 3384.1 \\
Van60\_VTL & 0.011 & 10 & 841 & 3709.5 & 0.006 & 1 & 126 & 1703.9 \\  \hline
\multicolumn{1}{l|}{Average} & 0.008 & 7 & 870 & 2773.0 & 0.008 & 6 & 200 & 2751.5
\end{tabular}
\end{table}

\begin{table}[ht]
\small
\centering
\caption{Experiment results for the B\&P W/O valid inequalities and B\&P W/O initial routes}
\label{tab:valid_tabu}
\begin{tabular}{l|cccc|cccc}
 & \multicolumn{4}{c|}{B\&P W/O valid inequalities} & \multicolumn{4}{c}{B\&P W/O initial routes} \\ \hline
Instance Type & Gap & \# Node & \begin{tabular}[c]{@{}c@{}}Avg. \\ columns\end{tabular} & Time & Gap & \# Node & \begin{tabular}[c]{@{}c@{}}Avg.\\ columns\end{tabular} & Time \\
\midrule
Van15\_T & 0.003 & 1 & 102 & 16.5 & 0.000 & 1 & 131 & 31.3 \\
Van15\_VT & 0.009 & 8 & 235 & 92.3 & 0.006 & 10 & 335 & 290.3 \\
Van15\_VTL & 0.010 & 48 & 119 & 196.9 & 0.005 & 18 & 250 & 496.6 \\
Van30\_T & 0.005 & 1 & 271 & 495.7 & 0.004 & 10 & 244 & 2460.5 \\
Van30\_VT & 0.011 & 52 & 419 & 7179.6 & 0.226 & 15 & 801 & 7287.5 \\
Van30\_VTL & 0.010 & 57 & 374 & 3163.3 & 0.473 & 22 & 448 & 6372.5 \\
Van60\_T & 0.005 & 1 & 578 & 1693.7 & 0.875 & 13 & 586 & 7120.6 \\
Van60\_TL & 0.016 & 3 & 4465 & 7233.0 & 1.268 & 3 & 3426 & 7248.8 \\
Van60\_VTL &  0.013 & 11 & 1308 & 7211.5 & 1.401 & 9 & 1354 & 7240.9 \\ \hline
\multicolumn{1}{l|}{Average} & 0.009 & 20 & 875 & 3031.4 & 0.473 & 11 & 842 & 4283.2
\end{tabular}
\end{table}

For Table \ref{tab:valid_tabu}, its evident that the removing valid inequalities from pricing model leds to a significant increase in the number of B\&B nodes. This increase in node exploration has, in turn, resulted in a 9.3\% longer runtime for the algorithm. Observing the outcomes for the `B\&P W/O initial routes' variant, there is a noticeable increase in both the optimality gap and runtime relative to the B\&P Full version. This indicates that dedicating effort to identifying a strong initial solution significantly aids the convergence of the B\&P algorithm. 

The B\&P version without GRASP exhibited a decrease in the number of generated columns, aligning with expectations since GRASP is capable of generating up to 10 columns with negative reduced costs during each column generation iteration. As evident in the Table \ref{tab:base_grasp}, for smaller instances, employing GRASP has resulted in accelerated convergence. However, for larger instances of types Van60\_TL and Van60\_VTL, GRASP does not seem to be very effective in identifying beneficial new columns.

In summary, our findings lead us to conclude that the implementation of Tabu Search for enhancing the initial columns, coupled with the incorporation of valid inequalities in the pricing model, significantly contributes to the superior performance of our B\&P algorithm. Furthermore, employing the GRASP algorithm for column generation has proved to be an effective strategy for small and medium-sized instances. 

{
\section{Evaluation of solution quality with a limited runtime}

This section presents an evaluation of the solutions generated by our B\&P algorithm within a restricted runtime of 600 seconds. For this experiment, we selected 15 instances from the Van and Kartal datasets, specifically choosing those with original termination times exceeding 600 seconds to allow for the algorithm's premature termination.

Table~\ref{table:600sec} provides the results for each instance. Our findings indicate an average optimality gap of 0.023\% with a 600-second runtime, demonstrating that the algorithm effectively reaches high-quality solutions early in the search process. This observation suggests that the B\&P algorithm allocates most of its computation time to proving optimality, a behavior commonly observed in branch-and-bound methods.

\begin{table}[]
\small
\centering
\caption{Experiment results for B\&P with 600 seconds time limit}
\label{table:600sec}
\begin{tabular}{@{}l|cccccc@{}}
\toprule
\textbf{Instance} & \textbf{UB} & \textbf{LB} & \textbf{Gap\%} & \textbf{Time} & \textbf{\# Nodes} & \begin{tabular}[c]{@{}c@{}} \textbf{Avg.} \\ \textbf{columns} \end{tabular} \\ \midrule
Van30\_VT1 & 170471 & 170434 & 0.021 & 611.0 & 3 & 684 \\
Van30\_VT2 & 166738 & 166703 & 0.021 & 603.3 & 3 & 853 \\
Van30\_VTL1 & 160298 & 160251 & 0.029 & 638.3 & 3 & 586 \\
Van30\_VTL2 & 158229 & 158180 & 0.031 & 629.9 & 3 & 541 \\
Van60\_VT1 & 166054 & 166031 & 0.014 & 623.7 & 1 & 616 \\
Van60\_VT2 & 173390 & 173371 & 0.010 & 639.1 & 3 & 661 \\
Van60\_VTL1 & 171346 & 171292 & 0.032 & 624.7 & 3 & 678 \\
Van60\_VTL2 & 175799 & 175748 & 0.029 & 619.2 & 3 & 523 \\
Kartal\_A3 & 24929.5 & 24921.3 & 0.033 & 609.3 & 105 & 55 \\
Kartal\_T4 & 71350.5 & 71335.2 & 0.021 & 604.2 & 135 & 84 \\
Kartal\_T6 & 97709.5 & 97681.4 & 0.029 & 604.8 & 181 & 78 \\
Kartal\_VT2 & 58759.3 & 58753.2 & 0.010 & 603.8 & 177 & 82 \\
Kartal\_VT9 & 102273 & 102239 & 0.033 & 603.4 & 161 & 87 \\
Kartal\_VTL4 & 45774.7 & 45763.9 & 0.024 & 605.4 & 71 & 111 \\
Kartal\_VTL7 & 46544.9 & 46540 & 0.011 & 604.8 & 117 & 81 \\ \midrule
Average &  &  & 0.023 &  &  &  \\ \bottomrule
\end{tabular}
\end{table}
}

{
\section{Price of perfect efficiency}
We also investigated the {\em price of perfect efficiency} of a solution considered ``efficient'' (such as MinUnD or MinIAAF), expressed simply by the Gini index of the unsatisfied demand values across individuals. 
In our test instances, perfect efficiency was associated with Gini indices ranging up to 0.56, with an average value of 0.19. Fig.~\ref{fig:Gini_obj_comparision} presents the average Gini index values for each problem instance type, comparing solutions obtained by using the objectives UnD and IAAF. 

\begin{figure}[!htb]
\centering
  \includegraphics[width=85ex]{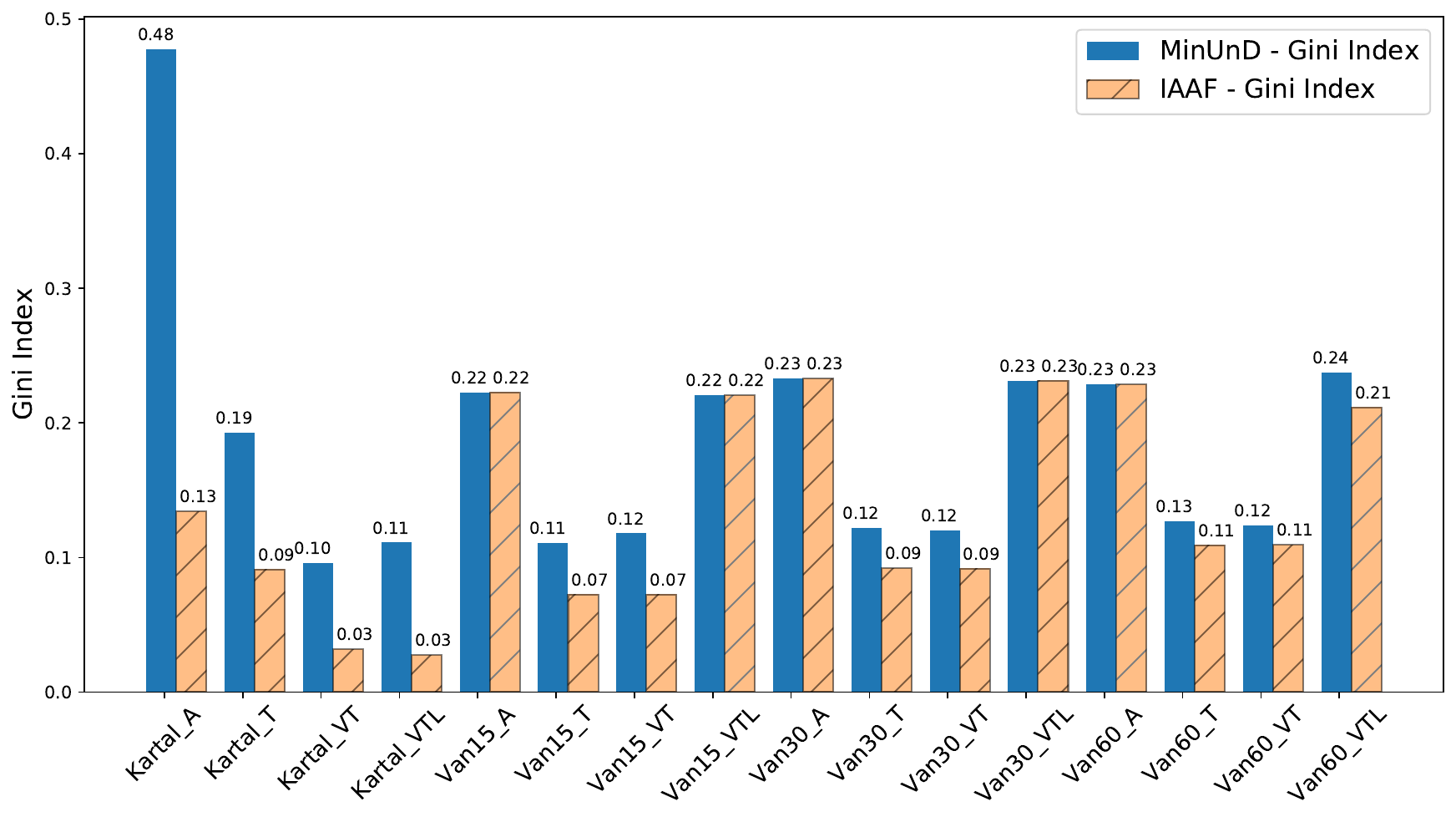}
  \vspace{-15pt}
  \caption{\centering Comparing Gini index achieved by models with MinUnD and MinIAAF objectives}
  \label{fig:Gini_obj_comparision}
\end{figure}
}
\newpage
{\small
\begin{spacing}{0.8}
\bibliography{refd}
\end{spacing}
\bibliographystyle{apalike}
}